\documentclass[lettersize,journal]{IEEEtran}

\usepackage{amsmath,amsfonts}
\usepackage{algorithm}
\usepackage{algpseudocode}
\usepackage{array}
\usepackage{textcomp}
\usepackage{stfloats}
\usepackage{url}
\usepackage{verbatim}
\usepackage{graphicx}
\usepackage{cite}
\usepackage{caption}
\usepackage{subcaption}
\usepackage{amsthm}
\usepackage{bm}
\usepackage[hidelinks]{hyperref}
\usepackage{url}
\usepackage[capitalise]{cleveref}
\usepackage{color}
\usepackage[most]{tcolorbox}
\usepackage{booktabs}
\usepackage{makecell}
\usepackage{multirow}
\usepackage{pifont}
\usepackage{newtxtext}
\usepackage{titletoc}
\usepackage{longtable}

\newcommand{\circlegray}{\textcolor{gray}{\ding{108}}} 
\newcommand{\cmark}{\textcolor{green!70!black}{\ding{51}}} 
\newcommand{\xmark}{\textcolor{red}{\ding{55}}}            

\newtheorem{theorem}{Theorem}

\newtheorem{lemma}{Lemma}
\newtheorem{remark}{Remark}
\newtheorem{definition}{Definition}

\newtheorem{property}{Property}
\newtheorem{example}{Example}
\newtheorem{problem}{Problem}
\usepackage{xcolor}

\xdefinecolor{blueZ}{RGB}{0,77,163}
\xdefinecolor{redZ}{RGB}{232,61,117}

\newcommand{\todo}[1]{{\color{black} #1}}

\usepackage{tikz,pgfplots}
\usepackage{float}
\usetikzlibrary{arrows,shapes,backgrounds,patterns,fadings,matrix,arrows,calc,
	intersections,decorations.markings,
	positioning,arrows.meta}
\usepgfplotslibrary{fillbetween}
\usepgfplotslibrary{statistics}
\pgfplotsset{width=5\columnwidth /5, compat = 1.13,
	height = 60\columnwidth /100, grid= major,
	legend cell align = left, ticklabel style = {font=\scriptsize},
	every axis label/.append style={font=\small},
	legend style = {font={\scriptsize}},title style={yshift=-7pt, font = \small} }

\usepackage[switch]{lineno}

\begin{document}
	
	\title{
		UniConFlow: A Unified Constrained Flow-Matching Framework for Certified Motion Planning
	}
	
	\author{
		Zewen Yang$^{*1}$,~\IEEEmembership{Member,~IEEE},
		Xiaobing Dai$^{*2}$, 
		Dian Yu$^{*3}$,~\IEEEmembership{Graduate Student Member,~IEEE},
		Zhijun Li$^{4}$,~\IEEEmembership{Fellow,~IEEE}, \\
		Majid Khadiv$^{3}$,~\IEEEmembership{Senior Member,~IEEE}, 
		Sandra Hirche$^{2,5}$,~\IEEEmembership{Fellow,~IEEE},
		Sami Haddadin$^{6}$,~\IEEEmembership{Fellow,~IEEE}
		\thanks{
			$^{*}$Equal contribution.
			}
		\thanks{
			Corresponding author: Xiaobing Dai \textless{}xiaobing.dai@tum.de\textgreater{}.
			}
		\thanks{
			This work is supported by the Deutsche Forschungsgemeinschaft (DFG) in the programme of “Research Training Groups”. Joint project METEOR (Machine Learning \& Control Theory), DFG Research Training Group, by the DAAD programme Konrad Zuse Schools of Excellence in Artificial Intelligence sponsored by the Federal Ministry of Education and Research, by the Federal Ministry of Research, Technology, and Space (BMFTR) for its support as part of the research program Communication Systems “Souverän. Digital. Vernetzt.”. Joint project 6G-life, project identification number: 16KIS2414, by the National Key Research and Development Program of China underGrant 2021YFF0501600, by the Huawei-TUM joint laboratory.
			}
		\thanks{
			$^{1}$Zewen Yang is with the Chair of Robotics and Systems Intelligence (RSI), Munich Institute of Robotics and Machine Intelligence (MIRMI), Technical University of Munich (TUM), 80992 Munich, Germany.
			$^{2}$Xiaobing Dai and Sandra Hirche are with the Chair of Information-oriented Control (ITR), School of Computation, Information and Technology (CIT), Technical University of Munich (TUM), 80333 Munich, Germany.
			$^{3}$Dian Yu and Majid Khadiv are with the Chair of AI Planning in Dynamic Environments, Munich Institute of Robotics and Machine Intelligence (MIRMI), Technical University of Munich (TUM), 80992 Munich, Germany.
			$^{4}$Zhijun Li is with the School of Mechanical Engineering, Translational Research Center, Tongji University, Shanghai 201804, China, affiliated with Shanghai Yangzhi Rehabilitation Hospital and also with Shanghai Key Laboratory of Wearable Robotics and Human-Machine Interaction, and also with Department of Automation, University of Science and Technology of China, Hefei 230026, China.
			$^{5}$Sandra Hirche is with the Munich Institute of Robotics and Machine Intelligence, and the Munich Data Science Institute, Technical University of Munich, Germany.
			$^{6}$Sami Haddadin is with the Mohamed Bin Zayed University of Artificial Intelligence, Abu Dhabi 23201, United Arab Emirates.
			}
	}
	
	\maketitle
	
	\begin{abstract}
		Generative models have become increasingly powerful tools for robot motion generation, enabling flexible and multimodal trajectory generation across various tasks. Yet, most existing approaches remain limited in handling multiple types of constraints. In this work, we propose UniConFlow, a unified constrained flow matching-based framework for trajectory generation that incorporates both equality and inequality constraints. Moreover, UniConFlow introduces a novel prescribed-time zeroing function that shapes a time-varying guidance field during inference, allowing the generation process to adapt to varying system models and task requirements. To further address the computational challenges of long-horizon and high-dimensional trajectory generation, we propose two practical strategies for the terminal constraint enforcement and inference process. Empirical validation across three experiments demonstrates that UniConFlow outperforms state-of-the-art generative planners and conventional optimization baselines, achieving superior performance on certified motion planning metrics such as safety, kinodynamic consistency, and action feasibility. 
		Project page is available at: \url{https://uniconflow.github.io}.
	\end{abstract}
	

	\section{Introduction}
	With the rapid development of robotic systems, such as quadrupeds~\cite{ha_IJRR2025_learning}, underactuated vehicles~\cite{Yang_EAAI2025_Safe}, and humanoid robots~\cite{Rozlivek_TRO2025_HARMONIOUS}, generating feasible and robust motion remains a longstanding challenge. 
	Conventional optimization-based methods often struggle with the high dimensionality and non-convexity of these tasks. 
	While imitation learning has emerged as a promising alternative, offering an efficient framework to learn behavioral policies directly from expert demonstrations~\cite{Ablett_TRO2025_Multimodal,Kim_TRO2024_Goal}. 
	However, deployment of such policies in real-world scenarios is hindered by critical challenges, including strong temporal dependencies, high-precision control requirements, and multimodal action distributions~\cite{Urain_TRO2025_A_Survey}.
	
	To address these limitations, generative models have recently emerged as powerful tools for robot policy learning. 
	Among these, methods such as Diffuser \cite{Janner_ICML2022_Planning}, Diffusion Policy \cite{Chi_RSS2023_Diffusion}, and Diffusion BC \cite{CarvalhoIROS2023Motion, pearceICLR2023Imitating} formulate policy learning as a conditional generative process for motion trajectories or actions. 
	These approaches typically corrupt trajectories with Gaussian noise through a forward diffusion process and train a model to reconstruct them via denoising processes. 
	Building on this foundation, works such as ChainedDiffuser \cite{Xian_CORL2023_ChainedDiffuser}, 3D Diffusion Policy \cite{Ze_RSS2024_DP3}, and 3D Diffuser Actor \cite{Ke_CoRL2024_3D}, among others, have extended diffusion-based models to more complex visuomotor settings, demonstrating strong generalization across long-horizon planning tasks and high-dimensional manipulation scenarios.
	Despite these advances, diffusion-based methods suffer from several key limitations. 
	Their reliance on a predefined forward diffusion process, typically Gaussian, not only constrains model flexibility but also incurs significant computational overhead due to the iterative nature of inference. 
	These multi-step sampling operations are particularly problematic in time-sensitive robotic applications. 
	
	Recent advancements in generative modeling have identified flow matching (FM)~\cite{Lipman_ICLR2023_Flow,liu2023ICLRflow} as a compelling alternative to diffusion models for robot learning. 
	FM leverages ordinary differential equations (ODEs) to transport noise to data, enabling greater numerical stability and faster inference compared to the iterative sampling of diffusion models~\cite{chisariCoRL2024learning,huNeurIPS2024AdaFlow}. 
	Conditional flow matching (CFM) extends FM to support conditional generation using inputs such as visual observations or task embeddings. 
	Applications of FM in robotics have so far focused on specific scenarios, including 3D point cloud~\cite{ZhangAAAI2025FlowPolicy}, Riemannian manifolds~\cite{BraunIROS2024Riemannian}, behavioral cloning~\cite{funk2024actionflowequivariantaccurateefficient,RouxelHumanoids2024Flow}, and language-conditioned planning~\cite{xuCoRL2024flow,zhang2025affordancebasedrobotmanipulationflow}. 
	While these generative frameworks improve learning efficiency and expressiveness, a central challenge in robot motion planning is to generate trajectories that are simultaneously safe, feasible, and {physically consistent}. 
	Concretely, in the paper, we consider that the generated trajectory of the robots must satisfy three conditions. 
	First, \emph{kinodynamic consistency}: the sequence of states must be physically reachable, meaning each successive state must follow from the previous one according to the robot's equations of dynamics. 
	Second, \emph{state safety}: at every time step, the robot's configuration must remain within a predefined safe region of the state space, for instance avoiding collisions with obstacles or respecting workspace boundaries. 
	Third, \emph{action feasibility}: the control commands required to execute the trajectory must lie within the robot's physical actuation limits, such as joint torque or velocity bounds. A trajectory satisfying all three conditions simultaneously is called a \textit{certified trajectory}.

	Recent efforts to incorporate constraints into generative planning frameworks 
	have pursued three main strategies: embedding constraint specifications during 
	model training~\cite{Ajay_ICLR2023_Is,zheng_arxiv2023_guidedflowsgenerativemodeling}, 
	integrating corrective guidance signals throughout the sampling 
	procedure~\cite{Yuan_ICCV2023_PhysDiff,Utkarsh_NeurIPS2025_Physics}, 
	or enforcing constraint adherence through post-generation 
	correction~\cite{Maze_AAAI2023_Diffusion,Giannone_NeurIPS2023_Aligning}. 
	Despite yielding improvements in constraint compliance, each paradigm exhibits 
	critical shortcomings: Training-time approaches lack the flexibility to 
	generalize to previously unseen constraints at deployment. Guidance-driven 
	methods provide directional biases toward constraint satisfaction but cannot 
	furnish formal guarantees. Post-generation correction strategies degrade 
	generative fidelity by pushing samples away from the learned distribution. 
	Most closely related to our work are SafeDiffuser~\cite{xiao_ICLR2025_Safediffuser} 
	and SafeFlow~\cite{dai2025safeflowmatchingrobot}, both of which apply control barrier function (CBF) guidance during the denoising process. 
	However, both methods are limited to safety-type inequality constraints 
	(e.g., obstacle avoidance) and neither addresses kinodynamic feasibility or 
	action admissibility. 
	SafeDiffuser enforces the classical CBF forward-invariance condition at each 
	denoising step, which requires the initial sample to be in the safe set. 
	Since sampling begins from a Gaussian prior, its theoretical safety certificate is therefore invalidated from the very first step. 
	SafeFlow partially addresses this by scheduling the constraint via a 
	prescribed-time function, however, its monotonically decreasing schedule 
	actively enforces the constraint throughout the entire generative 
	process, risking out-of-distribution deviations.
	To systematically characterize these distinctions, we summarize the 
	properties of diffusion- and flow-based motion planners in~\cref{tab_compare}.
	These fundamental limitations in robot motion planning reveal three challenges that must be simultaneously resolved: 
	(1) providing rigorous guarantees for state and action constraint satisfaction, 
	(2) preserving dynamic consistency to ensure physical feasibility of generated trajectories, and 
	(3) enabling the model to respect novel, previously unseen constraints at test time without necessitating costly model retraining.

	In this paper, we propose UniConFlow, a unified constrained flow-matching framework for trajectory generation that systematically enforces both equality and inequality constraints within a single generative architecture. Unlike existing guidance-based approaches that rely on heuristic penalty terms or post-processing, UniConFlow employs a novel prescribed-time zeroing function to modulate the inference dynamics, enabling efficient and training-free constraint satisfaction at test time without requiring retraining or auxiliary control modules. 
	By reformulating state and action constraints as barrier certification \cite{Ames_TAC2017_Control} and kinodynamic consistency as Lyapunov certification \cite{sontag1983lyapunov}, we derive principled guidance signals through a quadratic programming (QP) formulation. 
	The main contributions of this work are summarized as follows.
	\begin{itemize}
		\item We propose UniConFlow, a novel unified generation framework designed to ensure both equality and inequality constraints. 
		It integrates a guidance mechanism with stability and safety into the generative process, guaranteeing constraint satisfaction after generation. 
		\item  We introduce a prescribed-time zeroing function and integrate it into the UniConFlow framework, such that the guidance input remains small in the early stages of generation, helping UniConFlow keep the trajectories as close as possible to the training distribution.
		\item We provide the specific implementation details of UniConFlow on certified motion planning tasks, which are categorized into two types, namely trajectory planning and path planning, according to different data types. 
		\item To address the computational burden of long-horizon and high-dimensional trajectory generation, we propose two practical strategies for terminal constraint enforcement: (1) a violation-segment extraction protocol that precisely localizes constraint-violating portions of trajectories, enabling targeted refinement, and (2) a trajectory compression method to accelerate the optimization process in a reduced-dimensional space while preserving high reconstruction accuracy during decoding.
		\item We empirically validate UniConFlow on three diverse tasks: (1) a toy example involving a double inverted pendulum, (2) a real-to-sim car racing experiment where raw racetrack data is transformed to the proposed simulation environment, and (3) a sim-to-real manipulation task using a Franka robotic arm. 
		We compare against state-of-the-art generative and optimization-based approaches, demonstrating the superiority of UniConFlow in certified motion planning under constraints.
	\end{itemize}
	

	\begin{table*}[t]
		\centering
		\caption{Comparison of representative diffusion- and flow-based motion 
			planners across key properties. {Long-Horizon Planning}: the method 
			generates a complete task-level trajectory rather than a single short-horizon 
			action chunk. {Dataset Creation}: the method provides an explicit data 
			generation pipeline or a publicly released dataset. {Real-World 
				Experiments}: the method is validated on a physical robotic platform. 
			\cmark~denotes that the property is explicitly 
			addressed in the respective work, \xmark~that it is not, and \circlegray~that it is not applicable.}
		\label{tab_compare}
		\begin{tabular}{lccccccc}
			\hline
			Method & \makecell{Obstacle\\ Avoidance} & \makecell{Safety\\ Guarantee} & \makecell{Dynamics \\ Consistency} & \makecell{Admissible\\ Action} & \makecell{Long-Horizon \\ Planning}& \makecell{Dataset \\ Creation} & \makecell{Real-world \\ Experiments}\\
			\hline 
			Diffuser~\cite{Janner_ICML2022_Planning} & \xmark & \xmark & \xmark &\xmark&\xmark & \cmark &\xmark\\
			Decision Diffuser~\cite{Ajay_ICLR2023_Is} & \cmark & \xmark &\cmark &\xmark  &\xmark & \cmark &\xmark \\
			Diffusion-based Policy~\cite{Chi_RSS2023_Diffusion,Ze_RSS2024_DP3,Ke_CoRL2024_3D} & \xmark & \xmark &\circlegray &\xmark   &\xmark & \cmark &\cmark \\
			ChainedDiffuser~\cite{Xian_CORL2023_ChainedDiffuser}& \xmark & \xmark &\circlegray &\circlegray  &\cmark & \cmark &\cmark \\
			DPCA~\cite{spieler2024diffusion} & \cmark& \xmark & \xmark & \xmark  &\xmark & \xmark & \xmark \\
			DPCC~\cite{Romer_L4DC2025_Diffusion}& \cmark& \xmark & \cmark & \cmark & \xmark & \cmark & \xmark \\
			HDP~\cite{Ma2024CVPRHierarchical} & \xmark & \xmark & \cmark& \cmark& \xmark& \xmark & \cmark \\
			CoBL~\cite{Mizuta_IROS2024_CoBL} & \cmark& \xmark & \circlegray& \circlegray&  \cmark&\xmark& \xmark \\
			SafeDiffuser~\cite{xiao_ICLR2025_Safediffuser} & \cmark&\xmark& \circlegray & \circlegray & \cmark & \xmark & \xmark  \\
			PCFM~\cite{Utkarsh_NeurIPS2025_Physics} & \cmark & \circlegray& \circlegray & \circlegray  & \xmark& \cmark & \xmark \\
			SafeFlow~\cite{dai2025safeflowmatchingrobot} & \cmark & \cmark& \circlegray & \circlegray  & \cmark& \cmark & \cmark \\
			\hline
			UniConFlow (Ours) & \cmark & \cmark& \cmark& \cmark &  \cmark & \cmark& \cmark\\
			\hline
		\end{tabular}
		\vspace{-0.3cm}
	\end{table*}

	\section{Preliminaries}
	
	\subsection{Flow Matching Model}
	Flow matching is a family of generative models that learns to transform a simple prior distribution into a complex target distribution by matching velocity fields over time~\cite{Lipman_ICLR2023_Flow}. 
	The generation task aims to obtain a sample $\bm{\mathcal{T}} \in \mathbb{T} \subseteq \mathbb{R}^d$ from a distribution $q$ with $d \in \mathbb{N}_{>0}$, i.e., $\bm{\mathcal{T}} \sim q$.
	In practice, the true distribution $q$ is difficult to obtain, which is usually learned from the data set $\mathbb{D}_{\mathcal{T}} = \{ \bm{\mathcal{T}}^{(\iota)} \sim q \}_{\iota = 1, \cdots, M_{\mathcal{T}}}$ with $M_{\mathcal{T}} \in \mathbb{N}$.
	Specifically, flow matching focuses on predicting the velocity fields $\bm{v}(\cdot, \cdot): [0,1] \times \mathbb{T} \to \mathbb{R}^d$ over $t \in [0,1]$ to transform a simple known prior distribution $p$ into a complex unknown target $q$ \cite{Lipman_ICLR2023_Flow}.
	The matching of flow $\bm{v}(\cdot, \cdot)$ is conducted using parametric machine learning techniques with parameter $\bm{\theta} \in \Theta \subseteq \mathbb{R}^{d_{\theta}}$, and the estimated flow is denoted as $\bm{v}(\cdot, \cdot, \cdot): [0,1] \times \mathbb{T} \times \Theta \to \mathbb{R}^{d}$ simplified as $\bm{v}^{\bm{\theta}}_t(\bm{\mathcal{T}}) = \bm{v}(t, \bm{\mathcal{T}}, \bm{\theta})$.
	The induced flow map $\boldsymbol{\psi}_t(\cdot)$ with $t \in [0,1]$ generates a continuous probability path $p_t$ transitioning from $p_0=p$ to the target distribution $p_1=q$, such that
	\begin{align}
		\bm{\mathcal{T}}_t := \boldsymbol{\psi}_t(\bm{\mathcal{T}}_0) \sim p_t,
	\end{align}
	given an initial sample $\bm{\mathcal{T}}_0 \sim p$ at $t=0$. 
	The flow matching model is governed by an ordinary differential equation (ODE) with $\bm{\mathcal{T}}_0 \sim p_0 = p$, i.e.,
	\begin{align}
		\label{eqn_flow_matching_process}
		\mathrm{d} \boldsymbol{\psi}_t(\boldsymbol{\tau}) / \mathrm{d}t = \boldsymbol{v}_t^{\boldsymbol{\theta}}(\boldsymbol{\psi}_t(\boldsymbol{\tau})),
	\end{align}
	where $\bm{\psi}_t(\bm{\mathcal{T}}_0)$ simplified as $\bm{\mathcal{T}}_t \in \mathbb{R}^{d}$ is written as
	\begin{align}
		\bm{\mathcal{T}}_t = \bm{\psi}_t(\bm{\mathcal{T}}_0) = \bm{\mathcal{T}}_0 + \int\nolimits_0^t \bm{v}^{\bm{\theta}}_s(\bm{\psi}_s(\bm{\mathcal{T}}_0)) \mathrm{d} s
	\end{align}
	forming the probability path $p_t$, i.e., $\bm{\mathcal{T}}_t \sim p_t$.
	Given a training data $\bm{\mathcal{T}}_1 = \bm{\mathcal{T}} \in \mathbb{D}_{\mathcal{T}}$ satisfying $\bm{\mathcal{T}}_1 \sim p_1 = q$ and a noise sample $\bm{\mathcal{T}}_0 \sim p_0$, the deterministic path is given by
	\begin{align}
		\bm{\mathcal{T}}_t = \alpha(t) \bm{\mathcal{T}}_1 + \beta(t) \bm{\mathcal{T}}_0,
	\end{align}
	where $\alpha(\cdot), \beta(\cdot): [0,1] \to \mathbb{R}$ are predefined interpolation schedules satisfying $\alpha(0) = 0$, $\alpha(1) = 1$, $\beta(0) = 1$, $\beta(1) = 0$.
	Correspondingly, the target velocity $\bm{v}_t(\cdot) = \bm{v}(t, \cdot)$ along this interpolation path is written as
	\begin{align}
		\bm{v}_t(\bm{\mathcal{T}}_t) = \dot{\alpha}(t) \bm{\mathcal{T}}_1 + \dot{\beta}(t) \bm{\mathcal{T}}_0,
	\end{align}
	such that the parametric model characterized by $\bm{\theta}$ $\bm{v}_t^{\bm{\theta}}$ is trained to approximate this marginal velocity field by minimizing the flow matching loss defined as 
	\begin{align} \label{eqn_FM_loss}
		\mathcal{L}_{\text{FM}}(\bm{\theta}) = \mathbb{E}_{t \sim \mathcal{U}[0,1],\bm{\mathcal{T}}_t \sim p_t} \| \bm{v}_t^{\bm{\theta}}(\bm{\mathcal{T}}_t) - \bm{v}_t(\bm{\mathcal{T}}_t) \|_2^2.
	\end{align}
	Note that the direct computing of $\mathcal{L}_{\text{FM}}(\bm{\theta})$ in \eqref{eqn_FM_loss} is intractable due to the marginalization.
	Instead of direct evaluation, the conditional flow matching (CFM) loss is applied as
	\begin{align}
		\mathcal{L}_{\text{CFM}}(\bm{\theta}) = \mathbb{E}_{t, \bm{\mathcal{T}}_t, \bm{\mathcal{T}}} \| \boldsymbol{v}_t^{\boldsymbol{\theta}}(\bm{\mathcal{T}}_t) - \bm{v}_t(\bm{\mathcal{T}}_t | \bm{\mathcal{T}} ) \|_2^2,
	\end{align}
	where $t\sim \mathcal{U}[0,1]$, $\bm{\mathcal{T}}_t \sim p_t$ and $\bm{\mathcal{T}} = \bm{\mathcal{T}}_1 \sim q$.
	
	Note that the generation process in \eqref{eqn_flow_matching_process} is unconstrained, and in practice the sample $\bm{\mathcal{T}}$ requires to satisfy multiple constraints both in equality and inequality forms.
	Specifically, denote the $i$-th equality and $j$-th inequality constraints as $g_i(\bm{\mathcal{T}}) = 0$ and $h_j(\bm{\mathcal{T}}) \le 0$ for $i = 1, \cdots, N_g$ and $j = 1, \cdots, N_h$ respectively, where $N_g$ and $N_h$ represents the number of equality and inequality constraints.
	In the following subsections, control-inspired methods are introduced to address constraints.
	
	\subsection{Lyapunov Certification with Equality Constraints}
	\label{subsection_Lyapunov_certificate}
	In this subsection, the equality constraints $g_i(\bm{\mathcal{T}}) = 0$ are encoded into constraint function $g_i(\cdot): \mathbb{T} \to \mathbb{R}$ for all $i = 1, \cdots, N_g$ and $N_g \in \mathbb{N}$. 
	When the sample $\bm{\mathcal{T}}$ violates the constraint, we need to adjust the value of $g_i(\bm{\mathcal{T}}), \forall i = 1, \cdots, N_g$ such that it can approach $0$ by changing the evolution direction of $\bm{\mathcal{T}}$, i.e., $\dot{\boldsymbol{\mathcal{T}}}$. 
	This strategy is formulated as
	\begin{align} \label{eqn_gradient_descent}
		\dot{g}_i(\bm{\mathcal{T}}) = \frac{\mathrm{d} g_i(\bm{\mathcal{T}})}{\mathrm{d} \bm{\mathcal{T}}} \dot{\bm{\mathcal{T}}}: \begin{cases}
			< 0, & \text{if}~ g_i(\bm{\mathcal{T}}) > 0 \\
			= 0, & \text{if}~ g_i(\bm{\mathcal{T}}) = 0 \\
			> 0, & \text{otherwise}
		\end{cases}
	\end{align}
	for all $i = 1, \cdots, N_g$.
	Utilizing the sign of $g_i(\bm{\mathcal{T}})$, the formulation in \eqref{eqn_gradient_descent} is written as
	\begin{align} \label{eqn_gradient_descent_signed}
		\mathrm{sign}(g_i(\bm{\mathcal{T}})) \dot{g}_i(\bm{\mathcal{T}}) < - \gamma_g(| g_i(\bm{\mathcal{T}}) |)
	\end{align}
	for all $g_i(\bm{\mathcal{T}}) \ne 0$ and $\dot{g}_i(\bm{\mathcal{T}}) = 0$ if $g_i(\bm{\mathcal{T}}) = 0$, where $\gamma_g(\cdot): \mathbb{R}_{\ge 0} \to \mathbb{R}_{\ge 0}$ is continuous satisfying $\gamma_g(0) = 0$ and $\mathrm{sign}(\cdot): \mathbb{R} \to \{ -1, 0, 1 \}$ denotes a sign function as
	\begin{align}
		\mathrm{sign}(g) = \begin{cases}
			-1, & \text{if}~ g < 0 \\
			0, & \text{if}~ g = 0 \\
			1, & \text{otherwise}
		\end{cases}.
	\end{align}
	To avoid the discontinuity from $\mathrm{sign}(\cdot)$, an auxiliary function $V_i(\cdot): \mathbb{R} \to \mathbb{R}_{\ge 0}$ is introduced to quantify the constraint violation level of $g_i(\cdot)$, which can be expressed as
	\begin{align}
		V_i(g_i(\bm{\mathcal{T}})) = \gamma_{V,g,i}(| g_i(\bm{\mathcal{T}}) |), && \forall i = 1, \cdots, N_g
	\end{align}
	with the scalar function $\gamma_{V,g,i}(\cdot): \mathbb{R}_{\ge 0} \to \mathbb{R}_{\ge 0}$ belonging to class-$\mathcal{K}$ defined as follows.
	
	
	Using the introduced $V_i(\cdot)$, the condition in \eqref{eqn_gradient_descent} or \eqref{eqn_gradient_descent_signed} can also be reformulated as
	\begin{align} \label{eqn_CLF_short}
		\dot{V}_i(g_i(\bm{\mathcal{T}})) \le - \gamma_{V,i}(V_i(g_i(\bm{\mathcal{T}}))), && \forall i = 1, \cdots, N_g
	\end{align}
	with $\gamma_{V,i}(\cdot) \in \mathcal{K}$, which is similar to the stability condition in control theory \cite{khalil2015nonlinear}.
	Specifically, applying the Lyapunov theory, the constraint satisfaction under such update law for $\bm{\mathcal{T}}$ is summarized as follows.
	
	\begin{lemma}[Lyapunov Certificate for Equality Constraint]
		\label{def_clf}
		Given $N_g$ equality constraints $g_i(\cdot)=0$ for $i = 1, \cdots, N_g$ and any initial value $\bm{\mathcal{T}}(0)$.
		Let the update rate for $\bm{\mathcal{T}}$ satisfy
		\begin{align} \label{eqn_CLF}
			\frac{d V_i(g_i(\bm{\mathcal{T}}))}{d g_i(\bm{\mathcal{T}})} \frac{d g_i(\bm{\mathcal{T}})}{d \bm{\mathcal{T}}} \dot{\bm{\mathcal{T}}} \le - \gamma_{V,i}(V_i(g_i(\bm{\mathcal{T}}))),
		\end{align}
		for all $i = 1, \cdots, N_g$, then all equality constraints are satisfied asymptotically, i.e., $\lim_{t \to \infty} g_i(\bm{\mathcal{T}}(t)) = 0$.
	\end{lemma}
	\begin{IEEEproof}
		According to Lyapunov stability theory~\cite{sontag1983lyapunov}, it has $\lim_{t \to \infty} V_i(g_i(\bm{\mathcal{T}}(t))) = 0$, leading to 
		\begin{align}
			\lim\nolimits_{t \to \infty} | g_i(\bm{\mathcal{T}}(t)) | =& \lim\nolimits_{t \to \infty} \gamma_{V,g,i}^{-1}(V_i(g_i(\bm{\mathcal{T}}(t)))) \\
			=& \lim\nolimits_{t \to \infty} \gamma_{V,g,i}^{-1}(0) = 0 \nonumber
		\end{align}
		and then $\lim_{t \to \infty} g_i(\bm{\mathcal{T}}(t)) = 0$.
	\end{IEEEproof}
	
	\cref{def_clf} shows the relationship between the constraint satisfaction process and control theory, where an example for a possible update law for a single $g_i(\cdot)$ denotes
	\begin{align}
		\dot{\bm{\mathcal{T}}} =& - \lambda_{V,i} \Big( \frac{d g_i(\bm{\mathcal{T}})}{d \bm{\mathcal{T}}} \Big)^T \Big( \frac{d V_i(g_i(\bm{\mathcal{T}}))}{d g_i(\bm{\mathcal{T}})} \Big)^T, \\
		\lambda_{V,i} =& \gamma_{V,i}(V_i(g_i(\bm{\mathcal{T}}))) \Big\| \frac{d g_i(\bm{\mathcal{T}})}{d \bm{\mathcal{T}}} \Big\|^{-2} \Big\| \frac{d V_i(g_i(\bm{\mathcal{T}}))}{d g_i(\bm{\mathcal{T}})} \Big\|^{-2},
	\end{align}
	following the steepest gradient descent.
	However, finding a proper update law for multiple $g_i(\cdot)$ with $i = 1, \cdots, N_g$ requires optimization based methods, taking \eqref{eqn_CLF} as constraints.
	More details are shown in \cref{subsection_equality_constraint}.
	\subsection{Barrier Certification with Inequality Constraints}
	\label{subsection_barrier_certificate}
	The equality constraint satisfaction process can be formulated as a stability control task, wherein Lyapunov stability theory is employed to ensure convergence to a single equilibrium point. 
	In contrast, inequality constraints characterize an admissible set $\bm{\mathcal{T}}$, where the control objective is to ensure forward invariance of this set rather than convergence to a single point.
	Here, we consider $N_h$ inequality constraints $h_j(\cdot) \le 0$ with $N_h \in \mathbb{N}$, and each inequality constraint describe a feasible set of $\bm{\mathcal{T}}$ as
	\begin{align}
		\mathbb{T}_{j} = \{ \bm{\mathcal{T}} \in \mathbb{T} | - h_j(\bm{\mathcal{T}}) \ge 0 \} \subseteq \mathbb{T}
	\end{align}
	for all $j = 1, \cdots, N_h$. 
	Similar to the Lyapunov method in \eqref{eqn_CLF_short}, the barrier certificate restricts the derivative of $h_j(\cdot)$ as
	\begin{align}
		- \dot{h}_j(\bm{\mathcal{T}}) \ge - \gamma_{h,j}( - h_j(\bm{\mathcal{T}}) ), && \forall j = 1, \cdots, N_h
	\end{align}
	with $\gamma_{h,j}(\cdot): \mathbb{R} \to \mathbb{R}$ as an extended class-$\mathcal{K}$ function, which is defined as follows.
	\begin{definition} [Extended Class-$\mathcal{K}$ Function]
		A function $\gamma(\cdot): \mathbb{R} \to \mathbb{R}$ is called an extended class-$\mathcal{K}$ function, i.e., $\gamma(\cdot) \in \mathcal{K}_e$, if it is continuous, strictly monotonically increasing and satisfies $\gamma(0) = 0$.
	\end{definition}
	
	Following the ideas of control barrier functions (CBFs)~\cite{Ames_TAC2017_Control}, the satisfaction maintenance of inequality constraints is shown below, ensuring the state $\bm{\mathcal{T}}$ never exits the prescribed safe region $\mathbb{T}_{j}$ throughout future evolution.
	\begin{lemma}[Barrier Certificate for Inequality Constraint]
		\label{def_cbf}
		Given $N_h$ inequality constraints $h_j(\cdot)$ for $j = 1, \cdots, N_h$, suppose the initial value $\bm{\mathcal{T}}(0)$ satisfies $\bm{\mathcal{T}}(0) \in \bigcap_{j = 1}^{N_h} \mathbb{T}_{j}$.
		If the trajectory $\bm{\mathcal{T}}$ satisfies
		\begin{align} \label{eqn_CBF}
			(\mathrm{d} h_j(\bm{\mathcal{T}}) / \mathrm{d} \bm{\mathcal{T}}) \dot{\bm{\mathcal{T}}} \le - \gamma_{h,j}( - h_j(\bm{\mathcal{T}}) )
		\end{align}
		for all $j = 1, \cdots, N_h$, then the forward invariance of $\mathbb{T}_{j}$ is guaranteed, i.e., $h_j( \bm{\mathcal{T}}(t) ) \le 0$ for all $t \in \mathbb{R}_{\geq 0}$, indicating the satisfaction of all inequality constraints.
	\end{lemma}
	\begin{IEEEproof}
		The proof follows directly from the CBF condition~\cite{Ames_TAC2017_Control} and is omitted for brevity. 
	\end{IEEEproof}
	We remark that, for consistency with the equality constraint formulation in the preceding section, we adopt the convention $h(\cdot)\leq 0$ to define the inequality constraints. This is opposite to the standard CBF formulation, where $h(\cdot)>0$ defines the safe set.
	\cref{def_cbf} shows the inequality constraints $h_j(\cdot)$ are always satisfied, when the update law satisfies \eqref{eqn_CBF}.
	For a single inequality constraint, one update example is 
	\begin{align}
		\dot{\bm{\mathcal{T}}} = - \gamma_{h,j}( - h_j(\bm{\mathcal{T}}) ) \Big\| \frac{\mathrm{d} h_j(\bm{\mathcal{T}})}{\mathrm{d} \bm{\mathcal{T}}} \Big\|^{-1} \Big( \frac{\mathrm{d} h_j(\bm{\mathcal{T}})}{\mathrm{d} \bm{\mathcal{T}}} \Big)^{\top},
	\end{align} 
	using the steepest gradient descent method.
	While the inequality constraint satisfaction is guaranteed in \cref{def_cbf}, the initial condition $\bm{\mathcal{T}}(0) \in \bigcap_{j = 1}^{N_h} \mathbb{T}_{j}$ indicating $h_j(\bm{\mathcal{T}}(0)) \le 0$ for all $j = 1, \cdots, N_h$ does not naturally hold. 
	Consequently, \cite{xiao_ICLR2025_Safediffuser,Mizuta_IROS2024_CoBL} fail in enforcing safety constraints within reverse diffusion models for safe planning due to their technical limitations. 
	Specifically, the condition of forward invariance in CBFs relies on the assumption of initially safe samples~\cite{Ames_TAC2017_Control}, which is fundamentally violated for the reverse diffusion process, where the initial sample is usually obtained from Gaussian distribution. 
	Furthermore, due to the probabilistic nature of diffusion models, there is no guarantee that the generated trajectories will remain within the safe set throughout the process.
	For the case with $h_j(\bm{\mathcal{T}}(0)) > 0$, while the update condition \eqref{eqn_CBF} pushes $\bm{\mathcal{T}}$ gradually towards $\mathbb{T}_{j}$, the entrance into $\mathbb{T}_{j}$ is not guaranteed especially in the worst case with $\dot{h}_j(\bm{\mathcal{T}}) = - \gamma_{h,j}( - h_j(\bm{\mathcal{T}}) )$.
	The design of the update law $\dot{\bm{\mathcal{T}}}$, which guarantees the entrance of $\mathbb{T}_{j}$, is shown later in \cref{subsection_inequality_constraint}.
	
	In the following sections, the constrained generation problem is formally defined, where the equality and inequality constraints are addressed using Lyapunov and barrier certificates.

	\section{Problem Setting}
	\label{sec_problem_setting}
	In this section, a general form of the constrained generation problem is presented in \cref{subsection_problem_constrained_generation} with the introduced flow matching.
	Then, an explicit example for certified motion planning of robotic systems is discussed in \cref{subsection_problem_certified_motion_planning}.
	\subsection{Flow Matching-based Constrained Generation}
	\label{subsection_problem_constrained_generation}
	
	The generation process described by \eqref{eqn_flow_matching_process} is unconstrained, i.e., it does not enforce any structure or consider validity on the generated samples beyond fitting the distribution $q$. 
	However, in many real-world scenarios such as motion planning, photo design, or physical simulation, the generated samples $\bm{\mathcal{T}}$ may be required to satisfy a set of constraints.
	Let $\{g_i(\bm{\mathcal{T}}) = 0\}_{i=1}^{N_g}$ and $\{h_j(\bm{\mathcal{T}}) \leq 0\}_{j=1}^{N_h}$ denote the equality and inequality constraints, respectively. 
	All functions $g_i(\cdot), h_j(\cdot): \mathbb{R}^d \to \mathbb{R}$ are assumed to be known, differentiable, and task-specific for all $i = 1, \cdots, N_g$ and $j = 1, \cdots, N_h$. 
	The constrained generation problem is now formally stated as follows.
	\begin{definition} [Certified Sample]
		\label{definition_constrained_generation} 
		If a sample $\bm{\mathcal{T}}$ satisfies all the following equality and inequality constraints as
		\begin{subequations}
			\label{eqn_constrained_generation}
			\begin{align}
				\label{eqn_constrained_generation_eq}
				& g_i(\bm{\mathcal{T}}) = 0, && \forall i = 1, \cdots, N_g, \\
				\label{eqn_constrained_generation_ineq}
				& h_j(\bm{\mathcal{T}}) \le 0, && \forall j = 1, \cdots, N_h, 
			\end{align}
		\end{subequations}
		then it is called a certified sample.
	\end{definition}
	
	In this work, it is aimed to design a constrained generation framework, which is formulated as follows.
	\begin{problem} \label{problem_certified_FM}
		Design a general framework for the flow matching model, such that all generated samples are certified following \cref{definition_constrained_generation}.
	\end{problem}
	Constrained generation is applicable to a variety of tasks. 
	In this paper, we focus on robot motion planning as a representative use case. The following subsection introduces the certified robot motion planning problem.
	
	\subsection{Certified Robot Motion Planning}
	\label{subsection_problem_certified_motion_planning}
	We consider a general class of robotic systems described at time $\tau \in \mathbb{R}_{\geq 0}$ as 
	\begin{align} \label{eqn_robot_system_continuous_time}
		\dot{\bm{s}}(\tau) = \bm{f}(\tau, \bm{s}(\tau), \bm{a}(\tau)),
	\end{align}
	where $\bm{s} \in \mathbb{S} \subseteq \mathbb{R}^{d_s}$ denotes the system state and $\bm{a} \in \mathbb{A} \subseteq \mathbb{R}^{d_a}$ represents the control input with $d_s, d_a \in \mathbb{N}_{>0}$.
	The function $\bm{f}: \mathbb{S} \times \mathbb{A} \to \mathbb{R}^{d_s}$ encodes kinematics and dynamics and is considered as known and differentiable. 
	
	\begin{example}
		In robotic applications, a rigid-body manipulator is governed by the Euler–Lagrange equations, where the state $\bm{s}$ typically comprises the generalized coordinates $\bm{q} \in \mathbb{R}^{d_q}$ and their time derivatives $\dot{\bm{q}} \in \mathbb{R}^{d_q}$ with $d_q \in \mathbb{N}_{>0}$, i.e., $\bm{s} = [\bm{q}^{\top}, \dot{\bm{q}}^{\top}]^{\top}$ and $d_s = 2 d_q$.
		The control input $\bm{a} \in \mathbb{R}^{d_q}$ with $d_a = d_q$ represents generalized forces or torques acting on the system. 
		Specifically, the system dynamics in Euler–Lagrange form is written as
		\begin{align}
			\label{eq_arm_dynamics}
			\dot{\bm{s}} \!=\!
			[
				\dot{\bm{q}}^{\top},
				(\bm{M}^{-1}(\bm{q}) ( \bm{a} - \bm{C}(\bm{q}, \dot{\bm{q}}) \dot{\bm{q}} \!-\! \bm{g}(\bm{q}) ))^{\top}
			]^{\top},
		\end{align}
		where $\bm{M}(\bm{q}) \in \mathbb{R}^{d_q \times d_q}$ is the positive definite inertia matrix, $\bm{C}(\bm{q}, \dot{\bm{q}}) \in \mathbb{R}^{d_q \times d_q}$ captures Coriolis and centrifugal effects, and $\bm{g}(\bm{q}) \in \mathbb{R}^{d_q}$ represents gravitational forces. 
	\end{example}
	
	For practical data sampling, the action is set to constant during two time instances, i.e., $\bm{a}(\tau) = \bm{a}(\tau_k)$ for any $\tau \in [\tau^k, \tau^{k+1})$ with $k \in \mathbb{N}$, where the time sequence $\{ \tau^k \}_{k \in \mathbb{N}}$ is defined by $\tau^0 = 0$ and 
	\begin{align}
		\tau^{k+1} = \tau^k + \Delta_{\tau}, && \forall k \in \mathbb{N}
	\end{align}
	with constant time interval $\Delta_{\tau} \in \mathbb{R}_{>0}$.
	Then, the continuous dynamical system \eqref{eqn_robot_system_continuous_time} is discretized as 
	\begin{align}
		\bm{s}^{k+1} = \bm{f}^k(\bm{s}^k, \bm{a}^k)
	\end{align}
	for all $k = 0, \cdots, H$ with time horizon $H \in \mathbb{N}_{>0}$, where
	\begin{align}
		\bm{f}^k(\bm{s}^k, \bm{a}^k) = \bm{s}^k + \int\nolimits_{\tau^k}^{\tau^{k+1}} \bm{f}(\zeta, \bm{s}(\zeta), \bm{a}(\tau^k)) \mathrm{d} \zeta
	\end{align}
	is pre-calculated and differentiable.
	The motion planning task is to generate a trajectory $\bm{\mathcal{T}}$ over the finite horizon $H$ as a sequence of states and actions as
	\begin{align} \label{eqn_T}
		\bm{\mathcal{T}} = [ (\bm{s}^0)^{\top}\!, (\bm{a}^0)^{\top}\!, \cdots, (\bm{s}^{H-1})^{\top}\!, (\bm{a}^{H-1})^{\top}\!, (\bm{s}^H)^{\top} ]^{\top},
	\end{align}
	such that the total trajectory dimension is $d = (H+1)d_s + H d_a$.
	Note that the generated trajectory $\bm{\mathcal{T}}$ is certified if it satisfies the conditions of safety, feasibility, and dynamical consistency.
	A formal definition of such a certified trajectory $\bm{\mathcal{T}}$ is shown as follows.
	\begin{definition} [Certified Motion Trajectory] \label{definition_certified_robot_trajectory}
		If a sample trajectory $\bm{\mathcal{T}}$ satisfies the following conditions:
		\begin{itemize}
			\item 
			\textbf{Kinodynamic Consistency}: For all $k = 0, \dots, H-1$, the system evolves according to the known kinodynamics:
			\begin{align}
				\label{eqn_dynamics_consistency}
				\bm{s}^{k+1} = \bm{f}^k(\bm{s}^k, \bm{a}^k);
			\end{align}
			
			\item 
			\textbf{State Constraints for Safety}: For all $k = 0, \dots, H$, the state remains in a predefined safe set as
			\begin{align}
				\label{eqn_state_domain}
				\bm{s}^k \in \mathbb{C}_s \subseteq \mathbb{S};
			\end{align}
			
			\item 
			\textbf{Action Constraints for Feasibility}: The action lies within an admissible set for all $k = 0, \dots, H-1$ as
			\begin{align} \label{eqn_action_domain}
				\bm{a}^k \in \mathbb{C}_a \subseteq \mathbb{A},
			\end{align}
		\end{itemize}
		then the trajectory $\bm{\mathcal{T}}$ is called a certified trajectory with respect to safety, feasibility, and kinodynamical consistency.
	\end{definition}
	
	The constraints in \cref{definition_certified_robot_trajectory} ensure that the generated trajectories respect robot dynamics, avoid unsafe regions, and follow control limitations.
	To generate such certified trajectories without retraining the flow matching model, a constrained motion trajectory generation problem is defined below.
	
	\begin{problem} \label{problem_certified_FM_for_trajectory}
		Design a motion trajectory generation framework with a flow matching model, such that all generated trajectories are certified as defined in \cref{definition_certified_robot_trajectory}, i.e., satisfying safety, feasibility constraints and kinodynamical consistency.
	\end{problem}
	Moreover, we assume that the problem is well-posed, meaning there exists at least one trajectory $\bm{\mathcal{T}} \in \mathbb{T}$ that simultaneously satisfies the kinodynamic consistency, state safety, and action feasibility conditions specified in \cref{definition_certified_robot_trajectory}.
	
	\section{Unified Constrained Generation Framework}
	\label{sec_unified_framework}
	In this section, the constrained generation problem defined in \cref{problem_certified_FM} is addressed, where the equality and inequality constraints are discussed in \cref{subsection_equality_constraint} and \cref{subsection_inequality_constraint}, respectively.
	Then, the unified constrained generation framework with flow matching models is proposed in \cref{subsection_UniConGen} with its feasibility analysis.
	
	\subsection{Equality Constraint}
	\label{subsection_equality_constraint}
	In this subsection, the satisfaction of equality constraints $g_i(\cdot) = 0$ for all $i = 1, \cdots, N_g$ in \eqref{eqn_constrained_generation_eq} is considered within the flow matching framework.
	Different from the procedure in \cref{subsection_Lyapunov_certificate}, where a Lyapunov function $V_i(\cdot)$ is designed for each equality constraint $g_i(\cdot)$ with $i = 1, \cdots, N_g$, now we design a single Lyapunov function considering all constraints.
	Specifically, define the aggregated function as
	\begin{align} \label{eqn_g}
		g(\cdot) = \sum\nolimits_{i = 1}^{N_g} g_i^2(\cdot): \mathbb{R}^d \to \mathbb{R}_{\ge 0},
	\end{align}
	and it is straightforward to see that $g(\bm{\mathcal{T}}) = 0$ indicates $g_i(\cdot) = 0$ for all $i = 1, \cdots, N_g$. 
	The existence of a certified sample satisfying the concatenated equality constraint $g(\cdot)$ requires that there exists at least one sample satisfying all individual equality constraints $g_i(\cdot)$.
	Considering the generation process \eqref{eqn_flow_matching_process} with bounded $t \in [0,1]$, it is desired to guarantee the satisfaction of equality constraint $g(\bm{\mathcal{T}}_1) = 0$ from the initially violated condition $g(\bm{\mathcal{T}}_0) \ne 0$, forming a prescribed-time stability problem defined as follows. 
	\begin{definition}
		\label{definition_PT_stability} 
		Given a prescribed time $T_{\text{pre}} \in \mathbb{R}_{>0}$, a scalar dynamical function $g(\cdot): \mathbb{R}^d \to \mathbb{R}$ achieves prescribed time stability, if for arbitrary initial condition $g(\boldsymbol{\mathcal{T}}_0) \in \mathbb{R}$ it holds 
		\begin{align}
			g(\boldsymbol{\mathcal{T}}_t) = 0, && \forall t \in [T_{\text{pre}}, \infty), 
		\end{align}
		indicating $g_i(\boldsymbol{\mathcal{T}}_t) = 0$ for all $t \in [T_{\text{pre}}, \infty)$ and $i = 1, \cdots, N_g$,
		where the dynamics for $g(\cdot)$ denotes
		\begin{align} \label{eqn_g_dot_T}
			\dot{g}(\bm{\mathcal{T}}_t) = ( \mathrm{d} g(\bm{\mathcal{T}}_t) / \mathrm{d} \bm{\mathcal{T}}_t )^{\top} \dot{\bm{\mathcal{T}}}_t.
		\end{align}
	\end{definition}

	To achieve prescribed time (PT) stability w.r.t $g(\cdot)$, a prescribed time zeroing function (PTZF) is introduced as a reference, whose definition is detailed below.	
	\begin{definition} \label{definition_PTZF}
		Given a prescribed time $T_{\text{pre}} \in \mathbb{R}_{>0}$ and a function $\gamma_r(\cdot, \cdot): [0, T_{\text{pre}}) \times \mathbb{R} \to \mathbb{R}$ satisfying
		\begin{align}
			\gamma_r(t, r) \geq \underline{\gamma}_{r}(r), && \forall t \in [c_r T_{\text{pre}}, T_{\text{pre}}) ~\text{and}~ r \in \mathbb{R},
		\end{align}
		where $c_r \in [0, 1)$ is a factor controls the PT behavior and $\underline{\gamma}_{r}(\cdot): \mathbb{R} \to \mathbb{R}$ is an extended class-$\mathcal{K}$ function.
		Consider a scalar signal $r(\cdot): \mathbb{R}_{\ge 0} \to \mathbb{R}$, which follows the dynamics
		\begin{align} \label{eqn_system_r}
			\dot{r}(t) = \begin{cases}
				- \frac{T_{\text{pre}}}{(T_{\text{pre}} - t)^2} \gamma_r(t, r(t)), & \text{if~} t < T_{\text{pre}} \\
				0, & \text{otherwise}
			\end{cases},
		\end{align}
		then $r(\cdot)$ is a prescribed time zeroing function.
	\end{definition}
	The definition of PTZF in \cref{definition_PTZF} allows a wide range of choice on different extended class-$\mathcal{K}$ functions $\underline{\gamma}_{r}(\cdot)$, leading to different types of PTZFs. 
	Moreover, since all parameters $T_{pre}$, $\gamma_r(\cdot,\cdot)$ and $c_r$ are selected before generation, the explicit expression of $r(\cdot)$ and $\dot{r}(\cdot)$ can be derived analytically or numerically with one example shown below.
	
\begin{figure}
		\centering
		 \includegraphics[width=1\linewidth]{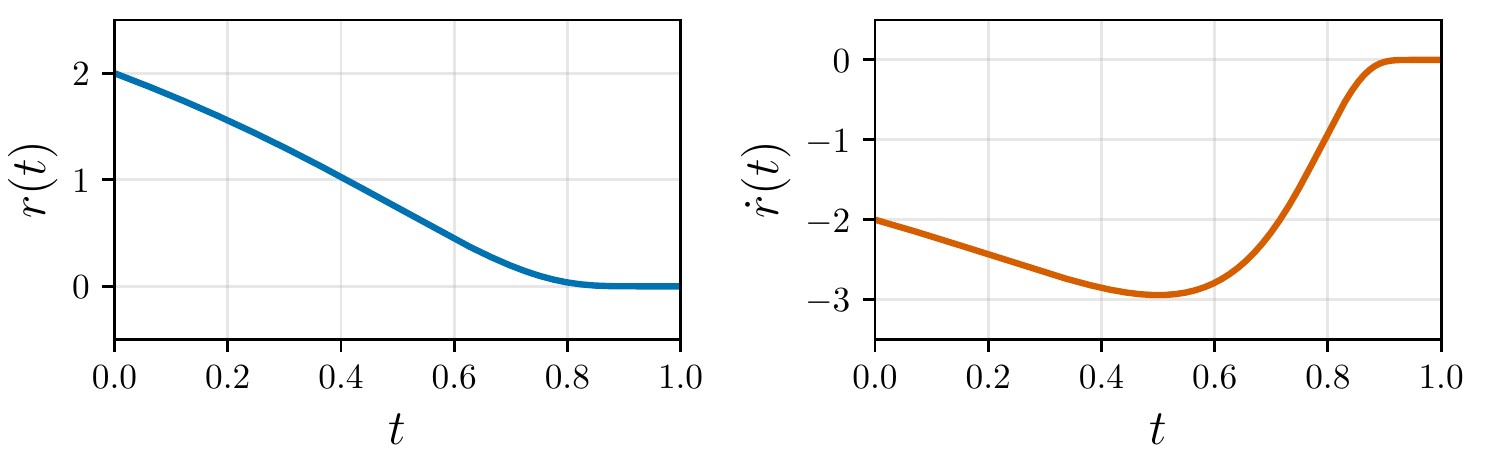}
		\vspace{-0.3cm}
		\caption{Function $r(t)$ over time with $c_r = 0$ and $\gamma_r(t,r) = \underline{\gamma}_r = r$ in \cref{definition_PTZF}. }
		\vspace{-0.3cm}
		\label{fig_rFunction}
	\end{figure}

	\begin{example}
		Let $\gamma_r(t, r) = \underline{\gamma}_{r}(r) = c_{\gamma} r$ with $c_{\gamma} \in \mathbb{R}_{>0}$, then the explicit solution of \eqref{eqn_system_r} is written as
		\begin{align}
			r(t) = \begin{cases}
				r(0) \mathrm{e}^{- c_{\gamma} t / (T_{\text{pre}} - t)}, & \text{if~} t < T_{\text{pre}} \\
				0, & \text{otherwise}
			\end{cases},
		\end{align}
		inducing $\dot{r}(t) = - r(0) T_{\text{pre}} \mathrm{e}^{- c_{\gamma} t / (T_{\text{pre}} - t)} / (T_{\text{pre}} - t)^2$.
	\end{example}
	To illustrate the change over time, we plot the function $r(t)$ for $t\in[0,1)$ with $c_{\gamma}=1$ and $T_{\text{pre}}=1$ in \cref{fig_rFunction}, demonstrating the prescribed time property, where $r(t)$ approaches to zero as $t$ goes from a fixed value to infinite.
	
	Moreover, there exists no restrictions on the initial value $r(0)$ and the transition behavior before $c_r T_{\text{pre}}$, enabling free evolution in the early stage $[0, c_r T_{\text{pre}})$.
	For the late stage with $t \in [c_r T_{\text{pre}}, T_{\text{pre}})$ in transition process, the term $T / (T - t)^2$ grows large as $t$ approaches prescribed time $T_{\text{pre}}$, inducing rapid growing trends of $r(\cdot)$ toward zero level. 
	In detail, the property of PTZF $r(\cdot)$ is shown as follows.
	\begin{property} 
		\label{property_PTZF}
		Given a prescribed time $T_{\text{pre}} \in \mathbb{R}_{>0}$.
		If signal $r(\cdot): \mathbb{R}_{\ge 0} \to \mathbb{R}$ is a PTZF in \cref{definition_PTZF}, it has
		\begin{align}
			r(t) = 0, && \forall t \in [T_{\text{pre}}, \infty)
		\end{align}
		for arbitrary initial value $r(0) \in \mathbb{R}$.
	\end{property}
	\begin{IEEEproof}
		See Appendix \ref{proof_of_property_PTZF}.
	\end{IEEEproof}
	
	\cref{property_PTZF} shows the PTZF always achieve prescribed time stability w.r.t $T_{\text{pre}}$ from arbitrary initial value.
	Given the constrained generation problem in \cref{definition_constrained_generation} with flow matching process \eqref{eqn_flow_matching_process}, the prespecified time $T_{\text{pre}}$ is set as $1$.
	Define a PTZF $\bar{g}(\cdot)$ with $T_{\text{pre}} = 1$, it is straightforward to see that $g(\bm{\mathcal{T}}_1) = 0$ if $g(\bm{\mathcal{T}}_t) \le \bar{g}(t)$ for all $t \in [0,1]$.
	This observation indicates that the generated sample $\bm{\mathcal{T}}_1$ satisfies all equality constraints, i.e., $g_i(\bm{\mathcal{T}}_1) = 0$ for $i = 1, \cdots, N_g$.
	To achieve $g(\bm{\mathcal{T}}_t) \le \bar{g}(t)$, $\forall t \in [0,1]$, the choice of $\bar{g}(\cdot)$ and the evolution of $g(\cdot)$ are restricted as shown below.
	\begin{theorem}
		\label{theorem_equality_constraint}
		Given a PTZF $\bar{g}(\cdot)$ with prescribed time $T_{\text{pre}} = 1$ and initial value $\bar{g}(0) \ge g(\bm{\mathcal{T}}_0)$, and a chosen class-$\mathcal{K}$ function $\gamma(\cdot): \mathbb{R}_{\ge 0} \to \mathbb{R}_{\ge 0}$, such that
		\begin{align} \label{eqn_system_g}
			\dot{g}(\bm{\mathcal{T}}_t) \le \gamma(\bar{g}(t) - g(\bm{\mathcal{T}}_t)) + \dot{\bar{g}}(t)
		\end{align}
		holds for all $t \in [0, T_{\text{pre}}]$.
		Then, it has $g(\bm{\mathcal{T}}_1) = 0$, indicating the satisfaction of $g_i(\bm{\mathcal{T}}_1) = 0$ for all $i = 1, \cdots, N_g$.
	\end{theorem}
	\begin{IEEEproof}
		See Appendix \ref{proof_theorem_equality_constraint}.
	\end{IEEEproof}
	\cref{theorem_equality_constraint} provides the guidance to achieve constrained generation in \cref{definition_constrained_generation} with equality constraints $g_i(\cdot)$, which requires the adjustment of flow $\dot{\bm{\mathcal{T}}}_t$ in \eqref{eqn_flow_matching_process} considering \eqref{eqn_g_dot_T} and \eqref{eqn_system_g}.
	In the next subsection, the inequality constraints in \cref{definition_constrained_generation} are considered.

	\subsection{Inequality Constraint}
	\label{subsection_inequality_constraint}
	Unlike the equality constraints $g_i(\cdot) = 0$ with $i = 1, \cdots, N_g$ which can be merged to a single function $g(\cdot)$, the inequality constraints $h_j(\cdot) \le 0$ require to be separately handled for $j = 1, \cdots, N_h$.
	As discussed in \cref{subsection_equality_constraint}, the inequality constraints may be violated at the initial sample $\bm{\mathcal{T}}_0$ indicating $h_j(\bm{\mathcal{T}}_0) > 0$, but should be enforced when the generation process ends at $t = 1$ resulting in $h_j(\bm{\mathcal{T}}_0) \le 0$ for $j = 1, \cdots, N_h$.
	This requirement is formulated as a prescribed time safety problem defined below.
	\begin{definition} \label{definition_PT_safety}
		Given a prescribed time $T_{\text{pre}} \in \mathbb{R}_{>0}$, a scalar dynamical system ${h}(\cdot): \mathbb{R}^d \to \mathbb{R}$ achieves prescribed time safety, if for arbitrary initial condition ${h}(\boldsymbol{\mathcal{T}}_0) \in \mathbb{R}$ it holds 
		\begin{align}
			{h}(\boldsymbol{\mathcal{T}}_t) \ge 0, && \forall t \in [T_{\text{pre}}, \infty),
		\end{align}
		where dynamics denotes
		\begin{align} \label{eqn_minus_hj_dot_T}
			\dot{h}(\bm{\mathcal{T}}_t) =  ( \mathrm{d} h(\bm{\mathcal{T}}_t) / \mathrm{d} \bm{\mathcal{T}}_t )^{\top} \dot{\bm{\mathcal{T}}}_t.
		\end{align}
	\end{definition}
	According to \cref{definition_PT_safety}, the enforced process for inequality constraints is prescribed time safety tasks w.r.t $- h_j(\bm{\mathcal{T}}_t)$ and $T_{\text{pre}} = 1$ for $j = 1, \cdots, N_h$.
	Specifically, for each $h_j(\cdot)$, define a PTZF $\bar{h}_j(\cdot)$ with prescribed time $T_{\text{pre}} = 1$, then the performance is shown in the following theorem.
	
	\begin{theorem}
		\label{theorem_inequality_constraint}
		For any $j = 1, \cdots, N_h$, given a PTZF $\bar{h}_j(\cdot)$ with prescribed time $T_{\text{pre}} = 1$ and initial value $\bar{h}_j(0) \ge h_j(\bm{\mathcal{T}}_0)$.
		If the dynamics for $h_j(\cdot)$ satisfies
		\begin{align} \label{eqn_system_hj}
			\dot{h}_j(\bm{\mathcal{T}}_t) \le \gamma(\bar{h}_j(t) - h_j(\bm{\mathcal{T}}_t)) + \dot{\bar{h}}_j(t)
		\end{align}
		for all $t \in [0, T_{\text{pre}}]$ with a class-$\mathcal{K}$ function $\gamma(\cdot): \mathbb{R}_{\ge 0} \to \mathbb{R}_{\ge 0}$, it is ensured that $- h_j(\bm{\mathcal{T}}_t)$ is prescribed time safety leading to the satisfaction of $h_j(\bm{\mathcal{T}}_1) \le 0$.
	\end{theorem}
	\begin{IEEEproof}
		See Appendix \ref{proof_theorem_inequality_constraint}.
	\end{IEEEproof}
	
	\cref{theorem_inequality_constraint} ensures that all inequality constraints in \eqref{eqn_constrained_generation} are satisfied, if $h_j(\cdot)$ follows \eqref{eqn_system_hj} for all $j = 1, \cdots, N_h$.
	While the condition \eqref{eqn_system_hj} shares the same structural form as the equality constraint dynamics in \eqref{eqn_system_g}, a fundamental difference lies in their behaviors.
	Specifically, due to the absence of a zero level lower bound as in $g(\cdot)$, the value of $h_j(\cdot)$ enters the negative real domain $\mathbb{R}_{\le 0}$ instead of converging to the origin. 
	Combining the results in \cref{theorem_equality_constraint} for equality constraints, the unified constrained generation framework with flow matching is proposed in the next subsection.
	
	\subsection{UniConFlow with Quadratic Programming}
	\label{subsection_UniConGen}
	
	To ensure that $\bm{\mathcal{T}}_1$ is a certified sample satisfying both equality and inequality constraints in \eqref{eqn_constrained_generation}, a guidance input $\bm{u}_t \in \mathbb{R}^d$ is introduced in the flow matching process \eqref{eqn_flow_matching_process}, such that the guided flow is written as
	\begin{align}
		\label{eqn_controlled_flow}
		\dot{\bm{\mathcal{T}}}_t = \bm{v}_t^{\bm{\theta}}(\bm{\mathcal{T}}_t) + \bm{u}_t
	\end{align}
	for all $t \in [0,1]$.
	Additionally, combining the dynamics \eqref{eqn_g_dot_T} and \eqref{eqn_minus_hj_dot_T}, the conditions \eqref{eqn_system_g} in \cref{theorem_equality_constraint} and \eqref{eqn_system_hj} in \cref{theorem_inequality_constraint} are rewritten as
	\begin{align} \label{eqn_QP_constraints}
		\rho_g(t) + \bm{\eta}_g^{\top}(t) \bm{u}_t &\le 0, \\
		\rho_{h,j}(t) + \bm{\eta}_{h,j}^{\top}(t) \bm{u}_t &\le 0, && \forall j = 1, \cdots, N_h,
	\end{align} 
	respectively, where $\rho_g(\cdot), \rho_{h,j}(\cdot): [0,1] \to \mathbb{R}$ are defined as
	\begin{align}
		\rho_g(t) &= \bm{\eta}_g^{\top}(t) \bm{v}_t^{\bm{\theta}}(\bm{\mathcal{T}}_t) - \gamma(\bar{g}(t) - g(\bm{\mathcal{T}}_t)) - \dot{\bar{g}}(t), \\
		\rho_{h,j}(t) &= \bm{\eta}_{h,j}^{\top}(t) \bm{v}_t^{\bm{\theta}}(\bm{\mathcal{T}}_t) - \gamma(\bar{h}_j(t) - h_j(\bm{\mathcal{T}}_t)) - \dot{\bar{h}}_j(t)
	\end{align}
	for $j = 1, \cdots, N_h$.
	Moreover, the functions $\bm{\eta}_g(\cdot), \bm{\eta}_{h,j}(\cdot): [0,1] \to \mathbb{R}^d$ for $j = 1, \cdots, N_h$ are defined as
	\begin{align}
		\bm{\eta}_g(t) = \mathrm{d} g(\bm{\mathcal{T}}_t) / \mathrm{d} \bm{\mathcal{T}}_t, &&
		\bm{\eta}_{h,j}(t) = \mathrm{d} h_j(\bm{\mathcal{T}}_t) / \mathrm{d} \bm{\mathcal{T}}_t
	\end{align}
	for $j = 1, \cdots, N_h$.
	The value of input $\bm{u}_t$ is designed to minimize the modification of original flow process in \eqref{eqn_flow_matching_process}, while ensuring the satisfaction of \eqref{eqn_QP_constraints}.
	Specifically, $\bm{u}_t$ obtained by solving quadratic optimization (QP) problem as
	\begin{subequations}
		\label{eqn_constrained_optimization}
		\begin{align} 
			&\min\nolimits_{\bm{u}_t \in \mathbb{R}^d} \bm{u}_t^{\top} \bm{P}_{u,t} \bm{u}_t \\
			\text{s.t.~} &\bm{\rho}(t) + \bm{\eta}(t) \bm{u}_t \preceq \bm{0}_{(N_h + 1) \times 1}
		\end{align}
		where $\bm{P}_{u,t} \in \mathbb{R}^{d \times d}$ is a positive definite matrix and
		\begin{align}
			&\bm{\rho}(t) = [\rho_g(t), \rho_{h,1}(t), \cdots, \rho_{h,N_h}(t)]^{\top}, \\
			&\bm{\eta}(t) = [\bm{\eta}_g(t), \bm{\eta}_{h,1}(t), \cdots, \bm{\eta}_{h,N_h}(t)]^{\top}.
		\end{align}
	\end{subequations}
	The operator $\preceq$ represents element-wise comparison, i.e., $\bm{\rho}_1 \preceq \bm{\rho}_2$ if and only if $\rho_{1,j} \leq \rho_{2,j}$ for all $j = 1, \cdots, (N_h + 1)$ with $\bm{\rho}_i = [\rho_{i,1}, \cdots, \rho_{i,N_h + 1}] \in \mathbb{R}^{N_h + 1}$ and $i = 1, 2$.
	If \eqref{eqn_constrained_optimization} is always feasible for all $t \in [0,1]$, the generated sample $\bm{\mathcal{T}}_1$ satisfies all constraints in \eqref{eqn_constrained_generation} according to \cref{theorem_equality_constraint,theorem_inequality_constraint}, which is summarized as follows.
	
	\begin{theorem}
		Consider \eqref{eqn_constrained_optimization} is feasible for all $t \in [0,1]$, and apply $\bm{u}_t$ from \eqref{eqn_controlled_flow}.
		Then, the generated sample $\bm{\mathcal{T}}_1$ from \eqref{eqn_controlled_flow} satisfies all constraints in \eqref{eqn_constrained_generation}.
	\end{theorem}
	\begin{IEEEproof}
		The proof is omitted as the result follows immediately from Theorems~\ref{theorem_equality_constraint} and~\ref{theorem_inequality_constraint}.
	\end{IEEEproof}
	
	However, the constraints in \eqref{eqn_constrained_optimization} may contradict with each other, resulting in potential infeasibility.
	To ensure the feasibility of optimization problem \eqref{eqn_constrained_optimization}, a slack variable $\bm{\delta}_t \in \mathbb{R}^{N_h + 1}$ is introduced, such that the optimization problem becomes
	\begin{subequations}
		\begin{align}
			&\min\nolimits_{\bm{u}_t \in \mathbb{R}^d, \bm{\delta}_t \in \mathbb{R}^{N_h + 1}} \bm{u}_t^{\top} \bm{P}_{u,t} \bm{u}_t + \bm{\delta}_t^{\top} \bm{P}_{\delta,t} \bm{\delta}_t \\
			\text{s.t.~} &\bm{\rho}(t) + \bm{\eta}(t) \bm{u}_t = \bm{\delta}_t,
		\end{align}
	\end{subequations}
	which is also written by defining $\bm{z}_t = [\bm{u}_t^{\top}, \bm{\delta}_t^{\top}]^{\top}$ as
	\begin{subequations}
		\label{eqn_QP}
		\begin{align} 
			&\min\nolimits_{\bm{z}_t \in \mathbb{R}^{d + N_h + 1}} \bm{z}_t^{\top} \bm{P}_{z,t} \bm{z}_t \\
			\text{s.t.~} &\bm{\rho}(t) + \bm{\eta}_z(t) \bm{z}_t = \bm{0}_{(N_h + 1) \times 1},
		\end{align}
	\end{subequations}
	with $\bm{P}_{z,t} = \mathrm{blkdiag}(\bm{P}_{u,t}, \bm{P}_{\delta,t})$ and $\bm{\eta}_z(t) = [\bm{\eta}(t), - \bm{I}_{N_h + 1}]$.
	It is easy to see \eqref{eqn_QP} is a quadratic programming problem, whose feasibility is always guaranteed due to the introduced slack variables $\bm{\delta}_t$.
	Specifically, the closed-form solution of \eqref{eqn_QP} with slack variables $\bm{\delta}_t$ is written as
	\begin{align} \label{eqn_QP_closed_form_solution}
		\bm{z}_t \!\!=\!\! \begin{cases}
			- (\bm{P}_{z,t}^{\top})^{-1} \bm{\eta}_z^{\top}(t) \bm{\Lambda}^{-1}(t) \bm{\rho}(t), & \bm{\rho}(t) \!\!\succ\!\! \bm{0}_{(N_h + 1) \times 1} \\
			\bm{0}_{(d + N_h + 1) \times 1}, & \text{otherwise}
		\end{cases}
		,\!
	\end{align}
	with $\bm{\Lambda}(t) = \bm{\eta}_z^{\top}(t) \bm{P}_{z,t}^{-1} \bm{\eta}_z(t)$ following Karush–Kuhn–Tucker (KKT) conditions and resulting in the value of input in \eqref{eqn_controlled_flow} as $\bm{u}_t = [\bm{I}_d, \bm{0}_{d \times (N_h + 1)}] \bm{z}_t$.
	The certification of the generated sample $\bm{\mathcal{T}}_1$ from the controlled flow process \eqref{eqn_controlled_flow} with $\bm{u}_t$ from \eqref{eqn_QP_closed_form_solution} is shown in the following theorem.
	
	\begin{theorem}
		\label{theorem_ut_slack_variable}
		Suppose that there exists a well-defined constant $c_{\gamma} \in \mathbb{R}_{>0}$ and a blow-up function $\varphi(\cdot): [0,1) \to \mathbb{R}_{>0}$ satisfying $\lim_{t \to 1^-} \varphi(t) = \infty$ and $\int_0^{1^-} \varphi(s) \mathrm{d} s = \infty$, such that the class-$\mathcal{K}$ function $\gamma(\cdot)$ used in \eqref{eqn_system_g} and \eqref{eqn_system_hj} satisfies $\gamma(a(t)) \le c_{\gamma} \varphi(t) a(t)$ for all possible function $a(\cdot): [0,1) \to \mathbb{R}$.
		Let the slack variable $\bm{\delta}_t = [\delta_{g,t}, \delta_{h,1,t}, \cdots, \delta_{h, N_h, t}]^{\top}$ obtained from \eqref{eqn_QP_closed_form_solution} as $\bm{\delta}_t = [\bm{0}_{(N_h + 1) \times d}, \bm{I}_{N_h + 1}] \bm{z}_t$, and assume $\bm{\delta}_t$ is bounded by well-defined constants $\bar{\delta}_g, \bar{\delta}_{h,j} \in \mathbb{R}_{>0}$ for all $j = 1, \cdots, N_h$, i.e., $\delta_{g,t} \le \bar{\delta}_g$ and $\delta_{h,j,t} \le \bar{\delta}_{h,j}$ for all $t \in [0,1]$.
		Apply the closed form solution of $\bm{u}_t$ from \eqref{eqn_QP_closed_form_solution} into \eqref{eqn_controlled_flow}, then the generated sample $\bm{\mathcal{T}} = \bm{\mathcal{T}}_1$ satisfies all constraints in \eqref{eqn_constrained_generation}.
	\end{theorem}
	\begin{IEEEproof}
		See Appendix \ref{proof_theorem_ut_slack_variable}.
	\end{IEEEproof}
	
	\cref{theorem_ut_slack_variable} shows that the certified sample $\bm{\mathcal{T}}_1$ in \cref{definition_constrained_generation} is always ensured, resulting from the guaranteed feasibility of \eqref{eqn_QP} with the slack variable.
	However, an additional requirement on $\gamma(\cdot)$ is necessary with the blow-up function $\varphi(\cdot)$, which is similar to some existing prescribed time approaches.
	
	\subsection{Comparison to Conventional Prescribed Time Approaches}
	\label{subsection_comparison}
	
	In \cref{subsection_equality_constraint,subsection_inequality_constraint}, the enforced processes for equality and inequality constraints are regarded as prescribed time stability and safety problems in \cref{definition_PT_stability,definition_PT_safety}, respectively.
	In the case of an equality constraint, the condition of prescribed time stability is 
	\begin{align} \label{eqn_system_g_conventional_PT}
		\dot{g}(\bm{\mathcal{T}}_t) \le - c_{\text{PT}} \varphi_{\text{PT}}(t) g(\bm{\mathcal{T}}_t)
	\end{align}
	with $c_{\text{PT}} \in \mathbb{R}_{>0}$.
	The function $\varphi_{\text{PT}}(\cdot): [0,1) \to \mathbb{R}_{>0}$ is a monotonically increasing blow-up function satisfying $\lim_{t \to 1^-} \varphi_{\text{PT}}(t) = \infty$ and $\int_0^{1^-} \varphi_{\text{PT}}(s) \mathrm{d} s = \infty$.
	It is obvious that the existing methods with \eqref{eqn_system_g_conventional_PT} are slightly different from \eqref{eqn_system_g} proposed in this work, which is designed specifically for generative models considering the behavior in the early generation stage.
	
	For a fair comparison, the class-$\mathcal{K}$ function $\gamma(\cdot)$ in \eqref{eqn_system_g} is chosen as $\gamma(a) = c_{\text{PT}} \varphi_{\text{PT}}(t) a$ with $\varphi_{\text{PT}}(t) = 1 / (1 - t)^2$, where the blow-up term $\varphi(\cdot)$ is added for feasibility as shown in \cref{theorem_ut_slack_variable}.
	Due to the usage of PTZF $\bar{g}$, its dynamics is also specified, such that the function $\gamma_r(t,a)$ in \eqref{eqn_system_r} is designed as $\gamma_r(t,a) = c_g a$ with $c_g < c_{\text{PT}}$.	
	Take the first part of the entire generation time as the early stage, i.e., $[0, c_r]$ with $c_r$ defined in \cref{definition_PTZF}.
	For the conventional condition \eqref{eqn_system_g_conventional_PT} for prescribed time stability, it is obvious to see the demand of a negative upper bound for $\dot{g}(\bm{\mathcal{T}}_t)$.
	Moreover, the value of $g(\bm{\mathcal{T}}_t)$ determined the restriction level for its derivative, indicating large modification with big $\bm{u}_t$ is required at the early generation stage with $t \!\in\! [0, c_r]$.
	In contrast, the proposed condition in \eqref{eqn_system_g} with PTZF allows a positive upper bound for $\dot{g}(\bm{\mathcal{T}}_t)$.
	Specifically, due to the choice of $T_{\text{pre}} \!=\! 1$ and $\gamma_r(t,a) \!=\! c_g a$, the dynamics of introduced PTZF $\bar{g}(\cdot)$ becomes
	\begin{align}
		\dot{\bar{g}}(t) = - c_g \bar{g}(t) / (1 - t)^2
	\end{align}
	whose solution is explicitly calculated as
	\begin{align} \label{eqn_bar_g_t}
		\bar{g}(t) = \mathrm{e}^{- c_g \int_0^t (1 - s)^{-2} \mathrm{d} s} \bar{g}(0) = \mathrm{e}^{- \frac{c_g}{1 - t}} \mathrm{e}^{c_g} \bar{g}(0)
	\end{align}
	for all $t \in [0,1]$.
	Considering $t \in [0,c_r]$, it has 
	\begin{align}
		\bar{g}(t) \ge \bar{g}(c_r) = \mathrm{e}^{- c_g c_r / (1 - c_r)} \bar{g}(0).
	\end{align}
	Substitute $\dot{\bar{g}}(t)$ and the upper bound of $\bar{g}(t)$ for $t \in [0,c_r]$ into the right-hand side of the condition \eqref{eqn_system_g} for equality constraint, then it is lower bounded by
	\begin{align} \label{eqn_RHS_g_bound}
		\gamma(\bar{g}(t) - g(\bm{\mathcal{T}}_t)) &+ \dot{\bar{g}}(t) = - \frac{c_{\text{PT}}}{(1 - t)^2}g(\bm{\mathcal{T}}_t) + \frac{c_{\text{PT}} - c_g}{(1 - t)^2} \bar{g}(t) \nonumber \\
		\ge& - \frac{c_{\text{PT}} g(\bm{\mathcal{T}}_t)}{(1 - c_r)^2} + \frac{(c_{\text{PT}} - c_g) \bar{g}(0)}{\mathrm{e}^{c_g c_r / (1 - c_r)}},
	\end{align}
	which can be positive due to the existence $\bar{g}(0)$.
	Furthermore, free generation with $\bm{u}_t = \bm{0}_{d \times 1}$ can be achieved by choosing sufficiently large $\bar{g}(0)$ at the early stage for generation, which is shown as follows. 
	\begin{theorem} \label{theorem_free_generation_condition}
		Suppose the equality and inequality constraints $g(\cdot)$ and $h_j(\cdot)$ for $j = 1, \cdots, N_h$ are Lipschitz continuous with Lipschitz constants $L_g, L_{h,j} \in \mathbb{R}_{,+}$ respectively, i.e., $\| \bm{\eta}_g(t) \| \le L_g$ and $\| \bm{\eta}_{h,j}(t) \| \le L_{h,j}$ for all $t \in [0,1]$.
		Moreover, assume the unconstrained flow $\bm{v}_t^{\bm{\theta}}(\cdot)$ is bounded by well-defined constant $\bar{v}^{\bm{\theta}} \in \mathbb{R}_{>0}$, i.e., $\| \bm{v}_t^{\bm{\theta}}(\bm{\mathcal{T}}) \| \le \bar{v}^{\bm{\theta}}$ for any $t \in [0,1]$ and $\bm{\mathcal{T}} \in \mathbb{R}^d$.
		Choose the initial value $\bar{g}(0), \bar{h}_j(0) \in \mathbb{R}$ for PTZF $\bar{g}(\cdot)$ and $\bar{h}_j(\cdot)$ satisfying
		\begin{align} \label{eqn_g_h_bar_initial_free_generation}
			\bar{g}(0) \ge& \xi_{\mathcal{T},0} g(\bm{\mathcal{T}}_0) + \xi_{\mathcal{T},\partial} L_g \bar{v}^{\bm{\theta}}, \\
			\bar{h}_j(0) \ge& \xi_{\mathcal{T},0} h_j(\bm{\mathcal{T}}_0) + \xi_{\mathcal{T},\partial} L_{h,j} \bar{v}^{\bm{\theta}}
		\end{align}
		respectively with the coefficients $\xi_{\mathcal{T},0}, \xi_{\mathcal{T},\partial} \in \mathbb{R}_{>0}$ written as
		\begin{align} \label{eqn_xi}
			&\xi_{\mathcal{T},0} = \frac{c_{\text{PT}}}{c_{\text{PT}} - c_g} \frac{\mathrm{e}^{c_g c_r / (1 - c_r)}}{(1 - c_r)^2}, \\
			&\xi_{\mathcal{T},\partial} = \frac{\mathrm{e}^{c_g c_r / (1 - c_r)}}{c_{\text{PT}} - c_g} \Big (1 + \frac{c_{\text{PT}} c_r}{(1 - c_r)^2} \Big),
		\end{align}
		then the optimal input denotes $\bm{u}_t = \bm{0}_{d \times 1}$, $\forall t \in [0, c_r]$.
	\end{theorem}
	\begin{IEEEproof}
		See Appendix \ref{proof_theorem_free_generation_condition}.
	\end{IEEEproof}
	\cref{theorem_free_generation_condition} shows the possibility of free generation using the proposed prescribed time approaches with PTZF, which is impossible for any other existing works.
	Note that the proposed conditions \eqref{eqn_system_g} and \eqref{eqn_system_hj} for prescribed time stability and safety can degrade to conventional methods, by letting $\gamma_r(t, \cdot)$ be $0$ for all $t \in [0,1]$.
	Therefore, the proposed PTZF-based prescribed time approaches are more general, allowing early free generation and evolution.
	
\begin{figure}[t]
		\centering
		\includegraphics[width=1\linewidth]{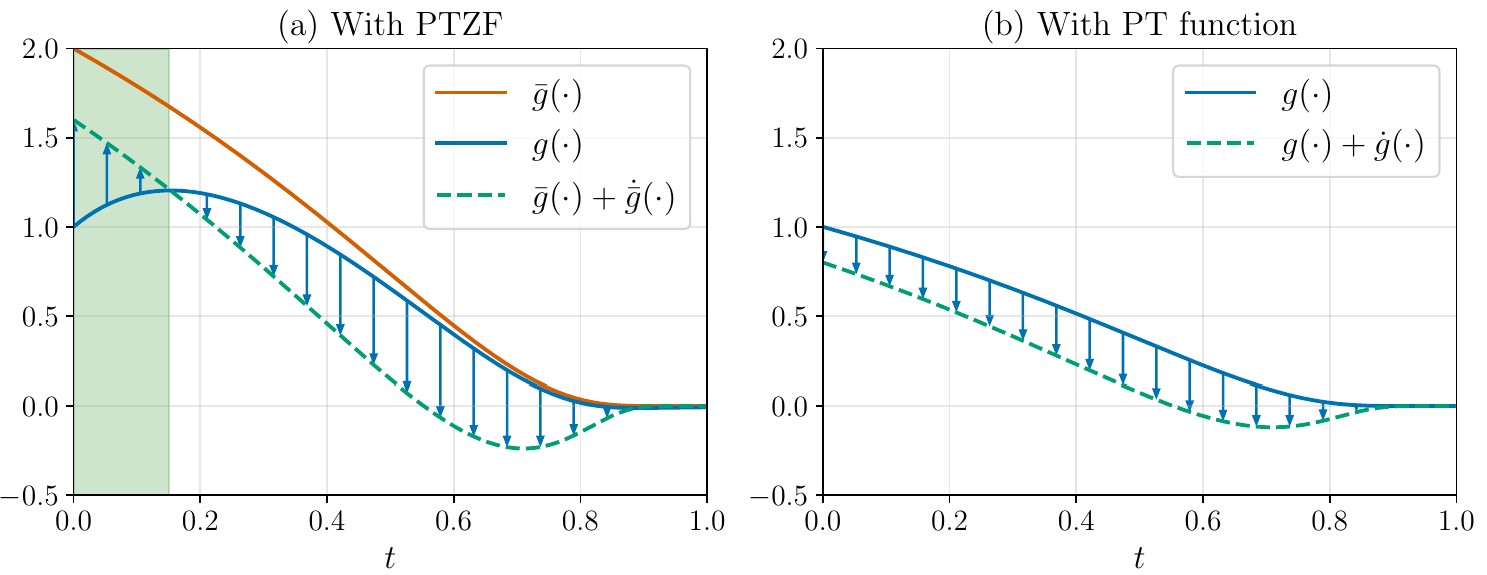}
		\caption{Function $g(t)$ over time with $c_r = 0$, $\gamma_r(t,r) = \underline{\gamma}_r = r$ in \cref{definition_PTZF} and $g(0) = 1$, $\bar{g}(0) = 2$ in \cref{theorem_equality_constraint} with PTZF (a) and with PT function (b). Green area denotes the free generation process.}
		\vspace{-0.3cm}
		\label{fig_gFunction}
	\end{figure}

	A numerical example is shown in \cref{fig_gFunction} for the equality constraint, illustrating the potential free generation phase using the proposed PTZF.

	\begin{remark}
		Considering the expression of $\xi_{\mathcal{T},0}$ in \eqref{eqn_xi}, when $\xi_{\mathcal{T},0} > 1$, it indicates $\bar{g}(0) > g(\bm{\mathcal{T}}_0)$ and $\bar{h}_j(0) > h_j(\bm{\mathcal{T}}_0)$, $\forall j = 1, \cdots, N_h$.
		This observation aligns with the requirement in \cref{theorem_equality_constraint,theorem_inequality_constraint}.
		Moreover, to ensure long free generation time with big $c_r$ (see the green area in \cref{fig_gFunction}), large initial values $\bar{g}(0)$ and $\bar{h}_j(0)$ are essential, which also increases the guidance strength in the later stage $(c_r, 1]$.
		In detail, exemplarily considering the equality constraint condition \eqref{eqn_system_g} with the worst case with $\bar{g}(t) = g(\bm{\mathcal{T}}_t)$, its time derivative is
		\begin{align}
			\dot{g}(\bm{\mathcal{T}}_t) \le \dot{\bar{g}}(t) = - c_g \mathrm{e}^{- \frac{c_g}{1 - t}} \mathrm{e}^{c_g} \bar{g}(0) / (1 - t)^2
		\end{align}
		according to \eqref{eqn_bar_g_t}, whose absolute value is monotonically increasing w.r.t $\bar{g}(0)$ for any fixed time $t \in [0,1)$.
		Therefore, there exists a trade-off between free generation and guidance strength. 
		To quantify this trade-off phenomenon is considered as future work.
	\end{remark}
	
	\begin{remark}
		\cref{theorem_free_generation_condition} shows the free generation under Lipschitz continuous constraint functions, including $\bar{g}(\cdot)$.
		Considering the expression of $g(\cdot)$ in \eqref{eqn_g}, its derivative is written as
		\begin{align}
			\frac{\mathrm{d} g(\bm{\mathcal{T}})}{\mathrm{d} \bm{\mathcal{T}}} = 2 \sum\nolimits_{i = 1}^{N_g} g_i(\bm{\mathcal{T}}) \frac{\mathrm{d} g_i(\bm{\mathcal{T}})}{\mathrm{d} \bm{\mathcal{T}}},
		\end{align}
		whose boundedness requires bounded and Lipschitz individual constraints $g_i(\cdot)$ for $i = 1, \cdots, N_g$.
		This condition is automatically fulfilled when $\bm{\mathcal{T}}$ is sampled from a bounded distribution.
		While theoretically the initial distribution $p$ is usually set as Gaussian leading to unbounded $\bm{\mathcal{T}}_0$, the practical implementation can filter the outlier sample out, ensuring bounded $\bm{\mathcal{T}}_0$.
		Then, the boundedness of $\bm{\mathcal{T}}_t$ naturally holds due to bounded $\dot{g}(\bm{\mathcal{T}}_t)$ from \eqref{eqn_dot_g_upper_bound} for all $t \in [0,1]$.
		Furthermore, combining the continuity of $g(\cdot)$ and bounded domain for $\bm{\mathcal{T}}$, the boundedness and Lipschitz continuity of $g_i(\cdot)$ and therefore $g(\cdot)$ are straightforward.
	\end{remark}
	\section{Certified Robot Motion Planning Strategy}
	\label{section_consistent_safe_generation}
	
	In this section, the certified robot motion planning in \cref{problem_certified_FM_for_trajectory} is investigated as an example for constrained generation task.
	Specifically, the requirements of dynamical consistency, safety, and feasibility are numerically modeled as different types of constraints in \cref{subsection_motion_planning_constraint} from the trajectory view.
	Next, the certified motion planning strategy is discussed in \cref{subsection_certified_motion_planning}, building the explicit connection between certified motion planning and constrained generation task\footnote{The certified path planning is also discussed in Appendix~\ref{subsection_certified_path_planning} due to page limitations.}.
	
	\subsection{Certified Motion Planning as Constrained Generation}
	\label{subsection_motion_planning_constraint}
	In this subsection, the requirements for a certified trajectory $\bm{\mathcal{T}}$ in \cref{definition_certified_robot_trajectory} is mathematically formulated to obtain a similar expression to constrained generation in \cref{definition_constrained_generation}.
	
	\subsubsection{Dynamical Consistency as Equality Constraints}
	\label{subsubsection_dynamical_consistency}
	
	Recalling the requirement of dynamical constraint in \eqref{eqn_dynamics_consistency}, it is easy to reformulate it into equality constraints as
	\begin{align}
		\label{eq_gk_con}
		g^k_{\text{con}}(\bm{\mathcal{T}}) =& \| \bm{s}^{k+1} - \bm{f}^k(\bm{s}^k, \bm{a}^k) \|
	\end{align}
	with $k = 0, \cdots, H - 1$.
	According to the construction of $g(\cdot)$ in \eqref{eqn_g} and \cref{theorem_equality_constraint}, the derivative of the squared form for each equality constraint is essential, which for dynamical consistency is written as
	\begin{align} \label{eqn_derivative_dynamical_consistency_constraint}
		\mathrm{d} &(g^k_{\text{con}}(\bm{\mathcal{T}}))^2 / \mathrm{d} t  = 2 (\bm{s}^{k+1} - \bm{f}^k(\bm{s}^k, \bm{a}^k))^T \\
		&\times \Big( \dot{\bm{s}}^{k+1} - \Big( \frac{\mathrm{d} \bm{f}^k(\bm{s}^k, \bm{a}^k)}{\mathrm{d} \bm{s}^k} \Big)^T \dot{\bm{s}}^k - \Big( \frac{\mathrm{d} \bm{f}^k(\bm{s}^k, \bm{a}^k)}{\mathrm{d} \bm{a}^k} \Big)^T \dot{\bm{a}}^k \Big). \nonumber
	\end{align}
	Moreover, introduce the selection matrices $\bm{E}_s^k \in \mathbb{R}^{d_s \times d}$ for system state with $k = 0, \cdots, H$ and $\bm{E}_a^k \in \mathbb{R}^{d_a \times d}$ for action with $k = 0, \cdots, H - 1$ as
	\begin{align} \label{eqn_election_matrix}
		&\bm{E}_s^k = [\bm{0}_{d_s \times (d_s + d_a) k}, \bm{I}_{d_s}, \bm{0}_{d_s \times (d_s + d_a) (H - k)}], \\
		&\bm{E}_a^k = [\bm{0}_{d_a \times (d_s (k+1) + d_a k)}, \bm{I}_{d_a}, \bm{0}_{d_a \times (d_s + d_a) (H -k -1)}],
	\end{align}
	such that $\bm{s}^k = \bm{E}_s^k \bm{\mathcal{T}}$ and $\bm{a}^k = \bm{E}_a^k \bm{\mathcal{T}}$.
	Then, the derivative in \eqref{eqn_derivative_dynamical_consistency_constraint} is further expressed as
	\begin{align} \label{eqn_dot_g_dynamical_consistency}
		\mathrm{d} (g^k_{\text{con}}(\bm{\mathcal{T}}))^2 / \mathrm{d} t = (\bm{\eta}^k_{\text{con}}(\bm{\mathcal{T}}))^T \dot{\bm{\mathcal{T}}}
	\end{align}
	where $\bm{\eta}^k_{\text{con}}(\cdot): \mathbb{R}^{d} \to \mathbb{R}^{d}$ is written as
	\begin{align}
		\bm{\eta}^k_{\text{con}}(\bm{\mathcal{T}}) \!=& 2 \!\Big(\! \bm{E}_s^{k+1} \!\!-\!\! \Big(\! \frac{\mathrm{d} \bm{f}^k(\bm{s}^k, \bm{a}^k)}{\mathrm{d} \bm{s}^k} \!\Big)^T \! \bm{E}_s^k \!\!-\!\! \Big(\! \frac{\mathrm{d} \bm{f}^k(\bm{s}^k, \bm{a}^k)}{\mathrm{d} \bm{a}^k} \!\Big)^T \! \bm{E}_a^k \!\Big)^T \nonumber \\
		&\times (\bm{s}^{k+1} - \bm{f}^k(\bm{s}^k, \bm{a}^k))
	\end{align}
	for $k = 0, \cdots, H - 1$.
	
	\subsubsection{State Constraint}
	\label{subsubsection_state_constraint}
	
	To ensure all generated states $\bm{s}^k$ are in the given safe domains $\mathbb{C}_s$ for all $k = 0, \cdots, H$ as in \eqref{eqn_state_domain}, we first mathematically define the domains $\mathbb{C}_s$ using the concept in control barrier functions.
	Specifically, following the definition of zeroing CBF, a differentiable continuous function $h_s(\cdot): \mathbb{S} \to \mathbb{R}$ is adopted to describe $\mathbb{C}_s$, such that
	\begin{align}
		\mathbb{C}_s \subseteq& \{ \bm{s} \in \mathbb{S} : h_s(\bm{s}) \ge 0 \}
	\end{align}
	and $\partial \mathbb{C}_s = \{ \bm{s} \in \mathbb{S} : h_s(\bm{s}) = 0 \}$.
	Then, the safety requirement $\bm{s}^k \in \mathbb{C}_s$ for all $k = 0, \cdots, H$ is tightened as
	\begin{align}
		h_s^k(\bm{\mathcal{T}}) = - h_s(\bm{s}^k) \le 0,
	\end{align}
	whose derivative used in \cref{theorem_inequality_constraint} is written as
	\begin{align} \label{eqn_dot_h_state_constraint}
		\dot{h}_s^k(\bm{\mathcal{T}}) = - ( \mathrm{d} h_s(\bm{s}^k) / \mathrm{d} \bm{s}^k )^T \dot{\bm{s}}^k = (\bm{\eta}^k_s(\bm{\mathcal{T}}))^T \dot{\bm{\mathcal{T}}}
	\end{align}
	with $\bm{\eta}^k_s(\bm{\mathcal{T}}) \!\!=\!\! - (\bm{E}_s^k)^T (\mathrm{d} h_s(\bm{s}^k) / \mathrm{d} \bm{s}^k)$ and the notation in \eqref{eqn_election_matrix}.
	
	\subsubsection{Action Feasibility}
	\label{subsubsection_action_constraint}
	
	Similarly as state constraint, a differentiable continuous function $h_a(\cdot): \mathbb{A} \to \mathbb{R}$ is used to describe feasible action set $\mathbb{C}_a$ in \eqref{eqn_action_domain}, such that
	\begin{align}
		\mathbb{C}_a = \{ \bm{a} \in \mathbb{A} : h_a(\bm{a}) \ge 0 \}
	\end{align}
	with $\partial \mathbb{C}_a = \{ \bm{a} \in \mathbb{A} : h_a(\bm{a}) = 0 \}$ for all $k = 0, \cdots, H-1$.
	Then, the feasibility requirement $\bm{a}^k \in \mathbb{C}_a$ becomes
	\begin{align}
		h_a^k(\bm{\mathcal{T}}) = - h_a(\bm{a}^k) \le 0
	\end{align}
	for $k = 0, \cdots, H-1$, whose derivative is written as
	\begin{align} \label{eqn_dot_h_action_constraint}
		\dot{h}_a^k(\bm{\mathcal{T}}) =& - ( \mathrm{d} h_a(\bm{a}^k) / \mathrm{d} \bm{a}^k )^T \dot{\bm{a}}^k = (\bm{\eta}^k_a(\bm{\mathcal{T}}))^T \dot{\bm{\mathcal{T}}}
	\end{align}
	with $\bm{\eta}^k_a(\bm{\mathcal{T}}) = - (\bm{E}_a^k)^T (\mathrm{d} h_a(\bm{a}^k) / \mathrm{d} \bm{a}^k)$.
	
	\subsubsection{Initial Condition Alignment}
	\label{subsubsection_initial_condition}
	
	Usually, the motion planning task asks to generate a trajectory starting from a safe current state $\bm{s}_{\text{cur}} \in \mathbb{C}_s$, leading to the requirement $\bm{s}^0 = \bm{s}_{\text{cur}}$.
	This demand is also formulated as an equality constraint as
	\begin{align}
		g_{\text{ini}}(\bm{\mathcal{T}}) = \| \bm{s}^0 - \bm{s}_{\text{cur}} \|,
	\end{align}
	and the derivative of its squared form is written as
	\begin{align} \label{eqn_dot_g_initial_condition}
		\mathrm{d} (g_{\text{ini}}(\bm{\mathcal{T}}))^2 / \mathrm{d} t  = 2 (\bm{s}^0 - \bm{s}_{\text{cur}})^T \dot{\bm{s}}^0 = \bm{\eta}_{\text{ini}}^T(\bm{\mathcal{T}}) \dot{\bm{\mathcal{T}}} 
	\end{align}
	with $\bm{\eta}_{\text{ini}}(\bm{\mathcal{T}}) = 2 (\bm{E}_s^0)^T (\bm{s}^0 - \bm{s}_{\text{cur}})$.
	
	\begin{remark}
		In conventional motion planning task, the initial condition serves as the guidance, which can be regarded as a soft constraint with potential violation.
		In contrast, the initial condition is considered as a hard equality constraint, leading to reliability of generated motion.
	\end{remark}
	
	This subsection formulate the requirements of certified motion planning in \cref{definition_certified_robot_trajectory} based on the trajectory $\bm{\mathcal{T}}$ in \eqref{eqn_T}.
	In the following subsections, the explicit expression of parameters in the optimization problem \eqref{eqn_constrained_optimization} for calculating the guidance term $\bm{u}_t$ in \eqref{eqn_controlled_flow} are shown.

	\subsection{Certified Trajectory Planning}
	\label{subsection_certified_motion_planning}
	
	Combining the requirements of dynamical consistency in \cref{subsubsection_dynamical_consistency} and initial condition in \cref{subsubsection_initial_condition}, the function $g(\cdot)$ in \eqref{eqn_g} for equality constraints is formulated as
	\begin{align} \label{eqn_dot_g_trajectory_planning}
		g(\bm{\mathcal{T}}_t) = g_{\text{ini}}^2(\bm{\mathcal{T}}_t) + \sum\nolimits_{k = 0}^{H - 1} (g^k_{\text{con}}(\bm{\mathcal{T}}_t))^2,
	\end{align}
	for $t \in [0,1]$.
	The derivative of $g(\bm{\mathcal{T}}_t)$ w.r.t $t$ is written as
	\begin{align} \label{eqn_g_trajectory_planning}
		\dot{g}(\bm{\mathcal{T}}_t) =& \big( \bm{\eta}_{\text{ini}}(\bm{\mathcal{T}}_t) + \sum\nolimits_{k = 0}^{H - 1} \bm{\eta}^k_{\text{con}}(\bm{\mathcal{T}}_t) \big)^T \dot{\bm{\mathcal{T}}}_t \\
		=& \bm{\eta}_g^T(\bm{\mathcal{T}}_t) \bm{v}_t^{\bm{\theta}}(\bm{\mathcal{T}}_t) + \bm{\eta}_g^T(\bm{\mathcal{T}}_t) \bm{u}_t \nonumber
	\end{align}
	following \eqref{eqn_dot_g_dynamical_consistency} and \eqref{eqn_dot_g_initial_condition} and using the controlled flow process in \eqref{eqn_controlled_flow}, where $\bm{\eta}_g(\cdot): \mathbb{R}^d \to \mathbb{R}^d$ denotes
	\begin{align}
		\bm{\eta}_g(\bm{\mathcal{T}}_t) =& \bm{\eta}_{\text{ini}}(\bm{\mathcal{T}}_t) + \sum\nolimits_{k = 0}^{H - 1} \bm{\eta}^k_{\text{con}}(\bm{\mathcal{T}}_t).
	\end{align}
	Next, the inequality constraints for system state $\bm{s}$ and action $\bm{a}$ in \cref{subsubsection_state_constraint} and \cref{subsubsection_action_constraint} are separately considered, which are reordered as
	\begin{align} \label{eqn_h_trajectory_planning}
		&h_j(\bm{\mathcal{T}}_t) = h_s^{j - 1}(\bm{\mathcal{T}}_t), && \forall j = 1, \cdots, H + 1, \\
		&h_j(\bm{\mathcal{T}}_t) = h_a^{j - H - 2}(\bm{\mathcal{T}}_t), && \forall j = H + 2, \cdots, 2 H + 1,
	\end{align}
	whose derivative following \eqref{eqn_dot_h_state_constraint} and \eqref{eqn_dot_h_action_constraint} is expressed as
	\begin{align} \label{eqn_dot_h_trajectory_planning}
		\dot{h}_j(\bm{\mathcal{T}}_t) = \bm{\eta}_{h,j}^T(\bm{\mathcal{T}}_t) \bm{v}_t^{\bm{\theta}}(\bm{\mathcal{T}}_t) + \bm{\eta}_{h,j}^T(\bm{\mathcal{T}}_t) \bm{u}_t
	\end{align}
	considering the controlled flow in \eqref{eqn_controlled_flow}.
	The vector function $\bm{\eta}_{h,j}(\cdot): \mathbb{R}^d \to \mathbb{R}^d$ is defined as
	\begin{align}
		&\bm{\eta}_{h,j}(\bm{\mathcal{T}}_t) = \bm{\eta}^{j - 1}_s(\bm{\mathcal{T}}_t), && \!\!\forall j = 1, \dots, H + 1, \\
		&\bm{\eta}_{h,j}(\bm{\mathcal{T}}_t) = \bm{\eta}^{j - H - 2}_a(\bm{\mathcal{T}}_t), && \!\!\forall j = H + 2, \dots, 2 H + 1.
	\end{align}
	With the redefined $g(\cdot)$ in \eqref{eqn_g_trajectory_planning} and reordered $h_j(\cdot)$ in \eqref{eqn_h_trajectory_planning}, it is straightforward to construct the constrained generation problem in \eqref{eqn_constrained_generation}.
	Moreover, their derivative in \eqref{eqn_dot_g_trajectory_planning} and \eqref{eqn_dot_h_trajectory_planning} allows direct application of the proposed UniConFlow in \cref{subsection_UniConGen}.
	
	Note that the generation of $\bm{\mathcal{T}}$ requires collecting all state and action pairs $\{ \bm{s}^k, \bm{a}^k \}_{k = 0, \cdots, H - 1}$ as demonstrations for flow matching model training, which induces high dimensional flow state and reduces training efficiency.
	
	\section{Experimental Implementation and Results}
	\label{sec_simulation}
	With the implementation details of the UniConFlow framework, we conduct three experiments to showcase its superior performance over SOTA baselines described in \cref{subsec_Experiment_Setting}. 
	The first experiment involves a double inverted pendulum as a toy simulation environment in \cref{subsec_double_pendulum}. 
	The second experiment evaluates a car-racing scenario with a real-to-sim pipeline in \cref{subsec_car_racing}. 
	The third experiment presents a sim-to-real deployment on a robotic manipulator in \cref{subsec_manipulator}.
	Finally, the ablation study is presented in~\cref{subsec_as}.

	\subsection{Experimental Setting}
	\label{subsec_Experiment_Setting}
	We evaluate the proposed UniConFlow framework against a comprehensive set of baselines\footnote{For details on the experimental setup, additional results, and training configurations, see Appendix \ref{sec_addiditonal_Experiment_Results}.}. 
	Specifically, we compare with 8 SOTA diffusion-based trajectory generation methods, 2 flow-based approaches with conventional optimization-based techniques. 
	All baselines are implemented following their original configurations and evaluated under the same experimental settings to ensure a fair comparison.
	\begin{itemize}
		\item \textbf{Diffuser}~\cite{Janner_ICML2022_Planning}: A diffusion-based motion planner that generates trajectories without ensuring safety, admissibility, or dynamic consistency. 
		\item \textbf{Truncation (Trunc)}~\cite{brockman_arXiv2016_openaigym}: A post-processing approach that enforces constraints by truncating violations in trajectories sampled from a diffusion model. 
		\item \textbf{Classifier Guidance (CG)}~\cite{Dhariwal_NEURIPS2021_Diffusion}: A diffusion model that uses classifier-based guidance in diffusion processes to guide generations toward constraint-gradient direction.
		\item \textbf{SafeDiffuser}~\cite{xiao_ICLR2025_Safediffuser}: A diffusion-based planner with three variants (\textbf{RoS}, \textbf{ReS}, \textbf{TVS}) that generates trajectories while projecting each sampled state onto the CBF-defined constraint set at every denoising step without safety guarantee.
		\item \textbf{Decision Diffuser (Diffuser-D)}~\cite{Ajay_ICLR2023_Is}: A conditional diffusion model trained to generate state trajectories based on task specifications and constraints, where the generated actions are inferred using a learned inverse dynamics. 
		\item \textbf{CoBL}~\cite{Mizuta_IROS2024_CoBL}: A gradient guided diffusion model that considers satisfying the safety and stability of goal points constraints during the denoising process. 
		\item \textbf{FM}~\cite{Lipman_ICLR2023_Flow}: A FM model that generates trajectories without considering safety, admissibility, or dynamic consistency.
		\item \textbf{PCFM}~\cite{Utkarsh_NeurIPS2025_Physics}: Trajectories are sampled from the pretrained FM model as used for prior demonstrations, considering boundary conditions as constraints, which are solved by nonlinear programming problems. 
		\item \textbf{SafeFlow}~\cite{dai2025safeflowmatchingrobot}: A FM-based planner guarantees safe trajectories while the action constraints are not considered.
		\item \textbf{FM-MPPI}: Trajectories are sampled from the pretrained FM model to serve as informed priors, which warm-start the MPPI controller for trajectory refinement.
		
	\end{itemize}
	
	To quantitatively analyze and evaluate the performance of all methods, the metrics employed across the three experiments are shown as follows:
	\begin{itemize}
		\item \textbf{Safety Rate (SR)}(\%): The proportion of generated trajectories that avoid collisions with obstacles or remain within the safe state space, where \textbf{SR-S} measures collision-free states and \textbf{SR-A} measures collision-free action rollouts.
		\item \textbf{Admissibility Rate (AR)}(\%): The proportion of trajectories that satisfy bounded actuation limits, confirming feasibility within the admissible action space $\mathbb{A}$. 
		\item \textbf{Total Success Rate (TSR)}(\%): The proportion of generated trajectories that satisfy equality and inequality constraints. 
		\item \textbf{Kinodynamic Consistency (KC)}: The forward kinodynamic consistency \textbf{(KC-F)}  is quantified by the root mean squared error (RMSE) between generated states and states obtained by forward-rolling out generated actions through the system dynamics model, defined by $\text{RMSE}_{\text{KC-F}} = (\frac{1}{H}\sum_{k=1}^{H} \|\boldsymbol{s}^{k+1} - \boldsymbol{f}(\boldsymbol{s}^k, \boldsymbol{a}^k)\|^2)^{\frac{1}{2}}$; and the inverse kinodynamic consistency \textbf{(KC-I)} is quantified by the RMSE between generated actions and actions reconstructed via inverse dynamics $\boldsymbol{f}_{\text{ID}}$ from consecutive state pairs, i.e., $\text{RMSE}_{\text{KC-I}} = (\frac{1}{H}\sum_{k=1}^{H} \|\boldsymbol{a}^k - \boldsymbol{f}_{\text{ID}}(\boldsymbol{s}^k, \boldsymbol{s}^{k+1})\|^2)^{\frac{1}{2}}$.
		\item \textbf{Cost}: The average cost value per trajectory.
		\item \textbf{Time} (ms): The average computation time of inference per trajectory.
	\end{itemize}
	
	\subsection{Toy Example: Double Inverted Pendulum}
	\label{subsec_double_pendulum}
	\begin{figure*}[t]
		\centering
		\includegraphics[width=1\linewidth]{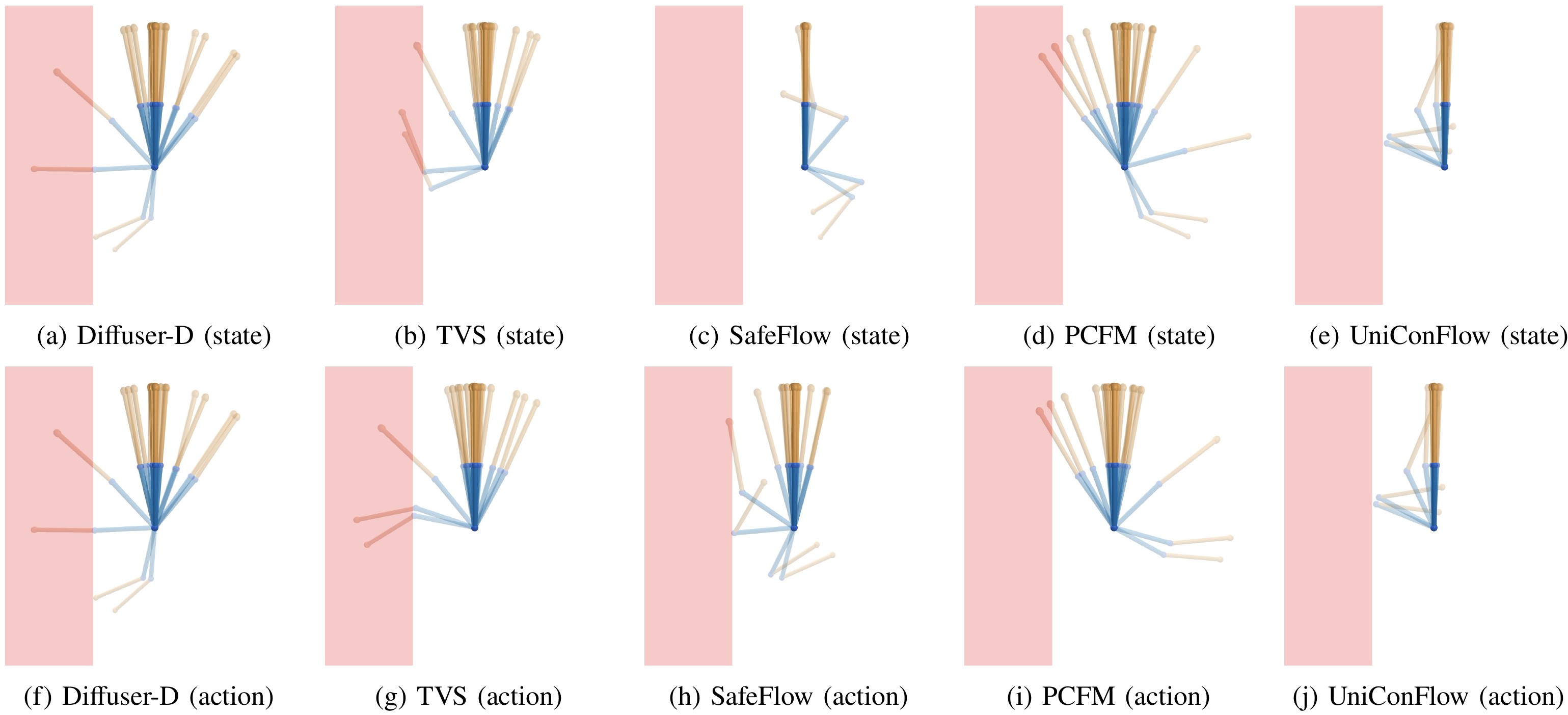}
		\caption{Visualization of the motion of the double inverted pendulum. State sequences (top row) and action rollouts (bottom row) for the stabilization task across different methods: Diffuser-D, TVS, SafeFlow, PCFM, and UniConFlow.}
		\label{fig_pen_motion}
	\end{figure*}
	
	\subsubsection{Pendulum Dynamics}
	Consider a double inverted pendulum consisting of two rigid links with equal masses \(m_1 = m_2 = 1.0 \) kg and lengths \(l_1 = l_2 = 1.0\) m. 
	The system operates in the vertical plane under gravity \(g = 9.8 \) m$/$s$^2$ and is actuated by torques \(\tau_1\) and \(\tau_2\) at the joints. 
	The generalized coordinates are the absolute angles \(q_1, q_2 \in [-\pi, \pi]\) relative to the vertical, with angular velocities \(\omega_1 = \dot{q}_1\) and \(\omega_2 = \dot{q}_2\). 
	The state vector is denoted by \(\boldsymbol{s} = [q_1, q_2, \omega_1, \omega_2]^\top\), and the control input by \(\boldsymbol{a} = [\tau_1, \tau_2]^\top\). 
	To express the system in first-order state-space form, the complete dynamics is rewritten as
	\begin{align}
		\label{eq_pen_dynamics}
		\dot{\boldsymbol{s}} \!=\! f(\boldsymbol{s}, \boldsymbol{a}) \!=\! [\omega_1, \omega_2, \boldsymbol{M}(\boldsymbol{s})^{-1} ( \boldsymbol{a} \!-\! \boldsymbol{C}(\boldsymbol{s}) \!-\! \boldsymbol{g}(\boldsymbol{s}) )]^{\top},
	\end{align}
	where the explicit model of the double inverted pendulum is provided in Appendix~\ref{subsub_ExplicitModelofInvertedPendulum}.

	\subsubsection{Experiment Setup and Inference Pipeline}
	\label{subsec:unicon_pen_pipeline}
	The control objective is to swing the pendulum to the upright equilibrium point $\boldsymbol{s}^{\mathrm{goal}} = [\pi, \pi, 0, 0]^\top$ from any random initial state. 
	Moreover, safety is imposed via a state constraint on the end-effector’s horizontal position, which must remain strictly above the wall located at $x=-1$.
	Let the base of the first link be at the origin of the world frame. And the Cartesian position of the tip of the second link is denoted by \((x_{\mathrm{ee}}(\boldsymbol{s}),y_{\mathrm{ee}}(\boldsymbol{s}))\). 
	With link lengths $l_1,l_2$ and angles $q_1, q_2$ measured from the vertical, the forward kinematics read
	\begin{align}
		x_{\mathrm{ee}}(\boldsymbol{s})
		&= l_1 \sin q_1 + l_2 \sin q_2, \\
		y_{\mathrm{ee}}(\boldsymbol{s})
		&= l_1 \cos q_1 + l_2 \cos q_2.
	\end{align}
	We set any configuration with $x_{\mathrm{ee}}(\boldsymbol{s}) < -1$ as unsafe, i.e., the pendulum has crossed a virtual wall placed at $x = -1$.
	Therefore, the safe set is $\mathbb{C}_{\mathrm{pen}}
	= \{ \boldsymbol{s} 
	| x_{\mathrm{ee}}(\boldsymbol{s}) \ge -1 \}$.
	The inequality functions are
	\begin{subequations}
		\label{eq_inequality_pen}
		\begin{align}
			&h_{\mathrm{pen}}(\boldsymbol{s}^k)
			= -1 - x_{\mathrm{ee}}(\boldsymbol{s}^k) \leq 0,\\
			&h_{\mathrm{pen}}^{a1}(\bm{a}^k) = | [1,0] \bm{a}^k |^2 -30^2 \leq 0, \\
			&h_{\mathrm{pen}}^{a2}(\bm{a}^k) = | [0,1] \bm{a}^k |^2 - 30^2 \leq 0.
		\end{align}
	\end{subequations}
	The construction of the equality constraint $g_i(\cdot)$ for dynamical consistency follows \eqref{eq_gk_con}, considering the pendulum dynamics \eqref{eq_pen_dynamics}.
	The parameters for the PTZF are chosen as $c_r = 0$, $T_{pre}=1$, $\gamma_r(t,r) = \underline{\gamma}_r(r) = r$.
	The initial values for  $r$, i.e., $\bar{g}(\cdot)$ and $\bar{h}_j(\cdot)$ in \cref{theorem_equality_constraint,theorem_inequality_constraint}, are chosen as $\bar{g}(0) = 2 g(\bm{\mathcal{T}}_0)$ and $\bar{h}_j = h_j(\bm{\mathcal{T}}_0)$.

	The UniConFlow inference procedure on the double inverted pendulum follows a two-stage structure.
	In the first stage (\emph{PTZF-based guided sampling}), the conditional FM model generates a full state-action trajectory under PTZF-based guidance. 
	The second phase, \emph{terminal refinement}, refines the trajectories derived from the action sequence rollout. 
	The primary goals are to ensure the trajectory remains within the training distribution by aligning closely with the reference path and to enforce strict constraint validity. 
	We address this optimization problem using the cross-entropy method (CEM), where the cost function is formulated as follows 
	\begin{subequations}
		\label{eq_pen_T}
		\begin{align}
			& \bm{\mathcal{T}} = \min\nolimits_{\bm{\mathcal{T}} \in \mathbb{R}^d} \| \bm{\mathcal{T}} - \bm{\mathcal{T}}_1 \|^2 \\
			\text{s.t.}~~& \eqref{eq_inequality_pen},  \forall k = 0, \cdots, H \\
			&g^k_{\text{con}}(\bm{\mathcal{T}}) = 0, \forall k = 0, \cdots, H-1.
		\end{align}
	\end{subequations}
	The overall procedure is provided in \cref{alg_unicon_pen_inference}.
	
	\begin{algorithm}[t]
		\caption{UniConFlow Inference for Inverted Pendulum}
		\label{alg_unicon_pen_inference}
		\begin{algorithmic}[1]
			\Require Conditional FM model $\bm{v}_t^{\bm{\theta}}$, guidance parameters, PTZFs $\bar{g}, \{\bar{h}_j\}_{j=1}^{N_h}$, discrete pendulum dynamics, constraint conditions, CEM hyperparameters.
			
			\Statex \textbf{PTZF-based Guided Sampling}
			\State Sample base noise trajectory $\bm{\mathcal{T}}_0 \sim p_0$;

			\While{$t_\ell <1$}
			\State $\bm{u}_{t_\ell} \gets$ QP \eqref{eqn_QP} or closed-form \eqref{eqn_QP_closed_form_solution} solution;
			\State $\bm{\mathcal{T}}_{t_{\ell}}
			\gets$ \texttt{ODE solver} with $\bm{v}_t^{\bm{\theta}}$ and $\bm{u}_{t_\ell}$;
			\EndWhile
			
			\Statex \textbf{Terminal Refinement} 
			\While{constraints violated}
			\State Run CEM over actions $\gets$ \eqref{eq_pen_T};
			\State Roll out pendulum dynamics from the entry state to obtain an updated $\bm{\mathcal{T}}$;
			\EndWhile
			
			\noindent\Return Certified trajectory
			$\bm{\mathcal{T}}^{\star}$.
		\end{algorithmic}
	\end{algorithm}

	\subsubsection{Result Analysis}
	\label{subsubsection_ratoy}
\begin{figure}
		\centering
		\includegraphics[width=1\linewidth]{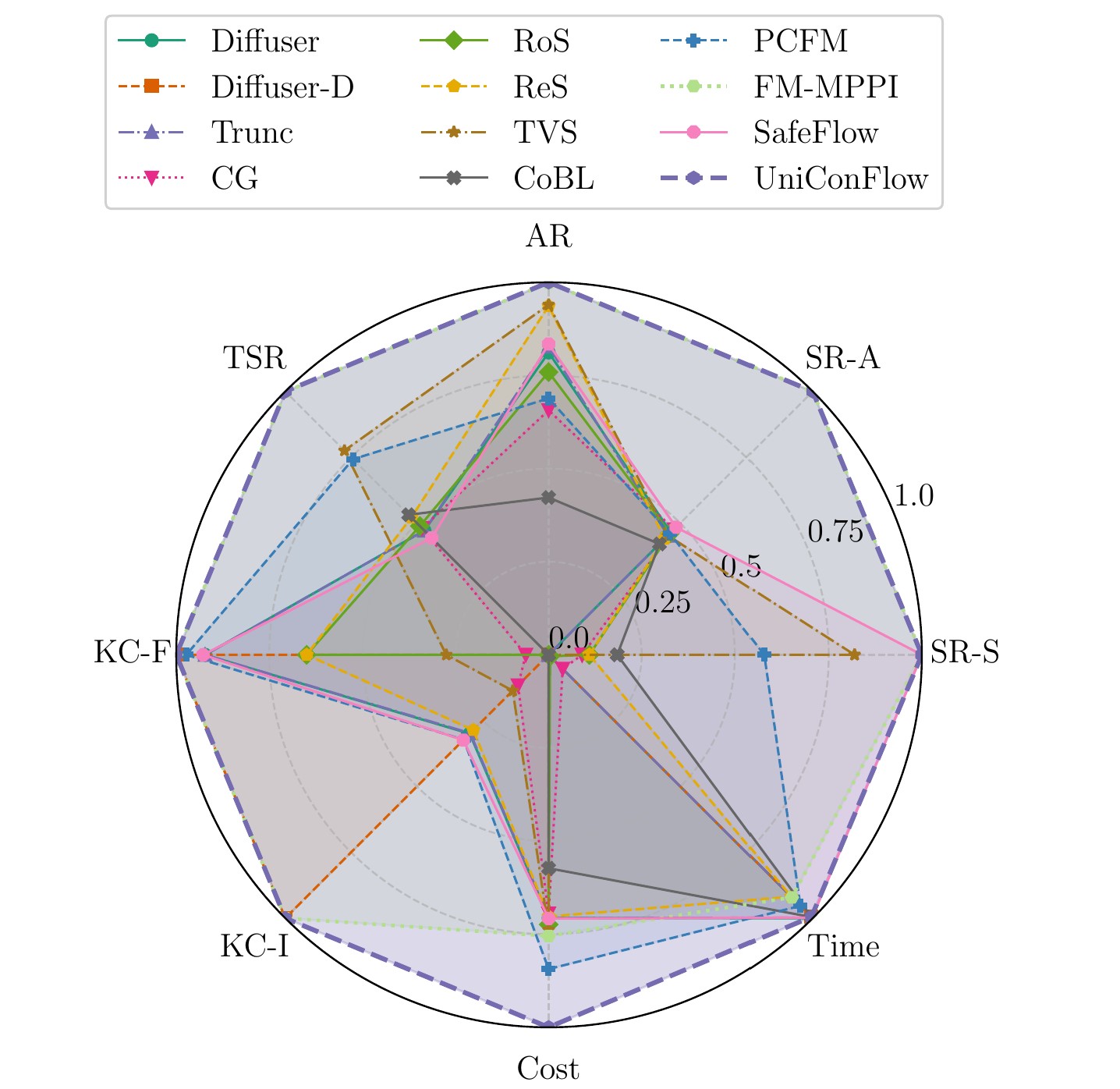}
		\vspace{-0.3cm}
		\caption{Performance comparison across normalized metrics for the double inverted pendulum scenario. SR-S, SR-A, AR, and TSR represent percentage-based success metrics, while KC-F, KC-I, Cost, and Time are inverted min-max normalized metrics (higher is better). }
		\vspace{-0.3cm}
		\label{fig_pen_radar}
	\end{figure}

	UniConFlow achieves unprecedented performance in constrained trajectory generation for the double inverted pendulum task, demonstrating perfect 100\% success rates across all success metrics (SR-S, SR-A, AR, TSR) with zero constraint violations (KC-F = 0.00, KC-I = 0.00), outperforming all baseline methods, including SOTA diffusion/FM and conventional approaches. 
	To visualize the comparison across all metrics intuitively, we present a radar chart in \cref{fig_pen_radar} to showcase the advantages of UniConFlow, where all normalized values are scaled to $[0,1]$ with inverted axes for lower-is-better metrics (KC-F, KC-I, Cost, Time) such that the outer edge uniformly represents optimal performance. 
	In \cref{tab_pen} (see Appendix \ref{subsubsection_ratoy}), both FM-MPPI and UniConFlow achieve perfect constraint satisfaction in state trajectory (100\% SR-S, 100\% SR-A) in the simple double inverted pendulum task. 
	However, UniConFlow has higher computation efficiency. It generates trajectories with lower cost and requires approximately 5 times less computation time. 
	Moreover, guidance-based methods (CG, RoS, ReS, TVS) are slow ($>130$ ms) and fail to eliminate violations, while vanilla generative models (Diffuser, FM) are fast but unsafe (86-89\% success) because they completely lack active constraint enforcement mechanisms. 
	These results demonstrate that UniConFlow offers the advantages of theoretical safety guarantees, minimal trajectory cost, and real-time inference suitable for safety-critical robotics.
	
	\subsection{Car Racing Experiment}
	\label{subsec_car_racing}
	
	\subsubsection{Nonholonomic Car Dynamics}
	In this experiment, a simplified model is employed to represent a planar, nonholonomic vehicle. 
	This model defines the vehicle's dynamic state, including its position, orientation, and longitudinal velocity over time, subject to bounded actuation inputs.
	The state vector of the model is formulated as \(\boldsymbol{s} = [x, y, \theta, v]^\top\), where \(x, y\) represent the global position coordinates of the rear-axle midpoint, \(\theta \) denotes the vehicle's heading angle relative to the global reference frame, and \(v\) denotes the forward speed.
	The control input vector is \(\boldsymbol{a} = [\delta, \tau]^\top\), with the bounded control inputs \(\tau \in [-35, 35]\) as the longitudinal acceleration command and \(\delta \in [-1,1]\) as the front-wheel steering angle. 
	The simplified dynamics is written in a state-space form  as
	\begin{align}
		\label{eq_dynamics_car}
		\dot{\boldsymbol{s}} \!=\! f(\boldsymbol{s}, \boldsymbol{a}) \!=\! 
		[
			v \cos(\theta),
			v \sin(\theta),
			v \tan(\delta) / L,
			\tau
		]^{\top},
	\end{align}
	where \(L = 2.7 \) m is the assumed wheelbase length.

	\subsubsection{Racetrack Environment Modeling}
\begin{figure*}[t]
		\centering
		\includegraphics[width=1\linewidth]{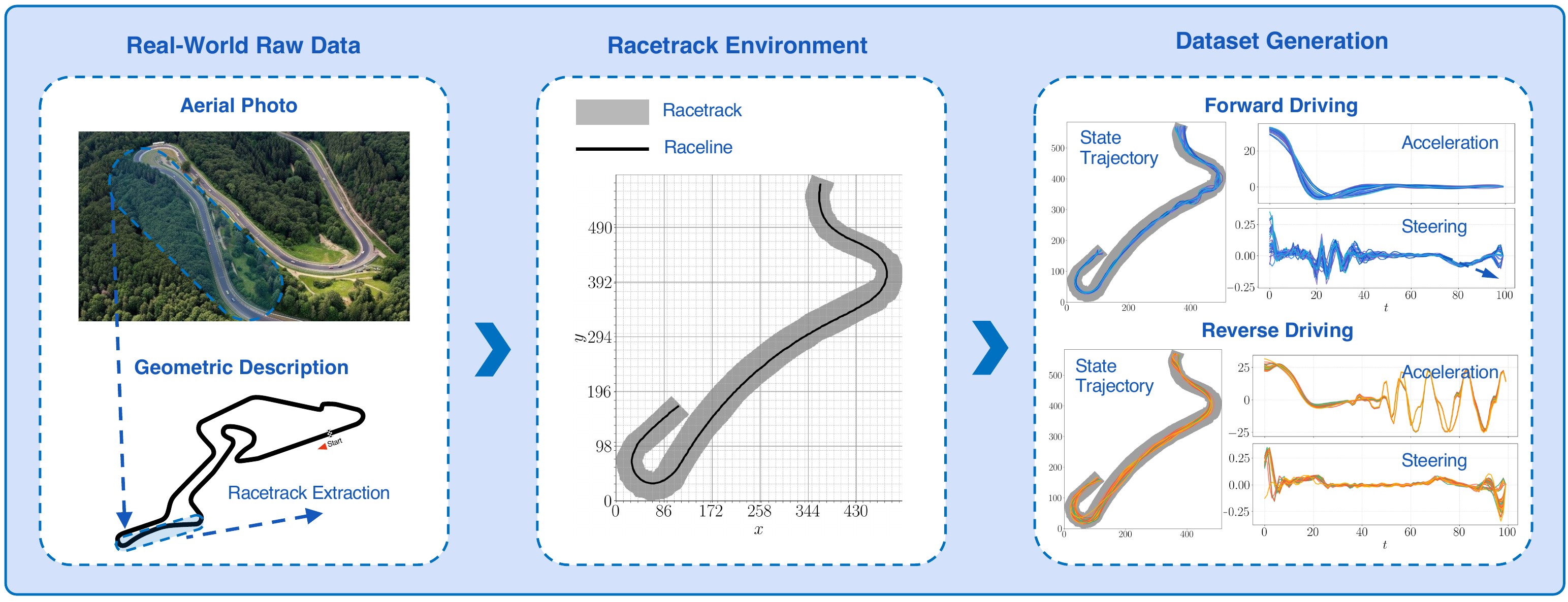}
		\caption{Dataset creation pipeline for generative models in car racing scenarios. 
			Geometric raw data including centerline, track boundaries, and raceline for the Nürburgring Nordschleife (left); extracted and rasterized track segment (middle); training data is generated through optimal control problems with random initial conditions for both forward and reverse driving (right).}
		\vspace{-0.3cm}
		\label{fig_trackdata_pipeline}
	\end{figure*}

	To construct a realistic yet controllable benchmark, we build the experimental racetrack from the Nürburgring Nordschleife data provided by the open-source \texttt{racetrack-database}~\cite{TUMFTM_racetrack_database_2025}. 
	The raw dataset contains a geometric description of the track centerline together with the left and right track widths, and, when available, an associated racing line. 
	The processing pipeline, illustrated in~\cref{fig_trackdata_pipeline}, comprises three stages for real-to-sim environment construction and dataset generation.
	
	First, we extract a representative segment of the full loop and compute its local geometry. 
	The sampled centerline is reparameterized, tangential directions are obtained by finite differences, and outward normals are used to reconstruct the left and right track boundaries from the stored width profiles. 
	This step yields a smooth planar track segment with explicit centerline, boundaries, and raceline.
	
	Second, we inflate the track to obtain an enlarged drivable corridor and rasterize the resulting polygonal region into a 2D small grid on the rasterized track map of size \(585 \times 514\) pixels. 
	The interior of the track is encoded as free space and the exterior as obstacles, while the raceline is discretized into a sequence of grid coordinates. 
	This representation is convenient for both motion planning and visualization, and can be generated at an arbitrary resolution.
	
	Finally, to assess the performance of the generative model and establish a benchmark, we create a training dataset by using an optimal control approach, where the details can be found in Appendix~\ref{subsubsec_Dataset_Creation_for_Car_Racing_Experiment}.
	
	Notably, although our experiments focus on the Nürburgring Nordschleife, environment representation can be readily modeled on other racetracks.

	\subsubsection{Experiment Setup and Inference Pipeline}
	\label{subsec:unicon_pipeline}
	The objective is to navigate the vehicle from a designated start state to a goal state while adhering closely to the reference raceline, remaining within the drivable corridor, and avoiding collisions with obstacles.
	We evaluate two obstacle scenarios: in the forward-driving case, three static ellipsoidal obstacles are randomly distributed along the track, whereas in the reverse-driving case, a single bone-shaped, irregular obstacle is positioned in the lane center.

	\begin{figure*}[h]
		\centering
		\includegraphics[width=1\linewidth]{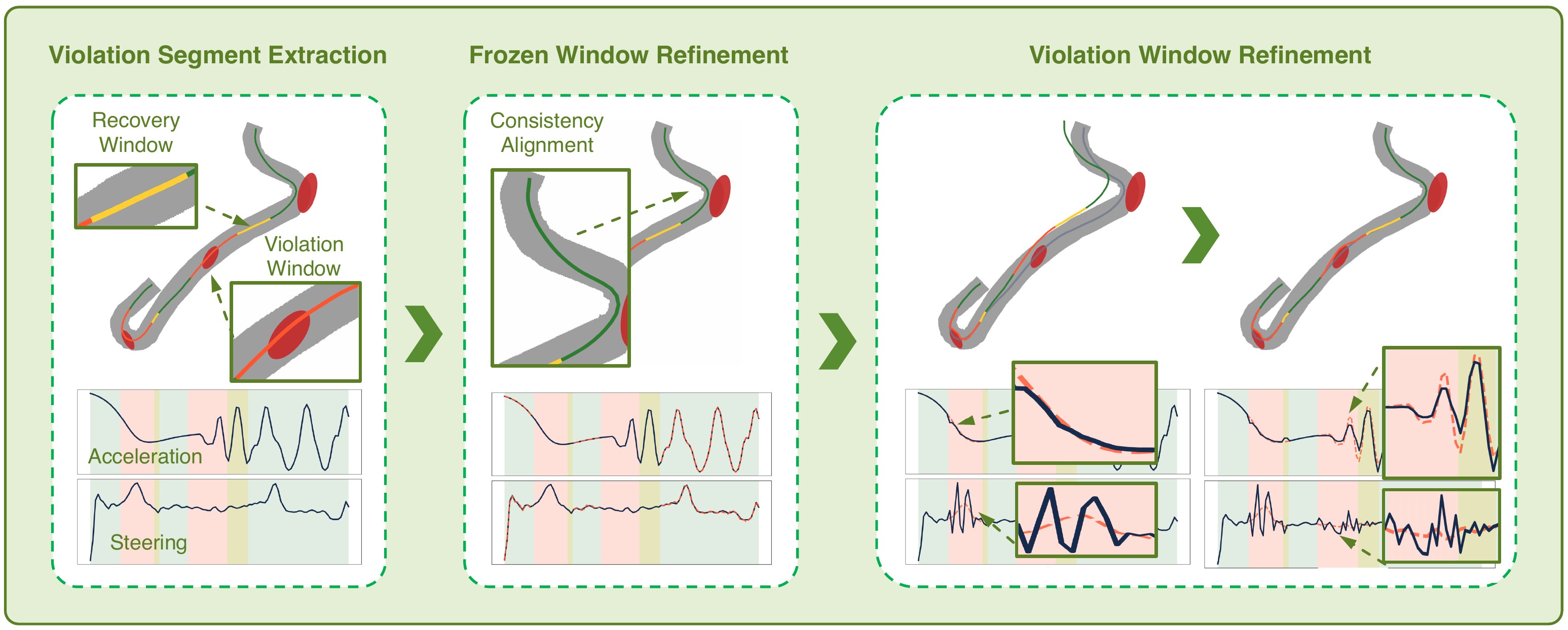}
		\caption{
			Terminal refinement process for the car racing scenario. The window-based segmentation strategy preserves the global trajectory shape (left). 
			Refine the safe segment where kinodynamic consistency may be violated, due to numerical errors (middle). The final refined trajectory achieves safety, admissibility, and kinodynamic consistency (right).
		}
		\vspace{-0.3cm}
		\label{fig_carref}
	\end{figure*}

	For each position $\boldsymbol{\xi}^k = [x^k, y^k]^\top$, let 
	$\boldsymbol{\xi}_{\mathrm{trk}}^{L}(k)$ and $\boldsymbol{\xi}_{\mathrm{trk}}^{R}(k)$
	denote the closest points on the left and right track boundaries, respectively.
	Let $\boldsymbol{n}^L(k)$ and $\boldsymbol{n}^R(k)$ be the corresponding inward-pointing unit normals.
	Then, we define the signed distances to the left and right boundaries as
	$d_{\mathrm{trk}}^{L}(\boldsymbol{\xi}^k)= \boldsymbol{n}^L(k)^\top \bigl(\boldsymbol{\xi}^k - \boldsymbol{\xi}_{\mathrm{trk}}^{L}(k)\bigr),~d_{\mathrm{trk}}^{R}(\boldsymbol{\xi}^k)= \boldsymbol{n}^R(k)^\top \bigl(\boldsymbol{\xi}_{\mathrm{trk}}^{R}(k) - \boldsymbol{\xi}^k\bigr)
	$.
	Both distances are positive inside the drivable corridor, zero on the corresponding boundary, and negative outside. 
	Let $d_{\mathrm{obs}}^{(o)}(\boldsymbol{\xi}^k)$ denote the signed distance to the $o$-th ellipsoidal obstacle, with negative values inside the obstacle and zero on its boundary.
	The signed distance to the union of all (possibly composite) obstacles is then defined as
	\begin{equation}
		d_{\mathrm{obs}}(\boldsymbol{\xi}^k)
		= \min\nolimits_{o} d_{\mathrm{obs}}^{(o)}(\boldsymbol{\xi}^k).
	\end{equation}
	We define the aggregated inequality constraint as
	\begin{equation}
		h_{\mathrm{car}}(\boldsymbol{s}^k)
		=
		-\min\big\{
		d_{\mathrm{trk}}^{L}(\boldsymbol{\xi}^k),\,
		d_{\mathrm{trk}}^{R}(\boldsymbol{\xi}^k),\,
		d_{\mathrm{obs}}(\boldsymbol{\xi}^k)
		\big\}.
		\label{eq:h_car_state}
	\end{equation}
	By construction, $h_{\mathrm{car}}(\boldsymbol{s}^k) \le 0$ if and only if $\boldsymbol{\xi}^k$ lies within the track corridor and outside all obstacles, while $h_{\mathrm{car}}(\boldsymbol{s}^k) > 0$ indicates either a track–limit violation or a collision with ellipsoidal obstacles.
	We also use the trajectory–level aggregate $h_{\mathrm{car}}(\bm{\mathcal{T}})
	= \max_{0 \le k \le H} h_{\mathrm{car}}(\boldsymbol{s}^k)$, which is nonpositive if and only if the entire rollout is safe.
	The inequality functions are
	\begin{subequations}
		\label{eq_inequality_car}
		\begin{align}
			&h_{\mathrm{car}}(\bm{\mathcal{T}}) \leq 0,\\
			&h_{\mathrm{car}}^{a1}(\bm{a}^k) = | [1,0] \bm{a}^k |^2 -100^2 \leq 0, \\
			&h_{\mathrm{car}}^{a2}(\bm{a}^k) = | [0,1] \bm{a}^k |^2 - \pi^2 \leq 0.
		\end{align}
	\end{subequations}
	The construction of the equality constraint $g_i(\cdot)$ for dynamical consistency follows \eqref{eq_gk_con}, considering the car dynamics \eqref{eq_dynamics_car}.
	
	The UniConFlow inference framework consists of PTZF-based guided sampling followed by terminal refinement. 
	While sufficient for simpler systems like the double inverted pendulum, long-horizon planning tasks involving multiple obstacles require an efficient approach to ensure constraint satisfaction. 
	To this end, we introduce a three-step strategy that mitigates the computational complexity during optimization: (1) violation segment extraction; (2) frozen-window refinement; and (3) violation-window refinement combined with global certification (see \cref{fig_carref}).
	
	In the stage of \emph{PTZF-based guided sampling}, the conditional FM model generates a full race–car trajectory \({\bm{\mathcal{T}}} \in \mathbb{R}^{604}\).
	During sampling, UniConFlow is applied as an ODE guidance term, so that the flow
	\(\dot{\bm{\mathcal{T}}}_t= \bm{v}_t^{\bm{\theta}}(\bm{\mathcal{T}}_t) + \bm{u}_t\)
	follows a PTZF direction for all equality and inequality constraints, which is the same procedure in \cref{subsec:unicon_pen_pipeline}
	
	The first step in the terminal refinement (\emph{violation segment extraction}) analyzes the sampled trajectory for constraint violations. 
	Different from the aggregated inequality constraint in~\eqref{eq:h_car_state}, the constraint $h_{\mathrm{car}}(\boldsymbol{s}^k) \le 0$ corresponds to a safe state. 
	We first mark all violating step indices
	\begin{align}
		\mathcal{I}_{\mathrm{vio}} := \big\{k \in \{0,\dots,H-1\} \big| h_{\mathrm{car}}(\boldsymbol{s}^k) > 0 \big\}.
	\end{align}
	Then the contiguous indices in $\mathcal{I}_{\mathrm{vio}}$ are grouped into raw violation intervals \([\tilde{k}_i^{-}, \tilde{k}_i^{+}]\), $i = 1,\dots,m$. 
	Each raw interval is then dilated in time by a user-chosen padding parameter $N_{\mathrm{pad}} \in \mathbb{N}$, such that $k_i^{-} := \max(0,\tilde{k}_i^{-} - N_{\mathrm{pad}}), ~
	k_i^{+} := \min(H-1,\tilde{k}_i^{+} + N_{\mathrm{pad}})$,
	and the corresponding violation window is defined as the index set
	$\mathcal{W}_{\mathrm{vio}}^{(i)} := \{k_i^{-},\dots,k_i^{+}\}$.
	Immediately after each enlarged violation window, we attach a recovery window of fixed length $N_{\mathrm{rec}} \in \mathbb{N}$:
	\[
	\mathcal{W}_{\mathrm{rec}}^{(i)} := 
	\{k_i^{+} + 1,\dots,k_i^{+} + N_{\mathrm{rec}}\} \cap \{0,\dots,H\},
	\quad i = 1,\dots,m.
	\]
	These recovery windows allow the trajectory to smoothly return from a locally repaired segment to the next frozen segment. 
	The remaining time indices are assigned to frozen windows.
	More precisely, we consider the complement of all violation and recovery indices, $\{0,\dots,H\}
	\setminus 
	\bigcup_{i=1}^{m}
	\big(\mathcal{W}_{\mathrm{vio}}^{(i)} \cup \mathcal{W}_{\mathrm{rec}}^{(i)}\big)$,
	and split it into $(m{+}1)$ contiguous index sets
	\(
	\{\mathcal{W}_{\mathrm{frz}}^{(i)}\}_{i=1}^{m+1}
	\),
	ordered along the time axis. 
	By construction, $\mathcal{W}_{\mathrm{frz}}^{(1)}$ contains $k=0$ and represents the frozen prefix, $\mathcal{W}_{\mathrm{frz}}^{(m+1)}$ the frozen suffix, and intermediate frozen windows lie between violation--recovery blocks.
	Throughout, each window $\mathcal{W}_{\mathrm{frz}}^{(i)}, \mathcal{W}_{\mathrm{vio}}^{(i)}, \mathcal{W}_{\mathrm{rec}}^{(i)}$ is a set of time indices.
	
	In the second step (\emph{frozen-window refinement}), UniConFlow enforces kinodynamic consistency on the frozen windows while preserving their overall shape.
	CEM updates all the corresponding states of the frozen windows based on the alignment cost $J_{\mathrm{pref}}$ in \eqref{eq:J_pref}, and the dynamics \eqref{eq_dynamics_car} are rolled out to update \((\bm{s}^k, \bm{a}^k)\) on these windows.
	We denote by $\{\hat{\bm{s}}^k\}_{k \in \mathcal{W}_{\mathrm{frz}}^{(i)}}$ the corresponding PTZF guided states after the PTZF-based guided sampling.
	The CEM alignment on the prefix refines the controls on each $\mathcal{W}_{\mathrm{frz}}^{(i)}$ such that the dynamics rollout matches the original trajectory.

	The third step (\emph{violation-window refinement and global certification}) repairs the remaining infeasible parts and produces a certified full-horizon trajectory. 
	For each violation window index \(i\), we refine the violation and recovery windows while keeping all previously certified segments fixed. 
	Starting from the current frozen prefix $\mathcal{W}_{\mathrm{frz}}^{(i)}$, CEM with cost $J_{\mathrm{vio}}^{(i)}$ in \eqref{eq:J_vio} is run on $\mathcal{W}_{\mathrm{vio}}^{(i)}$, and the dynamics \eqref{eq_dynamics_car} are rolled out to update $(\bm{s}^k,\bm{a}^k)$ on this window. 
	This is followed by a recovery window enforcement on $\mathcal{W}_{\mathrm{rec}}^{(i)}$ with cost $J_{\mathrm{rec}}^{(i)}$ in \eqref{eq:J_tr}, again followed by a rollout on $\mathcal{W}_{\mathrm{rec}}^{(i)}$. 
	These two CEM updates are repeated until all safety and track-limit constraints are satisfied on $\mathcal{W}_{\mathrm{vio}}^{(i)} \cup \mathcal{W}_{\mathrm{rec}}^{(i)}$. 
	The resulting certified subtrajectory $\mathcal{W}_{\mathrm{cer}}^{(i)}$ is obtained by concatenating, in temporal order, the $i$-th frozen window $\mathcal{W}_{\mathrm{frz}}^{(i)}$, the optimized violation window $\mathcal{W}_{\mathrm{vio}}^{(i)}$, and the corresponding recovery window $\mathcal{W}_{\mathrm{rec}}^{(i)}$.
	After processing all $m$ violation windows, a final CEM refinement is applied on the concatenation of the last certified window $\mathcal{W}_{\mathrm{cer}}^{(m)}$ and the frozen window $\mathcal{W}_{\mathrm{frz}}^{(m+1)}$, yielding the certified trajectory \(\bm{\mathcal{T}}^{\star}\), which strictly satisfies all equality and inequality constraints. 
	The overall algorithmic procedure is outlined in \cref{alg_unicon_inference_car}.

	\subsubsection{Result Analysis}
	
	\begin{algorithm}[t]
		\caption{UniConFlow Inference Pipline}
		\label{alg_unicon_inference_car}
		\begin{algorithmic}[1]
			\Require Conditional FM model $\bm{v}_t^{\bm{\theta}}$, guidance parameters, PTZFs $\bar{g}, \{\bar{h}_j\}_{j=1}^{N_h}$, dynamics $\bm{f}^k$, obstacle and track costs, CEM hyperparameters.
			
			\Statex \textbf{PTZF-based Guided Sampling}
			\State Sample a base noise trajectory $\bm{\mathcal{T}}_0 \sim p_0$ 
			\For{$t_\ell = 0$ to $1$}
			\State $\bm{u}_{t_\ell}$ $\gets$ QP \eqref{eqn_QP} or closed-form \eqref{eqn_QP_closed_form_solution} solutions;
			\State $\bm{\mathcal{T}}_{t_{\ell+1}} \gets \bm{\mathcal{T}}_{t_\ell} + \Delta t_\ell \big( \bm{v}_{t_\ell}^{\bm{\theta}}(\bm{\mathcal{T}}_{t_\ell}) + \bm{u}_{t_\ell} \big)$;
			\EndFor
			
			\Statex \textbf{Violation Segment Extraction} 
			\State Evaluate collisions and track violations along $\{{\bm{s}}^k\}_{k=0}^{H}$;
			\State Extract the frozen, violation and recovery windows: $\{\mathcal{W}_{\text{frz}}^{(i)}\}_{i}^{m+1}$,
			$\{\mathcal{W}_{\text{vio}}^{(i)}\}_{i}^m$ $\{\mathcal{W}_{\text{rec}}^{(i)}\}_{i}^m$;
			
			\Statex \textbf{Frozen Window Enforcement}
			\While{constraints violated}
			\State Run CEM on  $\{\mathcal{W}_{\text{frz}}^{(i)}\}_{i=1,\dots,m+1}$; 
			\State Rollout dynamics \eqref{eq_dynamics_car} to update  $(\bm{s}^k,\bm{a}^k)$ on  $\{\mathcal{W}_{\text{frz}}^{(i)}\}_{i=1,\dots,m+1}$;
			\EndWhile 
			
			\Statex \textbf{Violation Windows Enforcement}
			\For{$i = 1, \dots, m$}
			\While{constraints violated}
			\State Run CEM on $\mathcal{W}_{\text{vio}}^{(i)}$; 
			\State Rollout dynamics \eqref{eq_dynamics_car} to update $(\bm{s}^k,\bm{a}^k)$ on $\mathcal{W}_{\text{vio}}^{(i)}$;
			\State Run CEM on $\mathcal{W}_{\text{rec}}^{(i)}$;
			\State Get certified window $\mathcal{W}_{\text{cer}}^{(i)}$ by concatenating $\mathcal{W}_{\text{frz}}^{(i)}$ with the $i$-th $\mathcal{W}_{\text{rec}}^{(i)}$ and $\mathcal{W}_{\text{vio}}^{(i)}$;
			\EndWhile
			\EndFor
			\State Run CEM on the concatenated $\mathcal{W}_{\text{cer}}^{(m)}$ with $\mathcal{W}_{\text{frz}}^{(M+1)}$;
			
			\noindent\Return The certified trajectory of the full pipeline: 
			$\bm{\mathcal{T}}^{\star} = [(\bm{s}^{\star,k},\bm{a}^{\star,k})]_{k=0}^{H}$.
		\end{algorithmic}
	\end{algorithm}
	
	\begin{figure*}[t]
		\centering
		\includegraphics[width=1\linewidth]{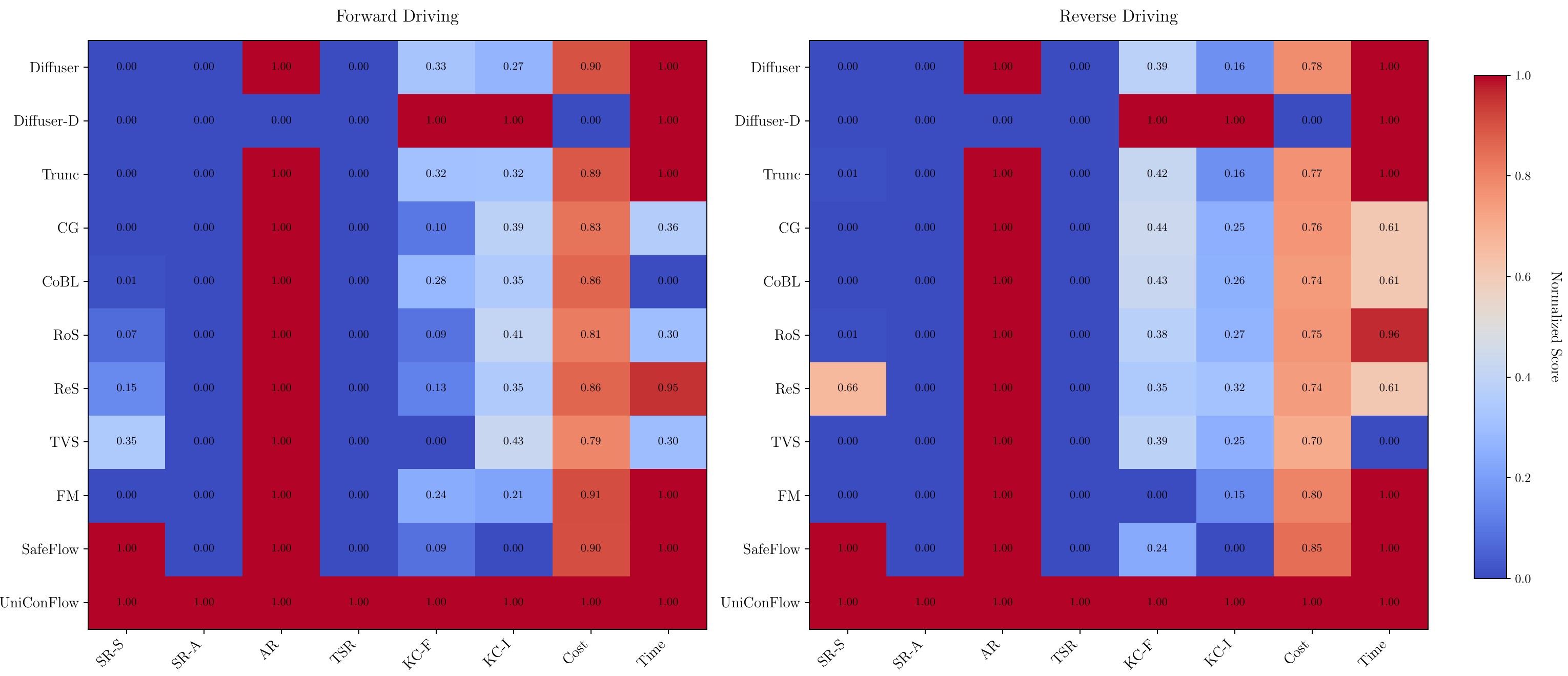}
		\caption{
			Performance heatmap for car racing scenarios in forward and reverse driving cases. 
			All values are normalized to $[0,1]$ with inversion applied to KC, Cost, and Time such that higher scores uniformly represent better performance.
			}
		\vspace{-0.3cm}
		\label{fig_car_racing_coolwarm}
	\end{figure*}
	
	\begin{figure}[t]
		\centering
		\includegraphics[width=1\linewidth]{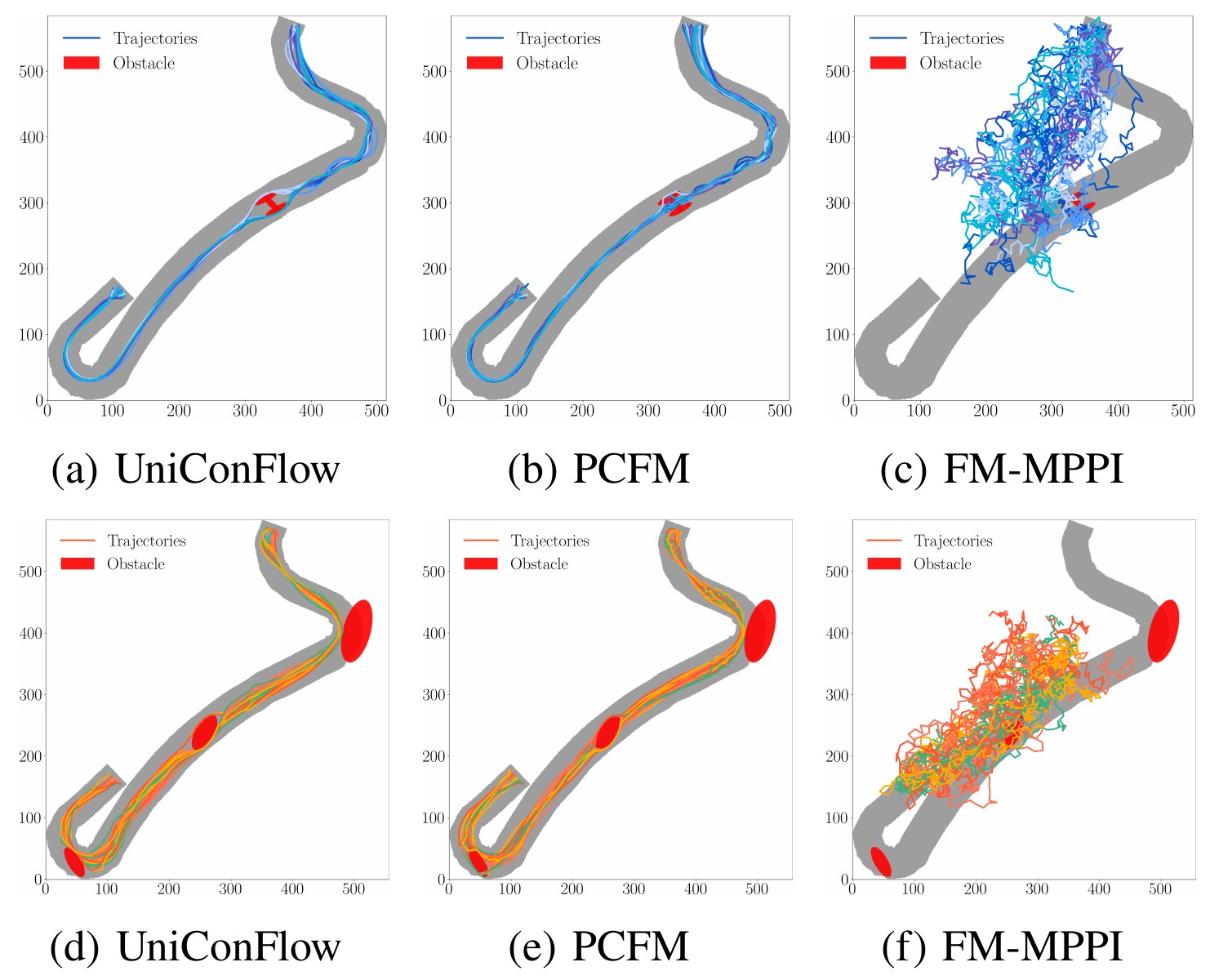}
		\caption{Visualization of state and action rollout trajectories generated by UniConFlow versus optimization-based methods.}
		\vspace{-0.3cm}
		\label{fig_cartrajectories}
	\end{figure}
	
	To provide an intuitive comparison of relative performance across methods, \cref{fig_car_racing_coolwarm} presents a normalized heatmap that demonstrates UniConFlow's consistently superior metrics across all evaluation criteria. 
	Note that PCFM and FM-MPPI are omitted from the heatmap visualization, as both methods failed to compute solutions within 180 seconds in the car racing task.  
	In \cref{fig_cartrajectories}, the state trajectories show that PCFM fails to avoid the obstacles and FM-MPPI produces random outputs, showing their computational intractability for real-time planning.
	
	In \cref{tab_forward_driving,tab_reverse_driving} (see Appendix \ref{subsubsection_arcar}) for the high-dimensional, dynamically constrained car racing scenarios, UniConFlow is the only method to successfully solve both forward and reverse driving tasks with perfect 100\% scores across all metrics and zero constraint violations. 
	In contrast, all baseline methods fail catastrophically: vanilla generative models (Diffuser, FM) and guidance-based approaches (CG, RoS, ReS, TVS, CoBL) achieve 0\% on the TSR metric despite maintaining 100\% AR, indicating generated trajectories are dynamically infeasible or unsafe trajectories with consistency violations (KC-F$>$23, KC-I$>$3.6). 
	Even SafeFlow, which matches UniConFlow's 100\% SR-S, completely fails on SR-A. 
	Moreover, the optimization-based methods (FM-MPPI, PCFM) timeout entirely ($>180$ s) with costs exceeding $6 \times 10^5$. 
	Therefore, UniConFlow not only eliminates all violations but simultaneously achieves lower trajectory costs and maintains real-time performance, demonstrating that its unified constraint enforcement framework scales robustly to complex non-holonomic vehicle dynamics where traditional generative and optimal control methods fundamentally break down. 
	
	\subsection{Manipulator Experiment}
	\label{subsec_manipulator}
	
	For the sim-to-real evaluation on the 7-DoF Franka Research 3 (FR3) manipulator, we first generate a training dataset in the MuJoCo~\cite{Todorov_IROS2012_MuJoCo} simulation environment via model predictive control (MPC). 
	The generative models are trained on this dataset before deploying the resulting UniConFlow planner directly on the hardware for evaluation.
	Compared to the car-racing benchmark, this task presents significantly greater challenges. Due to the high control frequency (1000 Hz) required by the FR3 to ensure real-time, responsive execution on the physical platform, the resulting state-action trajectories are both high-dimensional and long-horizon. 
	Consequently, even minor errors in state or action sequences can accumulate rapidly during open-loop execution. 
	As a result, kinodynamic inconsistency can easily lead to obstacle collisions or joint limit violations on the manipulator, making this benchmark a sensitive stress test for generative models.

	\subsubsection{Experiment Setup}
\begin{figure}[t]
		\centering
		\includegraphics[width=1\linewidth]{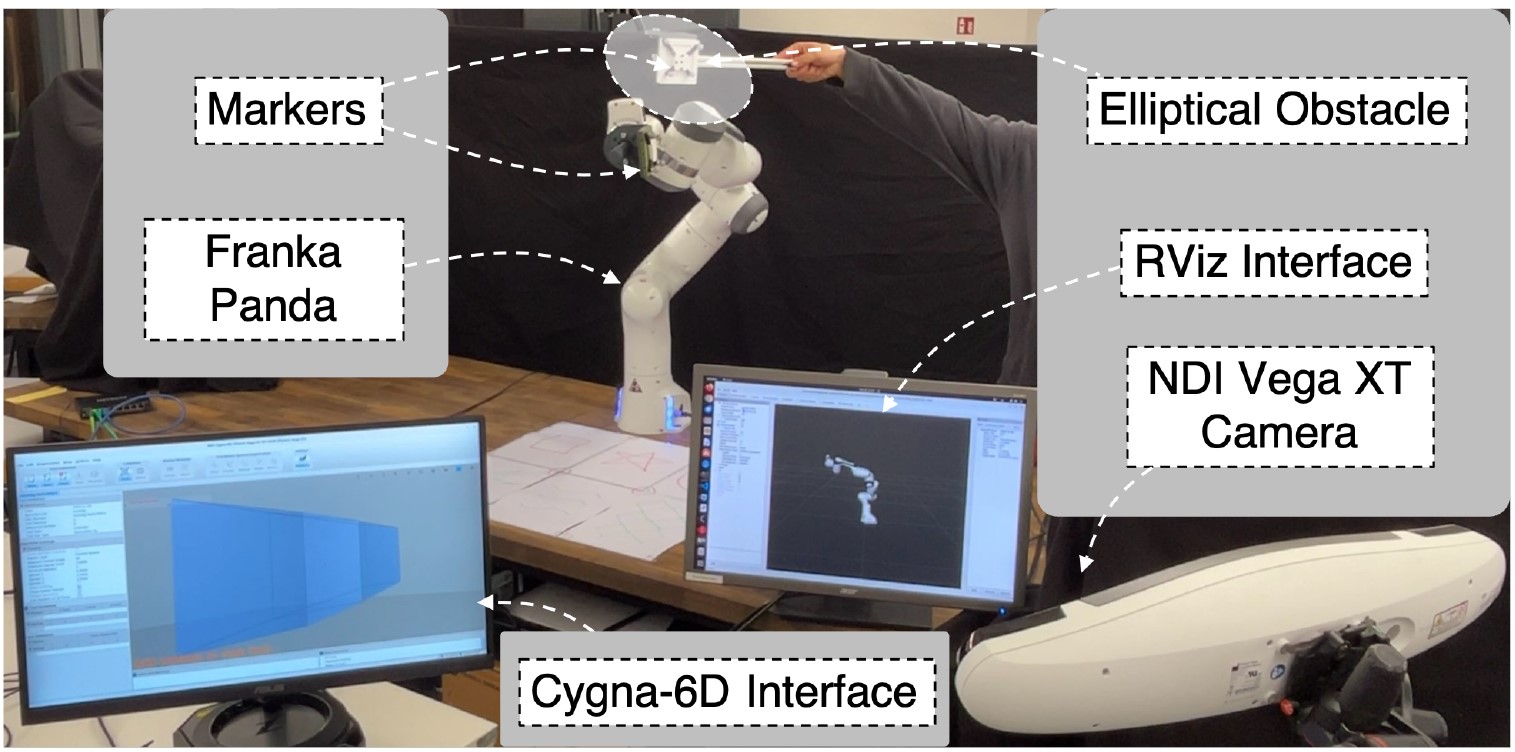}
		\caption{
			The experimental setup for the manipulator. 
			The obstacle is modeled as an ellipsoid for collision avoidance purposes, with its geometric center defined by the attached marker. 
			Marker positions are measured via an NDI Polaris Vega XT optical tracker visualized in Cigna-6D, while the robot state is visualized in RViz. 
		}
		\vspace{-0.3cm}
		\label{fig_experiment_setup_arm}
	\end{figure}

	The FR3 Panda arm is equipped with a rigid stick end-effector, which is used to track the 3D spatial trajectory while avoiding a set of obstacles placed around the motion trajectory. 
	We conduct two experiments to evaluate UniConFlow's performance across different motion scenarios: one in a pseudo-2D workspace for intuitive visualization, and another involving full 3D spatial motion.
	In the first experiment, the stick endpoint and obstacles are arranged such that the task can be equivalently described as a planar obstacle-avoidance problem on the $yz$ projection of the workspace. Specifically, the $x$-coordinate of the stick tip is fixed, while its $y$- and $z$-coordinates trace out the desired figure-eight trajectory. 
	Notably, while this pseudo-2D motion simplifies visualization, the control problem remains fully high-dimensional, as the manipulator must still solve for all seven joint angles to execute the planar end-effector motion. 
	In the second experiment, the spatial trajectory consists of a simple circular motion in 3D space. 
	The obstacle is placed along the center of the trajectory path. This setup tests UniConFlow's ability to generate collision-free motion in unconstrained three-dimensional workspaces, representing a more general manipulation scenario.
	To determine the obstacle positions relative to the robot, we employ a motion capture system. 
	Reflective markers are attached to the end-effector and to the center of the physical obstacle. 
	By tracking the relative distance between these markers, we derive the obstacle's spatial coordinates in the robot's frame. 
	The experimental setup is illustrated in \cref{fig_experiment_setup_arm}. 
	
	The generalized coordinates are the joint positions $\boldsymbol{q} \in \mathbb{R}^7$, and the control inputs are the joint torques $\boldsymbol{\tau} \in \mathbb{R}^7$. 
	Therefore, the state and action are defined as $\boldsymbol{s} = [\boldsymbol{q}^\top, \dot{\boldsymbol{q}}^\top]^\top \in \mathbb{R}^{14}, ~\boldsymbol{a} = \boldsymbol{\tau} \in \mathbb{R}^{7}$.
	The manipulator dynamics is the same as in \emph{Example 1}. 
	Let $\boldsymbol{p}_{ee}^{k}(\boldsymbol{q}) \in \mathbb{R}^3$ denote the stick endpoint position, and let
	$\boldsymbol{p}_{yz}^{k}(\boldsymbol{q}) \in \mathbb{R}^2$ denote its projection onto the $yz$ plane with fixed $x$.
	Because the figure–eight is drawn by moving the stick tip in this vertical plane, only the $yz$–projections of the stick tip and of all obstacles are relevant for safety.
	
	Obstacle are modeled as unions of ellipses in the $yz$ plane, and inequality constraints are encoded via functions $h_{\text{arm}}(\cdot)$, enforcing that $\boldsymbol{p}_{yz}^{k}(\boldsymbol{q})$ remains inside a prescribed safe corridor around the desired figure-eight while keeping a margin to all obstacles and admissible actions. 
	The distance of between each ellipse $i$ border and the $\boldsymbol{p}_{yz}^{k}(\boldsymbol{q})$ denotes $d_{\text{obs}}^{i}(\boldsymbol{p}_{yz}^{k}(\boldsymbol{q}))$. 
	Therefore, the aggregated inequality constraint is 
	\begin{equation}
		h_{\mathrm{arm}}(\boldsymbol{s}^k)
		=
		-\min\nolimits_{i=1,\dots,m}\big\{
		d_{\mathrm{obs}}^i(\boldsymbol{\xi}^k)
		\big\},
		\label{eq_h_arm_state}
	\end{equation}
	where $m$ is the number of obstacles. 
	The construction of the equality constraint $g_i(\cdot)$ for dynamical consistency follows \eqref{eq_gk_con}, considering the manipulator dynamics \eqref{eq_arm_dynamics}.

	\subsubsection{Trajectory Compression}
	To obtain expert demonstrations on the manipulator task, we first synthesize a periodic figure-eight motion using MPC running on the full joint-space dynamics introduced above in the case of pseudo-2D.
	The MPC solves a receding-horizon optimal control problem at $500\,\mathrm{Hz}$, tracking a reference figure-eight in the end-effector $yz$-plane while enforcing joint limits and torque bounds.
	At each control step, only the first input of the optimal sequence is applied, the horizon is shifted forward, and the optimization is repeated.
	After a short transient, this closed-loop scheme converges to a stable limit cycle in joint space that realizes the desired figure-eight motion in task space.
	The dataset is established by random sliding selection from the recorded 5 cycles.
	The joint trajectories in the manipulator benchmark contain $ 2.6\times 10^4$ time steps with a $14$-dimensional robot state
	$(\boldsymbol{q},\dot{\boldsymbol{q}})$ and a $7$-dimensional torque control $\boldsymbol{\tau}$ at each step.
	Therefore, running the ODE directly in such a high-dimensional space is computationally expensive and practically untenable for deployment on the real manipulator.
	
	To keep the ODE tractable, we perform the FM model in a compact compressed space and decode intermediate compressed states back to full state–action trajectories, where all UniCon constraints and PTZFs are evaluated.
	Thus, compression only affects numerical efficiency, while constraint enforcement always happens in the robot’s native space.
	We select a suitable encoder–decoder pair $(E,D)$, which is a cubic $B$–spline representation that directly parametrizes each joint coordinate by a small set of spline control points.

	\subsubsection{Inference Pipeline}
	
	\begin{figure*}
		\centering
		\includegraphics[width=1\linewidth]{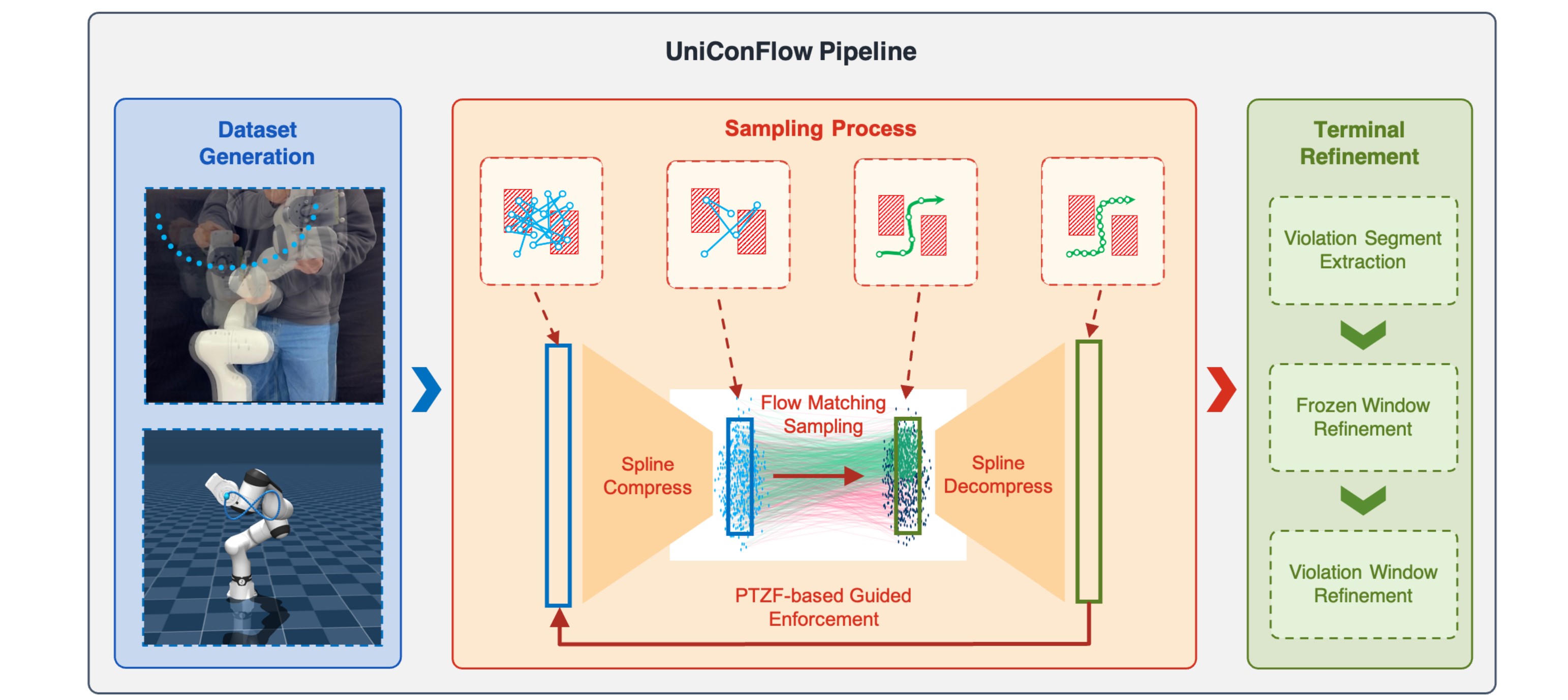}
		\caption{The datasets (left) are obtained either by expert demonstration with Franka arm or optimal control in a simulation environment. In the sampling process (middle), the rectangle filled with diagonal lines signifies the obstacles. The green path represents the collision-free trajectory navigating between red obstacles within the bounded workspace after PTZF-based guided sampling, while the blue path before PTZF-based guided sampling. The terminal refinement (right) is the same process in the car racing case in \cref{fig_carref}.}
		\label{fig_unipipe}
	\end{figure*}
	
	The inference procedure follows a two-stage structure analogous to the double inverted pendulum and car-racing benchmark. 
	To address the computational challenges posed by high-dimensional, long-horizon manipulator trajectories, we introduce a trajectory compression mechanism in the sampling stage. 
	During the first stage (PTZF-guided sampling), the conditional flow matching model generates trajectories in a compressed space defined by an encoder-decoder pair $(E,D)$. 
	Importantly, both $E$ and $D$ are deterministic, linear operators defined strictly by B-spline basis functions, ensuring an exact mathematical mapping without any learned reconstruction error.
	The process is illustrated in \cref{fig_unipipe}. 
	In the second stage, \emph{terminal refinement}, the procedure remains identical to that of the car-racing scenario, comprising three sub-steps: violation segment extraction, frozen-window refinement, and violation-window refinement.

	\subsubsection{Result Analysis}
	\begin{figure}[t]
		\centering
		\includegraphics[width=1\linewidth]{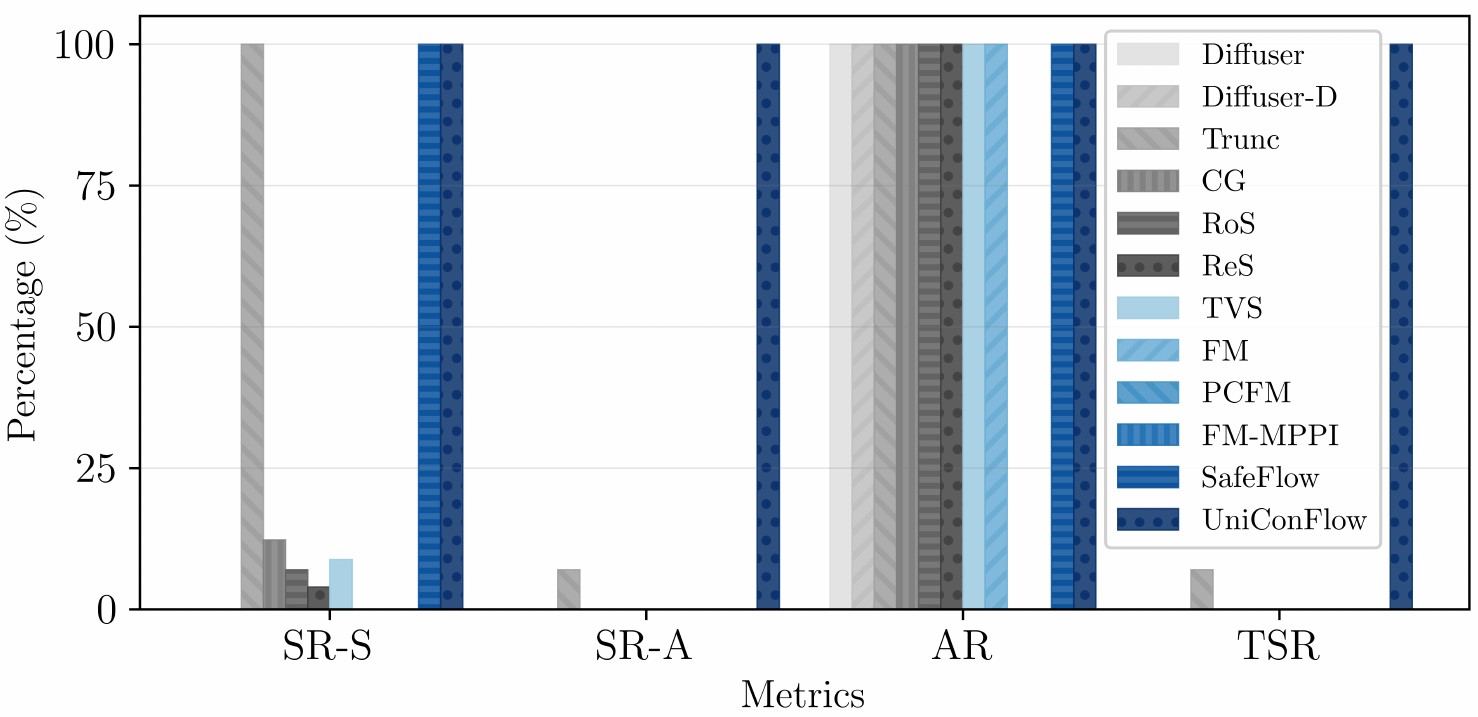}
		\caption{Success rate in 3D spatial motion scenario.}
		\vspace{-0.3cm}
		\label{fig_manipulator_3d_success_rates}
	\end{figure}
	
	\begin{figure}[t]
		\centering
		\includegraphics[width=1\linewidth]{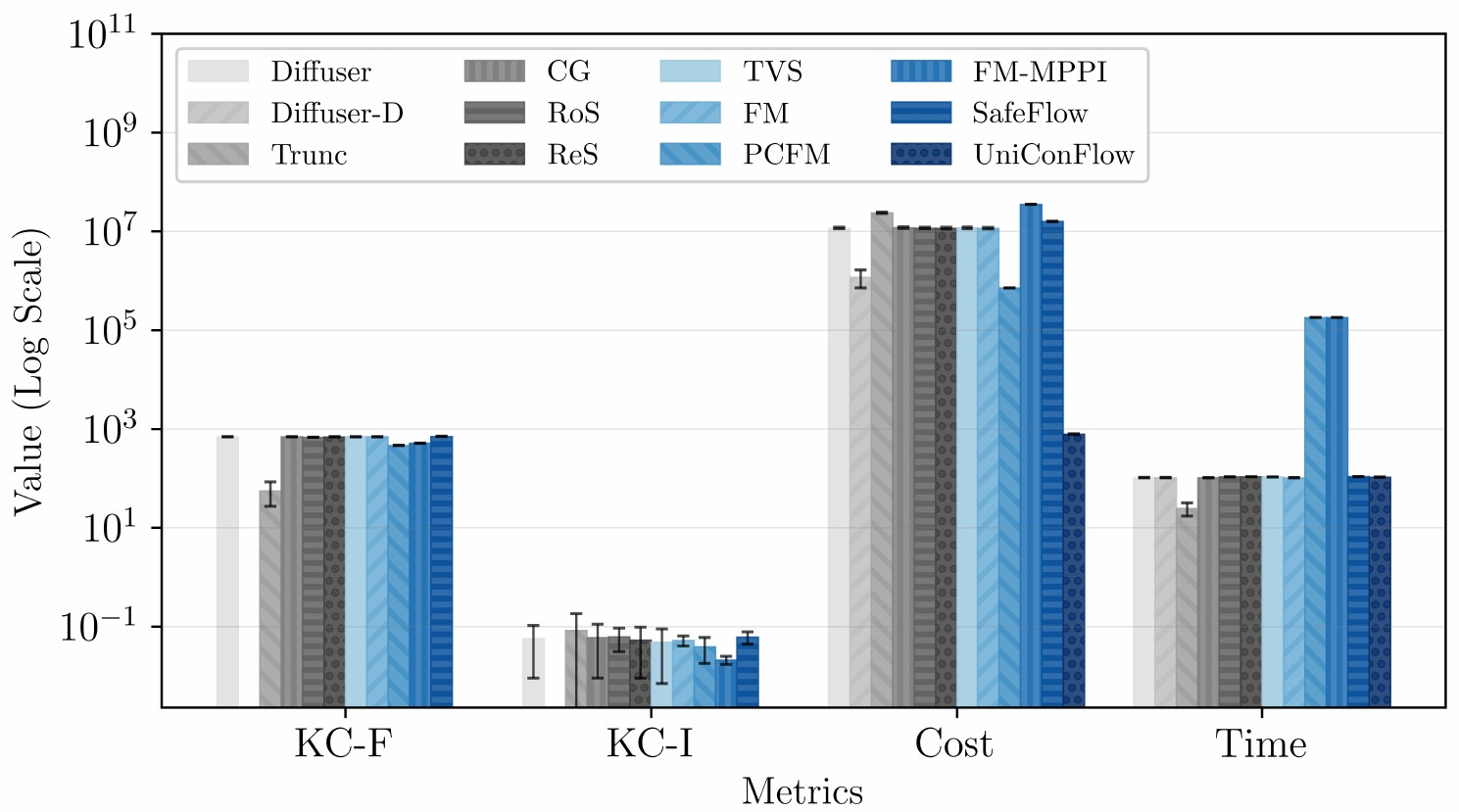}
		\caption{Values of the RMSE of kinodynamic consistency, cost, and inference time (ms) in 3D spatial motion scenario.}
		\vspace{-0.3cm}
		\label{fig_manipulator_3d_constraints_costs}
	\end{figure}
	In the high-dimensional manipulator pseudo-2D scenario (\cref{tab_arm_add_2D} in Appendix \ref{subsub_armain}), the results reveal a dramatic performance gap between UniConFlow and all baselines, with constraint violation serving as the critical differentiator. 
	UniConFlow is the only method achieving 100\% success across all success rate metrics, while SafeFlow achieves 100\% SR-S and AR due to the fact that the action sequences are not guided during the inference sampling.
	Moreover, UniConFlow achieves zero violations of kinodynamic consistency with a low cost, while SafeFlow suffers from massive violations with higher costs. 
	Vanilla models (Diffuser, FM) and guidance methods (CG, RoS, ReS, TVS, CoBL) all have high KC-F and KC-I values and costs exceeding $10^9\sim10^9$, indicating their generated trajectories are infeasible for the manipulator's physical constraints. 
	The optimization-based methods (PCFM, FM-MPPI) timeout entirely. 
	Therefore, iterative sampling-based approaches cannot apply to high-dimensional configuration spaces with complex obstacle geometries.

	The same observations hold consistently in the 3D spatial motion experiments, where \cref{fig_manipulator_3d_success_rates} demonstrates UniConFlow achieving 100\% across all four metrics, particularly excelling in SR-A and TSR, where all other methods except Trunc achieve zero success rates. 
	While most baselines maintain 100\% AR, they fail to generate executable action sequences that respect the manipulator's constraints. 
	The optimization-based methods (PCFM, FM-MPPI) exhibit even more failure, timing out ($>180$ s) before obtaining the solutions shown in \cref{fig_manipulator_3d_constraints_costs} and thus scoring 0\% on AR, demonstrating their inability to scale to the high-dimensional configuration space with complex 3D obstacle constraints. 
	Notably, in both scenarios (see \cref{tab_arm_add_2D,tab_arm_add_3D} in Appendix \ref{subsub_armain}), the cost of UniConFlow is $5\sim10$ orders of magnitude lower than any baseline while maintaining real-time performance ($32\sim 105$ ms), demonstrating that its unified constraint enforcement framework not only guarantees safety, admissibility, and kinodynamic consistency, but also generates the solutions for optimal and feasible trajectories.

	\subsection{Ablation Study}
	\label{subsec_as}
	To address the practical deployment concern raised regarding inference latency, we provide a detailed breakdown of the computational cost for each component of the UniConFlow inference pipeline across scenarios of increasing dimensionality. 
	In low-dimensional settings (pendulum and car racing), the total average inference time is approximately 2.5 ms, with the generative model ODE solver requiring 13 steps. 
	The full percentage breakdown for these cases is visualized in \cref{fig:computation_bar_chart} in Appendix~\ref{subsec_aas}.
	As task complexity grows, the pseudo-2D manipulator represents a medium-dimensional case requiring 133 ODE steps, while the 7-DoF 3D manipulator constitutes a high-dimensional case with 263 steps. The component-wise timing and proportional cost across all scenarios are reported in \cref{tab:computation_ablation}. 
	In all cases, the generative model ODE integration dominates the total inference budget (71–78\%), while the PTZF constraint evaluation and QP solving remain a modest fraction, scaling from 8.5\% in low-dimensional cases to 14.8 \% in the high-dimensional manipulator. 
	These results demonstrate that the QP and violation-detection stages do not constitute a bottleneck, and the total inference time of 105 ms for the 3D manipulator case remains within the range applicable to practical implementation. 
	
	To evaluate the true zero-shot efficacy of the flow-matching framework and quantify the role of terminal refinement, \cref{tab:ablation_terminal} decouples Stage 1 (PTZF output, without any post-processing) from Stage 2 (terminal refinement). 
	The PTZF-conditioned generative model already satisfies terminal constraints at high rates without correction: $88.4\%$, $98.1\%$, $89.2\%$, and $86.1\%$ for the pendulum, raceline, and 7-DoF manipulators, respectively, confirming that the PTZF meaningfully shapes the generative prior rather than deferring all safety enforcement downstream. 
	Notably, the Stage 1 state RMSE quantifies the pre-refinement kinodynamic consistency of the generated trajectories, while the mean per-step $\ell_2$ distance and the Fr\'echet distance quantify how much the terminal refinement changes the generated state and action trajectories, whose definitions and discussion are provided in Appendix~\ref{subsec_aas}.
	
	\begin{table*}
		\centering
		\caption{Computational Cost Ablation of Inference Components}
		\label{tab:computation_ablation}
		\setlength{\tabcolsep}{4pt}
		\renewcommand{\arraystretch}{1.2}
		\begin{tabular}{lcccccccc}
			\toprule
			\textbf{Pipeline Component} & \multicolumn{2}{c}{\makecell{\textbf{Pendulum}}} & \multicolumn{2}{c}{\makecell{\textbf{Raceline}}} & \multicolumn{2}{c}{\makecell{\textbf{Pseudo-2D Manip.}}} & \multicolumn{2}{c}{\makecell{\textbf{3D Manip.}}} \\
			\cmidrule(lr){2-3} \cmidrule(lr){4-5} \cmidrule(lr){6-7} \cmidrule(lr){8-9}
			& \textit{Time(ms)} & \textit{(\%)} & \textit{Time(ms)} & \textit{(\%)} & \textit{Time(ms)} & \textit{(\%)} & \textit{Time(ms)} & \textit{(\%)} \\
			\midrule
			Generative Model ODE      & 1.95 & 78.0 & 2.00 & 76.9 & 22.50 & 71.0 & 81.50 & 77.6 \\
			PTZF Constraints \& QP    & 0.35 & 14.0 & 0.22 & 8.5  & 3.20  & 10.1 & 15.50 & 14.8 \\
			Terminal Refinement       & 0.20 & 8.0  & 0.38 & 14.6 & 6.00  & 18.9 & 8.00  & 7.6  \\
			\midrule
			{Total Inference}  & {2.50} & {100} & {2.60} & {100} & {31.70} & {100} & {105.00} & {100} \\
			\bottomrule
		\end{tabular}
		\vspace{-0.3cm}
	\end{table*}

	\begin{table*}
		\centering
		\caption{Inference Components Impact Ablation of Terminal Enforcement on Generative Output}
		\label{tab:ablation_terminal}
		\setlength{\tabcolsep}{3pt}
		\renewcommand{\arraystretch}{1.1}
		\begin{tabular}{lcccccccc}
			\toprule
			\textbf{Task}
			& \multicolumn{2}{c}{{Stage 1 (PTZF Output)}}
			& \multicolumn{6}{c}{{Stage 2 (Terminal Refinement)}} \\
			\cmidrule(lr){2-3} \cmidrule(lr){4-9}
			& \textbf{Certified Rate}
			& \textbf{State RMSE}
			& \makecell{\textbf{State Corr.} \textbf{$\ell_2$ Dist.}}
			& \makecell{\textbf{State Fr\'echet} \textbf{Distance}}
			& \multicolumn{2}{c}{\makecell{\textbf{Action Corr.} \textbf{$\ell_2$ Dist.}}}
			& \multicolumn{2}{c}{\makecell{\textbf{Action Fr\'echet} \textbf{Distance}}} \\
			\midrule
			{Pendulum}
			& $88.4\%$ & $0.03$ & $0.07$ & $0.38$
			& \multicolumn{2}{c}{$0.31$}
			& \multicolumn{2}{c}{$1.13$} \\
			
			{Raceline}
			& $98.1\%$ & $0.07$ & $0.06$ & $0.52$
			& $0.31$  & $0.003$
			& $0.59$  & $0.009$ \\
			
			{Pseudo-2D Manip.}
			& $89.2\%$ & $29.84$ & $7.12$ & $13.49$
			& \multicolumn{2}{c}{$4.18$}
			& \multicolumn{2}{c}{$9.21$} \\
			
			{3D Manipulator}
			& $86.1\%$ & $17.18$ & $12.01$ & $15.79$
			& \multicolumn{2}{c}{$6.73$}
			& \multicolumn{2}{c}{$11.17$} \\
			\bottomrule
		\end{tabular}
		\vspace{-0.3cm}
	\end{table*}

	\section{Conclusion}
	\label{sec_conclusion}
	This paper introduced UniConFlow, a novel unified FM-based framework for motion generation that seamlessly integrates equality and inequality constraints without retraining, ensuring both safety, admissibility, and kinodynamic consistency for certified motion planning tasks. 
	By leveraging the proposed prescribed-time zeroing functions and control theory, UniConFlow achieves training-free constraint satisfaction and eliminates the need for auxiliary controllers, offering a flexible and efficient solution for dynamic tasks. 
	Experimental evaluations on three experiments, a double inverted pendulum, a nonholonomic vehicle, and the Franka manipulator, demonstrate that UniConFlow consistently outperforms SOTA generative model-based methods. 
	UniConFlow achieves perfect safety in position and action rollouts, as well as zero RMSE in kinodynamic consistency, indicating strict adherence to both safety and dynamic constraints. 
	Furthermore, compared to conventional optimization-based approaches, UniConFlow significantly reduces computation time, demonstrating feasibility for real-time deployment in practical robotic scenarios.


	\bibliography{ref_NoURL}
	\bibliographystyle{IEEEtran}
	
	\def\BiographyDistance{-10pt}
	\vspace{-11pt}
	
	\begin{IEEEbiography}[{\includegraphics[width=1in,height=1.25in,clip,keepaspectratio]{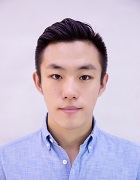}}]{Zewen Yang}
		(Member, IEEE)
		received the M.S. degree in control engineering from Northeast Forest University, in 2017. He pursued a Ph.D. in control science and engineering at College of Intelligent Systems Science and Engineering, Harbin Engineering University, Harbin, China, from 2017 to 2019 and joined the Chair of Information-oriented Control, School of Computation, Information and Technology, the Technical University of Munich, Munich, Germany, in machine learning and data-driven control until 2023. From 2023 to 2024, he was a postdoctoral researcher at the Robert Koch Institute, Berlin, Germany. Since 2025, he has been with the Chair of Robotics and Systems Intelligence at the Munich Institute of Robotics and Machine Intelligence, TUM. His current research interests include multi-agent systems, cooperative learning, generative models, control theory, and general robotics.
	\end{IEEEbiography}
	\vspace{\BiographyDistance}
	
	\begin{IEEEbiography}[{\includegraphics[width=1in,height=1.25in,clip,keepaspectratio]{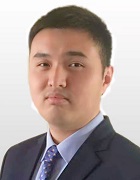}}]{Xiaobing Dai}
		received the B.Sc. mechanical engineering from the Tongji University, Shanghai, China, in 2018 with direction in mechatronics, building environment and civil engineering. He received double M.Sc degrees in Mechanical Engineering, Mechatronics and Robotics from the Technical University of Munich, Munich, Germany, in 2021. Since February 2022, he is a PhD student at the Chair of Information-oriented Control, TUM School of Computation, Information and Technology at the Technical University of Munich, Munich, Germany. His current research interests include efficient online machine learning, networked control systems, safe learning-based control.
	\end{IEEEbiography}
	\vspace{\BiographyDistance}
	
	\begin{IEEEbiography}[{\includegraphics[width=1in,height=1.25in,clip,keepaspectratio]{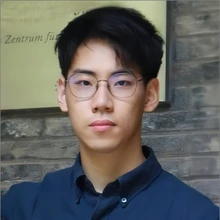}}]{Dian Yu}
		(Graduate Student Member, IEEE)
		received the B.Eng. degrees in Electrical Engineering and Automation from the University of Shanghai for Science and Technology (USST), Shanghai, China, and the Hamburg University of Applied Sciences (HAW Hamburg), Hamburg, Germany, and the M.Sc. degree in Robotics, Cognition, Intelligence from the Technical University of Munich (TUM), Munich, Germany, in 2026. He is currently a Research Associate and doctoral researcher with the Professorship of AI Planning in Dynamic Environments, Munich Institute of Robotics and Machine Intelligence (MIRMI), Technical University of Munich (TUM), Munich, Germany. His research interests include robot learning, motion planning, and control, with a focus on safe and constrained learning-based methods for robotic manipulation.
	\end{IEEEbiography}
	\vspace{\BiographyDistance}
	
	\vfill
	
	\begin{IEEEbiography}[{\includegraphics[width=1in,height=1.25in,clip,keepaspectratio]{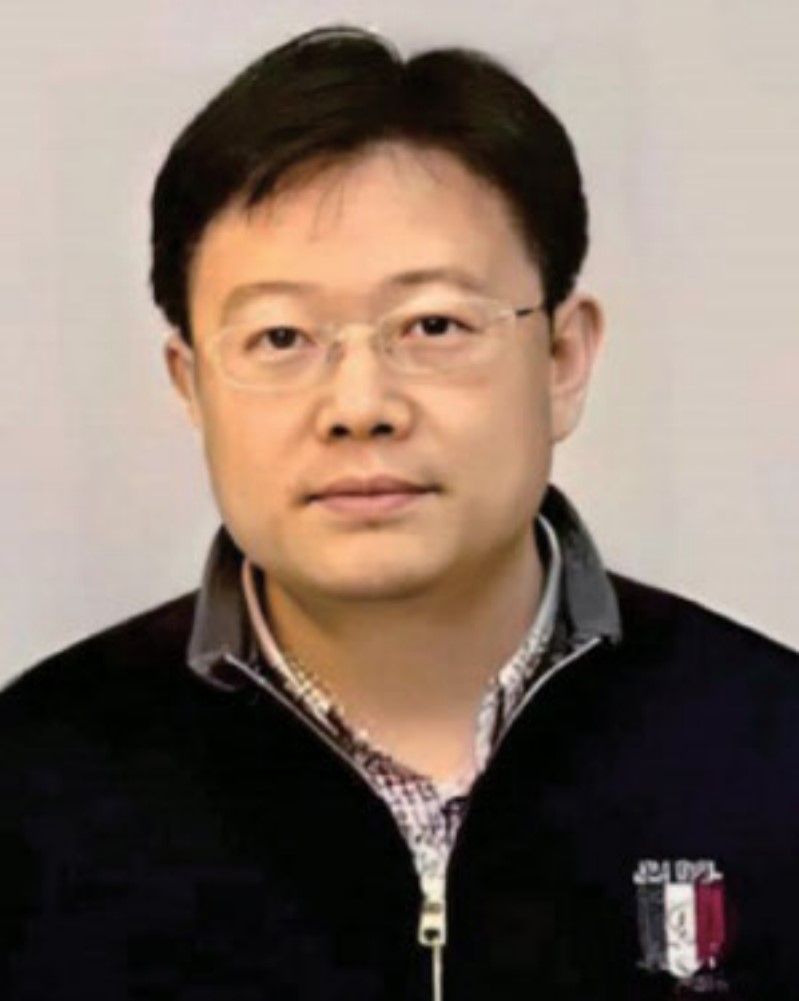}}]{Zhijun Li}
		(Fellow, IEEE)
		received the Ph.D. degree in mechatronics from Shanghai Jiao Tong University, Shanghai, China, in 2002.
		From 2003 to 2006, he was a Post-Doctoral Fellow with the University of Electro-Communications, Tokyo, Japan, and the National University of
		Singapore, Singapore. He is currently a Chair Professor with Tongji University, Shanghai, where he has been the Dean of the School of Mechanical Engineering since 2024. He has authored/co-authored more than 400 articles, where the prestigious contributions were wearable robotics and biomechatronics systems. His research interests include wearable robotics, biomechatronics systems, and intelligent control.
%
	\end{IEEEbiography}
	\vspace{\BiographyDistance}
	
	\begin{IEEEbiography}[{\includegraphics[width=1in,height=1.25in,clip,keepaspectratio]{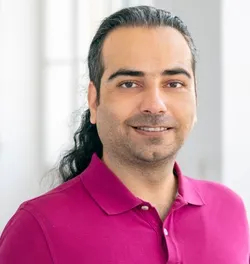}}]{Majid Khadiv}
		(Senior Member, IEEE)
		received the B.Sc. degree in mechanical engineering from Isfahan University of Technology (IUT), Isfahan, Iran, in 2010, and the M.Sc. and Ph.D. degrees in mechanical engineering from the K. N. Toosi University of Technology, Tehran, Iran, in 2012 and 2017, respectively. He is currently an Assistant Professor with the School of Computation, Information and Technology (CIT), Technical University of Munich (TUM), Munich, Germany. He is also a Principal Investigator with Munich Institute of Robotics and Machine Intelligence (MIRMI). Prior to joining TUM, he was a Postdoctoral Researcher at the Max Planck Institute for Intelligent Systems, Tübingen, Germany. His main research interests include the control of robotic systems, with a focus on locomotion and manipulation.
	\end{IEEEbiography}
	\vspace{\BiographyDistance}
	
	\begin{IEEEbiography}[{\includegraphics[width=1in,height=1.25in,clip,keepaspectratio]{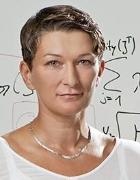}}]{Sandra Hirche}
		(Fellow, IEEE)
		received the Dipl.-Ing degree in aeronautical engineering from the Technical University of Berlin, Berlin, Germany, in 2002, and the Dr. Ing. degree in electrical engineering from the Technical University of Munich, Munich, Germany, in 2005. From 2005 to 2007, she was awarded a Post-doctoral scholarship from the Japanese Society for the Promotion of Science at the Fujita Laboratory, Tokyo Institute of Technology, Tokyo, Japan. From 2008 to 2012, she was an Associate Professor with the Technical University of Munich. Since 2013, she has served as Technical University of Munich Liesel Beckmann Distinguished Professor and has been with the Chair of Information-Oriented Control, Department of Electrical and Computer Engineering, Technical University of Munich. She has authored or coauthored more than 150 papers in international journals, books, and refereed conferences. Her main research interests include cooperative, distributed, and networked control with applications in human--machine interaction, multirobot systems, and general robotics. 
%
	\end{IEEEbiography}
	\vspace{\BiographyDistance}
	
	\begin{IEEEbiography}[{\includegraphics[width=1in,height=1.25in,clip,keepaspectratio]{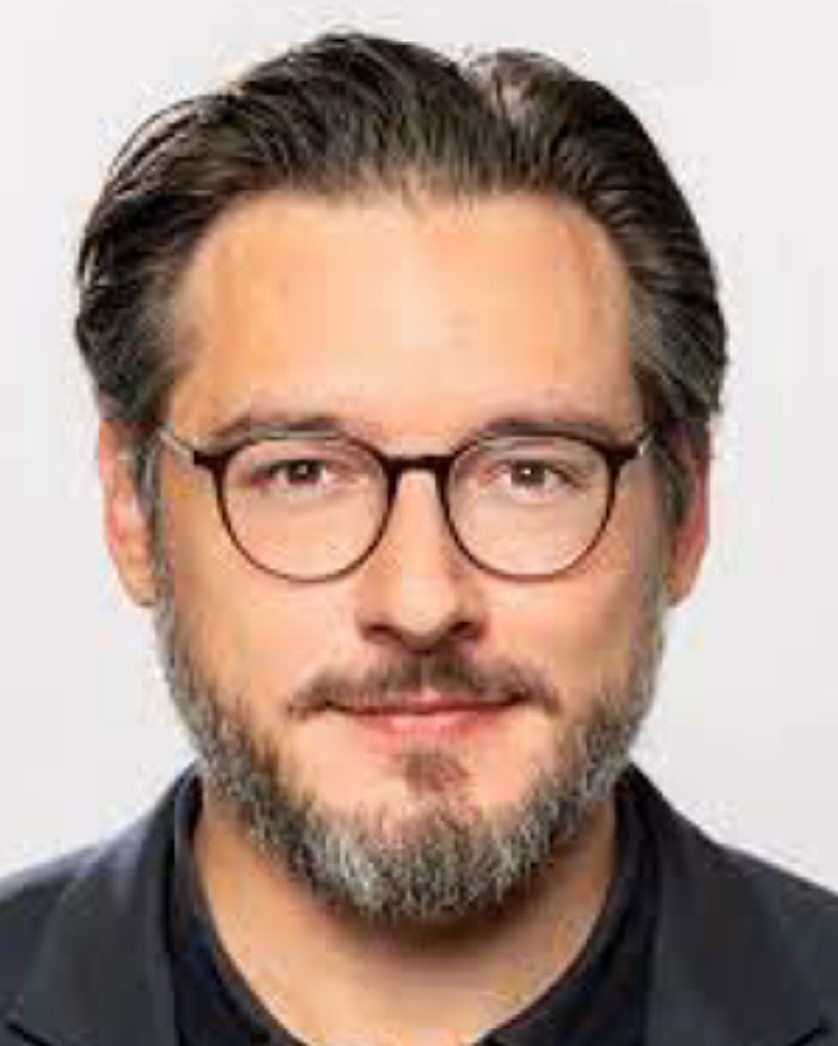}}]{Sami Haddadin}
		(Fellow, IEEE)
		received the Dipl.-Ing. degree in electrical engineering in 2005, the M.Sc. degree in computer science from the Technical University of Munich (TUM), Munich, Germany, in 2009, the dual Honours degree in technology management from Ludwig Maximilian University of Munich, Munich, and TUM, in 2007, and the Ph.D. degree in electrical engineering from RWTH Aachen University, Aachen, Germany, in 2011. He is currently a Full Professor and the Chair of the Robotics and Systems Intelligence, Technical University of Munich (TUM), the Founding Director of Munich Institute of Robotics and Machine Intelligence (MIRMI), and the Vice President of Research with the Mohamed Bin Zayed University of Artificial Intelligence, Abu Dhabi, United Arab Emirates.
	\end{IEEEbiography}
	
	\vfill
	
	\clearpage
	\appendices
	
	\startcontents[appendix] 
	\onecolumn
	\printcontents[appendix]{}{1}{}  
	\twocolumn  
	
	\clearpage
	\section{Proofs}
	\label{appendix_proof}
	
	\subsection{Proof of \cref{property_PTZF}}
	\label{proof_of_property_PTZF}
	
	The proof is separated into two parts with different horizon.
	For $t \in [c_r T_{\text{pre}}, T_{\text{pre}})$, define a new variable $\hat{t} \in \mathbb{R}$ as
	\begin{align}
		\hat{t} = t / (T_{\text{pre}} - t),
	\end{align}
	such that $t = T_{\text{pre}} \hat{t} / (1 + \hat{t})$.
	The derivatives of $t$ and $\hat{t}$ w.r.t $\hat{t}$ and $t$ are written as
	\begin{align}
		\frac{\mathrm{d} \hat{t}}{\mathrm{d} t} = \frac{T_{\text{pre}}}{(T_{\text{pre}} - t)^2}, && 
		\frac{\mathrm{d} t}{\mathrm{d} \hat{t}} = \frac{T_{\text{pre}}}{(1 + \hat{t})^2},
	\end{align}
	such that the derivative of $r(\cdot)$ w.r.t $\hat{t}$ denotes
	\begin{align}
		\frac{\mathrm{d} r}{\mathrm{d} \hat{t}} = \frac{\mathrm{d} r}{\mathrm{d} t} \frac{\mathrm{d} t}{\mathrm{d} \hat{t}} = - \frac{T_{\text{pre}}}{(T_{\text{pre}} - t)^2} \frac{T_{\text{pre}}}{(1 + \hat{t})^2} \gamma_r(t, r).
	\end{align}
	Considering $T_{\text{pre}} - t = T_{\text{pre}} / (1 + \hat{t})$, the derivative becomes
	\begin{align}
		\mathrm{d} r / \mathrm{d} \hat{t} = - \gamma_r(t, r) \le - \gamma_{r}(0, r),
	\end{align}
	which leads to $\lim_{\hat{t} \to \infty} r = 0$ according to \cite{khalil2015nonlinear}.
	Moreover, the definition of $t$ from $\hat{t}$ ensures $\lim_{\hat{t} \in \infty} t = T_{\text{pre}}$, indicating $\lim_{t \to T_{\text{pre}}} r = 0$ and $r(T_{\text{pre}}) = 0$ due to the continuity of $r$.
	After the prescribed time $T_{\text{pre}}$, the zero value of the derivative of $r$ in \eqref{eqn_system_r} guarantees $r(t) = r(T_{\text{pre}}) = 0$ for all $t \in [T_{\text{pre}}, \infty)$, which concludes the proof.		
	
	\subsection{Proof of \cref{theorem_equality_constraint}}
	\label{proof_theorem_equality_constraint}
	Define $\tilde{g}(t) = \bar{g}(t) - g(\bm{\mathcal{T}}_t)$, then the dynamical system \eqref{eqn_system_g} is written as
	\begin{align}
		\dot{\tilde{g}}(t) \ge - \gamma(\tilde{g}(t)), && \forall t \in [0, T_{\text{pre}}].
	\end{align}
	Moreover, due to the choice of $\bar{g}(0)$, it is easy to see $\tilde{g}(0) > 0$, which induces $\tilde{g}(t) \ge 0$ for all $t \in [0, T_{\text{pre}}]$ using comparison principle according to \cite{khalil2015nonlinear}.
	This observation is also equivalent to $g(\bm{\mathcal{T}}_1) \le \bar{g}(1) = 0$ following the result in \cref{property_PTZF}.
	Furthermore, considering the positivity of $g(\cdot)$ from its construction of $V(\cdot)$ in \eqref{eqn_g}, the zero level of $V(\bm{\mathcal{T}}_1)$ is guaranteed at the prescribed time $T_{\text{pre}} = 1$, leading to $g_i(\bm{\mathcal{T}}_1) = 0$ for all $i = 1, \cdots, N_g$ and concluding the proof.
	
	\subsection{Proof of \cref{theorem_inequality_constraint}}
	\label{proof_theorem_inequality_constraint}
	Define $\tilde{h}_j(t) = \bar{h}_j(t) - h_j(\bm{\mathcal{T}}_t)$, and the positivity of its initial value, i.e., $\tilde{h}_j(0) \ge 0$ is guaranteed due to the choice of $\bar{h}_j(0)$.
	Moreover, considering \eqref{eqn_system_r} and \eqref{eqn_system_hj}, the dynamical system of $\tilde{h}_j(\cdot)$ is written as
	\begin{align}
		\dot{\tilde{h}}_j(t) \ge - \gamma (\tilde{h}_j(t)), && \forall t \in \mathbb{R}_{\ge 0},
	\end{align}
	leading to the positivity of $\tilde{h}_j(t)$ using the comparison principle \cite{khalil2015nonlinear}.
	This observation is equivalent to $h_j(\bm{\mathcal{T}}_t) \le \bar{h}_j(t)$, and results to $h_j(\bm{\mathcal{T}}_1) \le \bar{h}_j(1) = 0$ according to \cref{property_PTZF} for PTZF $\bar{h}_j(\cdot)$, which concludes the proof.
	
	\subsection{Proof of \cref{theorem_ut_slack_variable}}
	\label{proof_theorem_ut_slack_variable}
	Due to the same structure of conditions in \eqref{eqn_system_g} and \eqref{eqn_system_hj} for equality and inequality constraints, this proof takes the equality constraint with $g(\cdot)$ as an example.
	Considering the slack variable $\delta_{g,t}$ obtained from \eqref{eqn_QP_closed_form_solution}, the dynamics of $g(\cdot)$ with slack variable $\delta_{g,t}$ is written as
	\begin{align} \label{eqn_system_g_slack_variable}
		\dot{\tilde{g}}(t) \ge - \gamma(\tilde{g}(t)) - \delta_{g,t}, && \forall t \in [0,1]
	\end{align}
	with $\tilde{g}(t) = \bar{g}(t) - g(\bm{\mathcal{T}}_t)$.
	Moreover, due to the property of $\gamma(\cdot)$ as $\gamma(a(t)) \le c_{\gamma} \varphi(t) a(t)$, the dynamics in \eqref{eqn_system_g_slack_variable} is becomes
	\begin{align}
		\dot{\tilde{g}}(t) \ge - c_{\gamma} \varphi(t) \tilde{g}(t) - \delta_{g,t},
	\end{align}
	whose solution following comparison principle as
	\begin{align}
		\tilde{g}(t) \ge& \mathrm{e}^{- c_{\gamma} \int_0^t \varphi(s) \mathrm{d} s} \tilde{g}(0) - \int_0^t \mathrm{e}^{- c_{\gamma} \int_s^t \varphi(\tau) \mathrm{d} \tau} \delta_{g,s} \mathrm{d} s \\
		\ge& \mathrm{e}^{- c_{\gamma} \int_0^t \varphi(s) \mathrm{d} s} \tilde{g}(0) - \bar{\delta}_g \int_0^t \mathrm{e}^{- c_{\gamma} \int_s^t \varphi(\tau) \mathrm{d} \tau} \mathrm{d} s \nonumber
	\end{align}
	due to the bounded $\delta_{g,t}$ as $\bar{\delta}_g$.
	Furthermore, the integration $\int_0^t \mathrm{e}^{- c_{\gamma} \int_s^t \varphi(\tau) \mathrm{d} \tau} \mathrm{d} s$ is further bounded by
	\begin{align}
		&\int_0^t \mathrm{e}^{- c_{\gamma} \int_s^t \varphi(\tau) \mathrm{d} \tau} \mathrm{d} s \\
		=& \lim_{\epsilon \to 0} \Big( \int_0^{t - \epsilon} \mathrm{e}^{- c_{\gamma} \int_s^t \varphi(\tau) \mathrm{d} \tau} \mathrm{d} s + \int_{t - \epsilon}^t \mathrm{e}^{- c_{\gamma} \int_s^t \varphi(\tau) \mathrm{d} \tau} \mathrm{d} s \Big) \nonumber \\
		\le& \lim_{\epsilon \to 0} \Big( \mathrm{e}^{- c_{\gamma} \int_{t - \epsilon}^t \varphi(\tau) \mathrm{d} \tau} \int_0^{t - \epsilon} 1 \mathrm{d} s + \int_{t - \epsilon}^t 1 \mathrm{d} s \Big) \nonumber \\
		=& \lim_{\epsilon \to 0} \Big( \mathrm{e}^{- c_{\gamma} \int_{t - \epsilon}^t \varphi(\tau) \mathrm{d} \tau} (t - \epsilon) + \epsilon \Big) = t \mathrm{e}^{- c_{\gamma} \varphi(t) }, \nonumber
	\end{align}
	such that the lower bound of $\tilde{g}(t)$ is further expressed as
	\begin{align}
		\tilde{g}(t) \ge& \mathrm{e}^{- c_{\gamma} \int_0^t \varphi(s) \mathrm{d} s} \tilde{g}(0) - \bar{\delta}_g t \mathrm{e}^{- c_{\gamma} \varphi(t) }
	\end{align}
	for all $t \in [0,1)$.
	Additionally, due to the properties of blow-up function $\varphi(\cdot)$ in the theorem, it is obvious to see $\mathrm{e}^{- c_{\gamma} \int_0^{1^-} \varphi(s) \mathrm{d} s} = 0$ as well as $\mathrm{e}^{- c_{\gamma} \varphi(1^-) } = 0$, which induces $\lim_{t \to 1^-} \tilde{g}(t) \ge 0$ and $\tilde{g}(1) \ge 0$ considering the continuity of $\tilde{g}(\cdot)$.
	This observation also indicates $g(\bm{\mathcal{T}}_1) \le \bar{g}(1) = 0$ since $\bar{g}$ is a PTZF satisfying \cref{property_PTZF}, which results the satisfaction of equality constraints $g_i(\bm{\mathcal{T}}_1) = 0$ from $g(\bm{\mathcal{T}}_1) = 0$ due to the zero lower bound of $g(\cdot)$ from its construction in \eqref{eqn_g}.

	Following the same procedure as equality constraint with $g(\cdot)$, the result $\tilde{h}_j(1) = \bar{h}_j(1) - h_j(\bm{\mathcal{T}}_1) \ge 0$ is guaranteed with $j = 1, \cdots, N_h$.
	This observation leads to $h_j(\bm{\mathcal{T}}_1) \le \bar{h}_j(1) = 0$, which indicates the satisfaction of inequality constraint in \eqref{eqn_constrained_generation} and therefore concludes the proof.
	
	\subsection{Proof of \cref{theorem_free_generation_condition}}
	\label{proof_theorem_free_generation_condition}	
	The zero optimal guidance input $\bm{u}_t$ indicates the unconstrained flow $\bm{v}_t^{\bm{\theta}}(\bm{\mathcal{T}})$ satisfies all constraints in \eqref{eqn_constrained_optimization} with $t \in [0, c_r]$.
	First, the equality constraint with $g(\cdot)$ is investigated, whose derivative without guidance input following Cauchy-Schwarz inequality is bounded by
	\begin{align} \label{eqn_dot_g_upper_bound}
		\dot{g}(\bm{\mathcal{T}}_t) =& \bm{\eta}_g^T(t) \bm{v}_t^{\bm{\theta}}(\bm{\mathcal{T}}_t) \le \| \bm{\eta}_g(t) \| \| \bm{v}_t^{\bm{\theta}}(\bm{\mathcal{T}}_t) \| \le L_g \bar{v}^{\bm{\theta}}
	\end{align}
	considering the Lipschitz continuity of $g(\cdot)$ and bounded unconstrained flow $\bm{v}_t^{\bm{\theta}}(\cdot)$.
	This observation also induces bounded value of $g(\bm{\mathcal{T}}_t)$ as
	\begin{align}
		g(\bm{\mathcal{T}}_t) =& g(\bm{\mathcal{T}}_0) + \int_0^t \dot{g}(\bm{\mathcal{T}}_s) \mathrm{d} s \le g(\bm{\mathcal{T}}_0) + \int_0^t L_g \bar{v}^{\bm{\theta}} \mathrm{d} s \nonumber \\
		=& g(\bm{\mathcal{T}}_0) + L_g \bar{v}^{\bm{\theta}} t \le g(\bm{\mathcal{T}}_0) + L_g \bar{v}^{\bm{\theta}} c_r
	\end{align}
	considering the domain $t \in [0, c_r]$, such that the right-hand side in \eqref{eqn_system_g} following \eqref{eqn_RHS_g_bound} is further bounded by
	\begin{align}
		\gamma(\bar{g}(t) &- g(\bm{\mathcal{T}}_t)) + \dot{\bar{g}}(t) \\
		\ge& - \frac{c_{\text{PT}} g(\bm{\mathcal{T}}_0)}{(1 - c_r)^2} - \frac{c_{\text{PT}} L_g \bar{v}^{\bm{\theta}} c_r}{(1 - c_r)^2} + \frac{(c_{\text{PT}} - c_g) \bar{g}(0)}{\mathrm{e}^{c_g c_r / (1 - c_r)}} \nonumber \\
		=& \frac{c_{\text{PT}} - c_g}{\mathrm{e}^{c_g c_r / (1 - c_r)}} \big( \bar{g}(0) - \xi_{\mathcal{T},0} (g(\bm{\mathcal{T}}_0) - c_r L_g \bar{v}^{\bm{\theta}}) \big). \nonumber
	\end{align}
	Moreover, the choice of $\bar{g}(0)$ in \eqref{eqn_g_h_bar_initial_free_generation} ensures 
	\begin{align}
		\bar{g}(0) - \xi_{\mathcal{T},0} (g(\bm{\mathcal{T}}_0) - c_r L_g \bar{v}^{\bm{\theta}}) \ge \frac{\mathrm{e}^{c_g c_r / (1 - c_r)}}{c_{\text{PT}} - c_g} L_g \bar{v}^{\bm{\theta}},
	\end{align}
	leading to the lower bound for $\gamma(\bar{g}(t) - g(\bm{\mathcal{T}}_t)) + \dot{\bar{g}}(t)$ as
	\begin{align}
		\gamma(\bar{g}(t) - g(\bm{\mathcal{T}}_t)) + \dot{\bar{g}}(t) \ge L_g \bar{v}^{\bm{\theta}} \ge \dot{g}(\bm{\mathcal{T}}_t)
	\end{align}
	for all $t \in [0,c_r]$ using the result in \eqref{eqn_dot_g_upper_bound}.
	This observation the condition \eqref{eqn_system_g} holds with unconstrained flow $\bm{v}_t^{\bm{\theta}}(\bm{\mathcal{T}})$ in the early generation stage $[0,c_r]$.
	
	Following the similar procedure as equality constraint $g(\cdot)$, it is straightforwardly to see that the condition \eqref{eqn_system_hj} for $h_j(\cdot)$ holds with the initial value of its corresponding PTZF as $\bar{h}_j(0)$ in \eqref{eqn_g_h_bar_initial_free_generation} under unconstrained flow $\bm{v}_t^{\bm{\theta}}(\bm{\mathcal{T}})$ in $[0,c_r]$ for all $j = 1, \cdots, N_h$.
	Combining the discussion for equality and inequality constraints, it is concluded that all constraints are satisfied with zero guidance input in the early stage, i.e., $\bm{u}_t = \bm{0}_{d \times 1}$ for all $t \in [0, c_r]$, which completes the proof.
	
	\section{Certified Path Planning}
	\label{subsection_certified_path_planning}
	
	In some scenario, only the system states $\bm{s}^k$ for $k = 0, \cdots, H$ are recorded, forming the state of flow as
	\begin{align}
		\bm{\mathcal{T}}^s = [(\bm{s}^0)^T, \cdots, (\bm{s}^H)^T]^T
	\end{align}
	with dimension $d_{\mathcal{T}}^s = (H + 1) d_s$.
	The dynamics of $\bm{\mathcal{T}}^s$ following controlled flow matching framework \eqref{eqn_controlled_flow} denotes
	\begin{align}
		\dot{\bm{\mathcal{T}}}^s_t = \bm{v}^{s,\bm{\theta}^s}_t(\bm{\mathcal{T}}^s_t) + \bm{u}^s_t
	\end{align}
	where $\bm{u}^s_t$ is the guidance term and $\bm{v}^{s,\bm{\theta}^s}_t(\cdot): \mathbb{R}^{d_{\mathcal{T}}^s} \to \mathbb{R}^{d_{\mathcal{T}}^s}$ is a parametric model for the flow with learned parameter $\bm{\theta}^s \in \mathbb{R}^{d_{\theta}^s}$ and $d_{\theta}^s \in \mathbb{N}$.
	To consider the dynamics constraint in \eqref{eqn_dynamics_consistency} and action feasibility in \eqref{eqn_action_domain}, a virtual action flow is introduced with flow state $\bm{\mathcal{T}}^a$ defined as
	\begin{align}
		\bm{\mathcal{T}}^a = [(\bm{a}^0)^T, \cdots, (\bm{a}^{(H-1)})^T]^T
	\end{align}
	with dimension $d_{\mathcal{T}}^a = H d_a$.
	Considering the lack of action data, the distribution of $\bm{\mathcal{T}}^a$ draws less attention and requires no model training.
	Instead, only the action feasibility and the dynamics constraint are essential, such that the dynamics of $\bm{\mathcal{T}}^a$ is only driven by a correction term $\bm{u}^a \in \mathbb{R}^{d_{\mathcal{T}}^a}$ as
	\begin{align} \label{eqn_controlled_virtual_action_flow}
		\dot{\bm{\mathcal{T}}}^a_t = \bm{u}^a_t
	\end{align}
	for all $t \in [0,1]$.
	It is easy to see the trajectory $\bm{\mathcal{T}}$ is a reordering of flow state $\bm{\mathcal{T}}^s$ and virtual action flow state $\bm{\mathcal{T}}^a$, which is simplified as $\bm{\mathcal{T}} = \bm{\mathcal{F}}_p(\bm{\mathcal{T}}^s, \bm{\mathcal{T}}^a)$ with known $\bm{\mathcal{F}}_p(\cdot, \cdot): \mathbb{R}^{d_{\mathcal{T}}^s} \times \mathbb{R}^{d_{\mathcal{T}}^a} \to \mathbb{R}^d$.
	The combined equality constraint $g(\cdot)$ denotes
	\begin{align}
		g(\bm{\mathcal{T}}^s_t, \bm{\mathcal{T}}^a_t) =& g_{\text{ini}}^2(\bm{\mathcal{F}}_p(\bm{\mathcal{T}}^s_t, \bm{\mathcal{T}}^a_t)) \\
		&+ \sum\nolimits_{k = 0}^{H - 1} (g^k_{\text{con}}(\bm{\mathcal{F}}_p(\bm{\mathcal{T}}^s_t, \bm{\mathcal{T}}^a_t)))^2, \nonumber
	\end{align}
	considering kinodynamical consistency in \cref{subsubsection_dynamical_consistency} and initial condition in \cref{subsubsection_initial_condition}.
	The state and action constraints in \cref{subsubsection_state_constraint} and \cref{subsubsection_action_constraint} are reordered as
	\begin{align}
		h_j(\bm{\mathcal{T}}^s_t, \bm{\mathcal{T}}^a_t) = h_s^{j - 1}(\bm{\mathcal{F}}_p(\bm{\mathcal{T}}^s_t, \bm{\mathcal{T}}^a_t))
	\end{align}
	for all $j = 1, \cdots, H + 1$ and
	\begin{align}
		h_j(\bm{\mathcal{T}}^s_t, \bm{\mathcal{T}}^a_t) = h_a^{j - H - 2}(\bm{\mathcal{F}}_p(\bm{\mathcal{T}}^s_t, \bm{\mathcal{T}}^a_t))
	\end{align}
	for all $j = H + 2, \cdots, 2 H + 1$. 
	To construct QP problem for obtaining $\bm{u}$ as in \eqref{eqn_QP}, it is essential to represent the derivative of trajectory $\bm{\mathcal{T}}$ and a function of $\bm{\mathcal{T}}^s$, $\bm{\mathcal{T}}^a$ and $\bm{u}^s$, $\bm{u}^a$, which is 
	
	written as
	\begin{align} \label{eqn_relationship_P_T_path_planning}
		\dot{\bm{\mathcal{T}}}_t =& \bm{J}_{\mathcal{F},p}^s(\bm{\mathcal{T}}^s_t, \bm{\mathcal{T}}^a_t) \dot{\bm{\mathcal{T}}}^s_t + \bm{J}_{\mathcal{F},p}^a(\bm{\mathcal{T}}^s_t, \bm{\mathcal{T}}^a_t) \dot{\bm{\mathcal{T}}}^a_t \\
		=& \bm{J}_{\mathcal{F},p}^s(\bm{\mathcal{T}}^s_t, \bm{\mathcal{T}}^a_t) (\bm{v}_t^{s,\bm{\theta}^s}(\bm{\mathcal{T}}^s_t) + \bm{u}^s_t) + \bm{J}_{\mathcal{F},p}^a(\bm{\mathcal{T}}^s_t, \bm{\mathcal{T}}^a_t) \bm{u}^a_t, \nonumber
	\end{align}
	following the constrained flow \eqref{eqn_controlled_flow} and \eqref{eqn_controlled_virtual_action_flow}.
	The Jacobian matrix $\bm{J}_{\mathcal{F},p}^s(\bm{\mathcal{T}}^s_t, \bm{\mathcal{T}}^a_t) = ( \mathrm{d} \bm{\mathcal{T}}_t / \mathrm{d} \bm{\mathcal{T}}^s_t )^T$ is explicitly written as 
	\begin{align}
		\bm{J}_{\mathcal{F},p}^s \!=\! \begin{bmatrix}
			\bm{J}^{s,0}_{p,s,0} & \bm{J}^{s,0}_{p,s,1} & \cdots &\bm{J}^{s,0}_{p,s,H} \\
			\bm{J}^{a,0}_{p,s,0} & \bm{J}^{a,0}_{p,s,1} & \cdots & \bm{J}^{a,0}_{p,s,H} \\
			\vdots & \ddots & \vdots \\
			\bm{J}^{s,H}_{p,s,0} & \bm{J}^{s,H}_{p,s,1} & \cdots & \bm{J}^{s,H}_{p,s,H}
		\end{bmatrix},
	\end{align}
	where $\bm{J}^{s,q}_{p,s,k} = \delta(q,k) \bm{I}_{d_s}$ for all $q, k = 0, \cdots, H$ with a Kronecker delta function $\delta(\cdot,\cdot): \mathbb{N} \times \mathbb{N} \to \{0, 1\}$ defined as
	\begin{align} \label{eqn_delta_function}
		\delta(p,q) = \begin{cases}
			1, & \text{if~} p = q \\
			0, & \text{otherwise}
		\end{cases}
	\end{align}
	and $\bm{J}^{a,q}_{p,s,k} = \bm{0}_{d_a \times d_s}$ for all $q = 0, \cdots, H-1$ and $k = 0, \cdots, H$.
	The Jacobian matrix $\bm{J}_{\mathcal{F},p}^a(\bm{\mathcal{T}}^s_t, \bm{\mathcal{T}}^a_t) = ( \mathrm{d} \bm{\mathcal{T}}_t / \mathrm{d} \bm{\mathcal{T}}^a_t )^T$ related to the virtual action is expressed in detail as
	\begin{align}
		\bm{J}_{\mathcal{F},p}^a \!=\! \begin{bmatrix}
			\bm{J}^{s,0}_{p,a,0} & \bm{J}^{s,0}_{p,a,1} & \cdots &\bm{J}^{s,0}_{p,a,H-1} \\
			\bm{J}^{a,0}_{p,a,0} & \bm{J}^{a,0}_{p,a,1} & \cdots & \bm{J}^{a,0}_{p,a,H-1} \\
			\vdots & \ddots & \vdots \\
			\bm{J}^{s,H}_{p,a,0} & \bm{J}^{s,H}_{p,a,1} & \cdots & \bm{J}^{s,H}_{p,a,H-1}
		\end{bmatrix},
	\end{align}
	where $\bm{J}_{a,q}^{p,a,k} = \delta(q,k) \bm{I}_{d_a}$ for all $q, k = 0, \cdots, H-1$  and $\bm{J}^{s,q}_{p,a,k} = \bm{0}_{d_s \times d_a}$ for all $q = 0, \cdots, H$ and $k = 0, \cdots, H-1$.
	
	With the derived relationship in \eqref{eqn_relationship_P_T_path_planning}, the derivative of combined equality constraint $g(\cdot)$ is written as
	\begin{align}
		\dot{g}(\bm{\mathcal{T}}^s_t,\! \bm{\mathcal{T}}^a_t) \!=& \big( \bm{\eta}_{\text{ini}}(\bm{\mathcal{F}}_p(\bm{\mathcal{T}}^s_t,\! \bm{\mathcal{T}}^a_t)) \!+\! \sum\nolimits_{k \!=\! 0}^{H \!-\! 1} \! \bm{\eta}^k_{\text{con}}(\bm{\mathcal{F}}_p(\bm{\mathcal{T}}^s_t,\! \bm{\mathcal{T}}^a_t)) \big)^T \nonumber \\
		&\times\! (\bm{J}_{\mathcal{F},p}^s(\bm{\mathcal{T}}^s_t,\! \bm{\mathcal{T}}^a_t) \dot{\bm{\mathcal{T}}}^s_t \!\!+\!\! \bm{J}_{\mathcal{F},p}^a(\bm{\mathcal{T}}^s_t,\! \bm{\mathcal{T}}^a_t) \dot{\bm{\mathcal{T}}}^a_t) \\
		=& \bm{\eta}_{g,v}^T(\bm{\mathcal{T}}^s_t, \bm{\mathcal{T}}^a_t) \bm{v}_t^{s, \bm{\theta}^s}(\bm{\mathcal{T}}^s_t) + \bm{\eta}_{g,u}^T(\bm{\mathcal{T}}^s_t, \bm{\mathcal{T}}^a_t) \bm{u}^{aug}_t, \nonumber
	\end{align}
	where the augmented guidance is expressed as $\bm{u}^{aug}_t = [(\bm{u}^s_t)^T, (\bm{u}^a_t)^T]^T$ and
	\begin{align}
		\bm{\eta}_{g,v}(\bm{\mathcal{T}}^s_t, \bm{\mathcal{T}}^a_t) \!=& (\bm{J}_{\mathcal{F},p}^s(\bm{\mathcal{T}}^s_t,\! \bm{\mathcal{T}}^a_t))^T \bm{\eta}_{g,v}^{base}(\bm{\mathcal{T}}^s_t,\! \bm{\mathcal{T}}^a_t), \\
		\bm{\eta}_{g,u}^T(\bm{\mathcal{T}}^s_t, \bm{\mathcal{T}}^a_t) \!=& [\bm{J}_{\mathcal{F},p}^s(\bm{\mathcal{T}}^s_t,\! \bm{\mathcal{T}}^a_t),\! \bm{J}_{\mathcal{F},p}^a(\bm{\mathcal{T}}^s_t, \bm{\mathcal{T}}^a_t)]^T \bm{\eta}_{g,v}^{base}(\bm{\mathcal{T}}^s_t,\! \bm{\mathcal{T}}^a_t) \nonumber
	\end{align}
	with the base vector $\bm{\eta}_{g,v}^{base}(\bm{\mathcal{T}}^s_t, \bm{\mathcal{T}}^a_t)$ defined as
	\begin{align}
		\bm{\eta}_{g,v}^{base}(\bm{\mathcal{T}}^s_t, \bm{\mathcal{T}}^a_t) =& \bm{\eta}_{\text{ini}}(\bm{\mathcal{F}}_p(\bm{\mathcal{T}}^s_t, \bm{\mathcal{T}}^a_t)) \\
		&+ \sum\nolimits_{k = 0}^{H - 1} \bm{\eta}^k_{\text{con}}(\bm{\mathcal{F}}_p(\bm{\mathcal{T}}^s_t, \bm{\mathcal{T}}^a_t)). \nonumber
	\end{align}
	Moreover, the derivative of $h_j(\cdot)$ with $j = 1, \cdots, H + 1$ for state constraint is written as
	\begin{align} \label{eqn_dot_h_path_planning}
		\dot{h}_j(\bm{\mathcal{T}}^s_t, \bm{\mathcal{T}}^a_t) =& \bm{\eta}_{h,v,j}^T(\bm{\mathcal{T}}^s_t, \bm{\mathcal{T}}^a_t) \bm{v}_t^{s,\bm{\theta}^s}(\bm{\mathcal{T}}^s_t) \\
		&+ \bm{\eta}_{h,u,j}^T(\bm{\mathcal{T}}^s_t, \bm{\mathcal{T}}^a_t) \bm{u}^{aug}_t, \nonumber
	\end{align}
	where $\bm{\eta}_{h,v,j}(\bm{\mathcal{F}}_p(\bm{\mathcal{T}}^s_t, \bm{\mathcal{T}}^a_t))$ and $\bm{\eta}_{h,u,j}(\bm{\mathcal{F}}_p(\bm{\mathcal{T}}^s_t, \bm{\mathcal{T}}^a_t))$ denote
	\begin{align}
		\bm{\eta}_{h,v,j}(\bm{\mathcal{T}}^s_t,\! \bm{\mathcal{T}}^a_t) \!=& (\bm{J}_{\mathcal{F},p}^s(\bm{\mathcal{T}}^s_t,\! \bm{\mathcal{T}}^a_t))^T \bm{\eta}^{j \!-\! 1}_s(\bm{\mathcal{F}}_p(\bm{\mathcal{T}}^s_t,\! \bm{\mathcal{T}}^a_t)), \nonumber \\
		\bm{\eta}_{h,u,j}(\bm{\mathcal{T}}^s_t,\! \bm{\mathcal{T}}^a_t) \!=& [\bm{J}_{\mathcal{F},p}^s(\bm{\mathcal{T}}^s_t,\! \bm{\mathcal{T}}^a_t),\! \bm{J}_{\mathcal{F},p}^a(\bm{\mathcal{T}}^s_t, \bm{\mathcal{T}}^a_t)]^T \\
		&\times \bm{\eta}^{j \!-\! 1}_s(\bm{\mathcal{F}}_p(\bm{\mathcal{T}}^s_t,\! \bm{\mathcal{T}}^a_t)). \nonumber
	\end{align}
	Similarly, the derivative of $h_j(\cdot)$ for action feasibility has the identical form as \eqref{eqn_dot_h_path_planning} but with
	\begin{align}
		\bm{\eta}_{h,v,j}(\bm{\mathcal{T}}^s_t,\! \bm{\mathcal{T}}^a_t) \!=& (\bm{J}_{\mathcal{F},p}^s(\bm{\mathcal{T}}^s_t,\! \bm{\mathcal{T}}^a_t))^T \bm{\eta}^{j - H - 2}_a(\bm{\mathcal{F}}_p(\bm{\mathcal{T}}^s_t,\! \bm{\mathcal{T}}^a_t)), \nonumber \\
		\bm{\eta}_{h,u,j}(\bm{\mathcal{T}}^s_t,\! \bm{\mathcal{T}}^a_t) \!=& [\bm{J}_{\mathcal{F},p}^s(\bm{\mathcal{T}}^s_t,\! \bm{\mathcal{T}}^a_t),\! \bm{J}_{\mathcal{F},p}^a(\bm{\mathcal{T}}^s_t, \bm{\mathcal{T}}^a_t)]^T \\
		&\times \bm{\eta}^{j - H - 2}_a(\bm{\mathcal{F}}_p(\bm{\mathcal{T}}^s_t,\! \bm{\mathcal{T}}^a_t)) \nonumber
	\end{align}
	with $j = H + 2, \cdots, 2 H + 1$.
	Therefore, the generation problem for certified path planning is formulated as a constrained optimization problem w.r.t augmented guidance term $\bm{u}^{aug}_t$, which is solved by UniConFlow framework.
	Since small $\bm{u}_t$ is required to minimize the difference between the distribution from the training data set $\mathbb{D}_{\mathcal{T}}$ and the safe distribution, the objective function in \eqref{eqn_constrained_optimization} can be designed as
	\begin{align}
		\min\nolimits_{\bm{u}^{aug}_t \in \mathbb{R}^d} (\bm{u}^{aug}_t)^T \bm{P}^{aug}_{u,t} \bm{u}^{aug}_t
	\end{align}
	with $\bm{P}^{aug}_{u,t} = \mathrm{blkdiag}(\bm{P}^s_{u,t}, \bm{P}^a_{u,t})$, where $\bm{P}^s_{u,t} \succ \bm{P}^a_{u,t}$ for bigger penalty on $\bm{u}^s_t$, allowing relatively larger $\bm{u}^a_t$.
	Denote $\bm{u}^{aug}_t$ as the optimal solution from UniConFlow, then
	\begin{align}
		\bm{u}^s_t = [\bm{I}_d, \bm{0}_{d \times d}] \bm{u}^{aug}_t, &&
		\bm{u}^a_t = [\bm{0}_{d \times d}, \bm{I}_{d}] \bm{u}^{aug}_t.
	\end{align}
	are the guidance terms for constrained flow \eqref{eqn_controlled_flow} and virtual action flow \eqref{eqn_controlled_virtual_action_flow}, respectively.

	\section{Details of Experiment and Additional Results}
	\label{sec_addiditonal_Experiment_Results}
	
	\subsection{Double Inverted Pendulum}
	\subsubsection{Explicit Model of Inverted Pendulum}
	\label{subsub_ExplicitModelofInvertedPendulum}
	The equations of motion are derived using the Euler-Lagrange formulation, resulting in the second-order dynamics
	\begin{equation}
		\label{eq_second_order_dynamics}
		\boldsymbol{M}(\boldsymbol{q}) \ddot{\boldsymbol{q}} + \boldsymbol{C}(\boldsymbol{q}, \dot{\boldsymbol{q}}) + \boldsymbol{g}(\boldsymbol{q}) = \boldsymbol{a},
	\end{equation}
	where \(\boldsymbol{q} = [q_1, q_2]^{\top}\) is the vector of generalized coordinates, and \(\boldsymbol{M}(q) \in \mathbb{R}^{2 \times 2}\) is the symmetric, positive-definite inertia matrix defined as 
	\begin{align*}
		\boldsymbol{M}(\boldsymbol{q})  = \begin{bmatrix}
			(m_1 + m_2) l_1^2 & m_2 l_1 l_2 \cos(q_2 - q_1) \\
			m_2 l_1 l_2 \cos(q_2 - q_1) & m_2 l_2^2
		\end{bmatrix},
	\end{align*}
	the matrix \(\boldsymbol{C}(\boldsymbol{q}, \dot{\boldsymbol{q}})\) defines the Coriolis and centrifugal terms as
	\begin{align*}
		\boldsymbol{C}(\boldsymbol{q}, \dot{\boldsymbol{q}}) = \begin{bmatrix}
			- m_2 l_1 l_2 (2 \omega_1 \omega_2 + \omega_2^2) \sin(q_2 - q_1) \\
			m_2 l_1 l_2 \omega_1^2 \sin(q_2 - q_1)
		\end{bmatrix},
	\end{align*}
	the vector \(\boldsymbol{g}(\boldsymbol{q}) \in \mathbb{R}^2\) is gravitational forces 
	\begin{align*}
		\boldsymbol{g}(q) = \begin{bmatrix}
			(m_1 + m_2) g l_1 \sin q_1 \\
			m_2 g l_2 \sin q_2
		\end{bmatrix}.
	\end{align*}

	\subsubsection{Dataset Creation}
	To train and evaluate UniConFlow on the toy problem introduced in \cref{subsec_double_pendulum}, we construct an expert trajectory dataset for the double inverted pendulum using MPC.
	We record the training dataset with discrete time indices, i.e., at each step, the state is \(\boldsymbol{s}^k = [q_1^k, q_2^k, \omega_1^k, \omega_2^k]^\top\) and the control input is \(\boldsymbol{a}^k = [\tau_1^k, \tau_2^k]^\top\).

	For each MPC solve, we consider a prediction horizon of length \(N \in \mathbb{N}\) and optimize over a sequence of predicted states
	\(\{\boldsymbol{s}^k\}_{k=0}^{N}\) and controls \(\{\boldsymbol{a}^k\}_{k=0}^{N-1} \).
	The stage and terminal costs are defined as
	\begin{subequations}\label{eq_J_pen}
		\begin{align}
			&J_{\mathrm{pen}}
			= \sum_{k=0}^{N-1} \ell_k
			+ \ell_N(\boldsymbol{s}^N,\boldsymbol{s}^{\mathrm{goal}}), \\
			&\ell_k
			= (\boldsymbol{s}^k - \boldsymbol{s}^{\mathrm{goal}})^\top
			\boldsymbol{Q} (\boldsymbol{s}^k - \boldsymbol{s}^{\mathrm{goal}})
			+ (\boldsymbol{a}^k)^\top \boldsymbol{R} \boldsymbol{a}^k, \\
			&\ell_N(\boldsymbol{s}^N,\boldsymbol{s}^{\mathrm{goal}})
			= (\boldsymbol{s}^N - \boldsymbol{s}^{\mathrm{goal}})^\top
			Q (\boldsymbol{s}^N - \boldsymbol{s}^{\mathrm{goal}}),
		\end{align}
	\end{subequations}
	where \(\boldsymbol{Q} \in \mathbb{R}^{4 \times 4}\) and \(\boldsymbol{R} \in \mathbb{R}^{2 \times 2}\) are positive-definite weighting matrices that penalize deviations from the target and control effort, respectively.
	At a given current state \(\boldsymbol{s}_{\mathrm{cur}}\), the MPC problem is formulated as
	\begin{subequations}\label{eq_ocp_pen}
		\begin{align}
			\min \quad
			& J_{\mathrm{pen}} \ \text{in}~\eqref{eq_J_pen}, \\
			\text{s.t.} \quad
			& \boldsymbol{s}^0 = \boldsymbol{s}_{\mathrm{cur}}, \\
			& \boldsymbol{s}^{k+1}
			= \boldsymbol{s}^k + {f}(\boldsymbol{s}^k,\boldsymbol{a}^k),
			\quad k = 0,\dots,N-1.
		\end{align}
	\end{subequations}
	Notably, only the first control input \(\boldsymbol{a}^{0}_{\star}\) of the optimal sequence is applied to the true system; the next state
	\(\boldsymbol{s}_{\mathrm{next}}\) is obtained by integrating the exact nonlinear dynamics \(f\) with a fourth-order Runge-Kutta over one sampling interval \(\Delta t = 0.1s\).
	
	The dataset is built by running this MPC controller from randomized initial conditions.
	For each rollout, the initial angles \(q_1^0,q_2^0\) are sampled independently and uniformly from \([0, 2\pi)\), and the initial angular velocities are set to zero, yielding
	\(\boldsymbol{s}_0 = [q_1^0,q_2^0,0,0]^\top\).
	The closed-loop system is then simulated for \(H\) time steps.
	Stacking all successful rollouts yields a collection of expert trajectories, where each includes a sequence of \(H-1\) control inputs and \(H\) discrete states.
	In total, we generate $5000$ expert rollouts with the MPC controller.

	\subsubsection{Implementation Details}
	\paragraph{Obstacle Modeling.}
	As described in \cref{subsec_double_pendulum}, safety is enforced by a
	planar wall constraint on the end-effector position of the second link.
	Let $(x_{\mathrm{ee}}(\boldsymbol{s}),y_{\mathrm{ee}}(\boldsymbol{s}))$
	denote the Cartesian coordinates of the tip.
	We place a virtual vertical wall at $x_{\mathrm{wall}} = -1.0\,\mathrm{m}$
	and define the safe set by
	\[
	x_{\mathrm{ee}}(\boldsymbol{s}^k) > x_{\mathrm{wall}}
	\quad\Leftrightarrow\quad
	h_{\mathrm{pen}}(\boldsymbol{s}^k)
	= -1 - x_{\mathrm{ee}}(\boldsymbol{s}^k) \le 0,
	\]
	for all time indices $k$ along the trajectory.
	
	\paragraph{Cost.}
	For the double inverted pendulum, we use a finite–horizon quadratic cost around
	the goal state $\boldsymbol{s}^{\mathrm{goal}} = [\pi,\pi,0,0]^\top$.
	Let $\boldsymbol{s}^k = [q_1^k, q_2^k, \omega_1^k, \omega_2^k]^\top$ and
	$\boldsymbol{a}^k = [\tau_1^k,\tau_2^k]^\top$ for $k = 0,\dots,N$.
	The stage and terminal costs are
	\begin{subequations}
		\label{eq:impl_pen_cost}
		\begin{align}
			&J_{\mathrm{pen}}
			= \sum_{k=0}^{N-1} \ell_k
			+ \ell_N(\boldsymbol{s}^N,\boldsymbol{s}^{\mathrm{goal}}), \\
			&\ell_k
			= (\boldsymbol{s}^k - \boldsymbol{s}^{\mathrm{goal}})^\top
			\boldsymbol{Q}\,(\boldsymbol{s}^k - \boldsymbol{s}^{\mathrm{goal}})
			+ (\boldsymbol{a}^k)^\top \boldsymbol{R}\,\boldsymbol{a}^k, \\
			&\ell_N(\boldsymbol{s}^N,\boldsymbol{s}^{\mathrm{goal}})
			= (\boldsymbol{s}^N - \boldsymbol{s}^{\mathrm{goal}})^\top
			\boldsymbol{Q}\,(\boldsymbol{s}^N - \boldsymbol{s}^{\mathrm{goal}}),
		\end{align}
	\end{subequations}
	with diagonal weight matrices $\boldsymbol{Q}
	= \mathrm{diag}(10.0,\,10.0,\,1.0,\,1.0),~
	\boldsymbol{R}
	= \mathrm{diag}(0.1,\,0.1)$.

	\subsubsection{Additional Results}
	\label{subsec_double_pendulum_add}
	To visualize the evolution of the state of the tip of the inverted pendulum over time, we plot the state sequence and action rollout sequence for all approaches. 
	
	\begin{figure*}[t]
		\centering
		\includegraphics[width=1\linewidth]{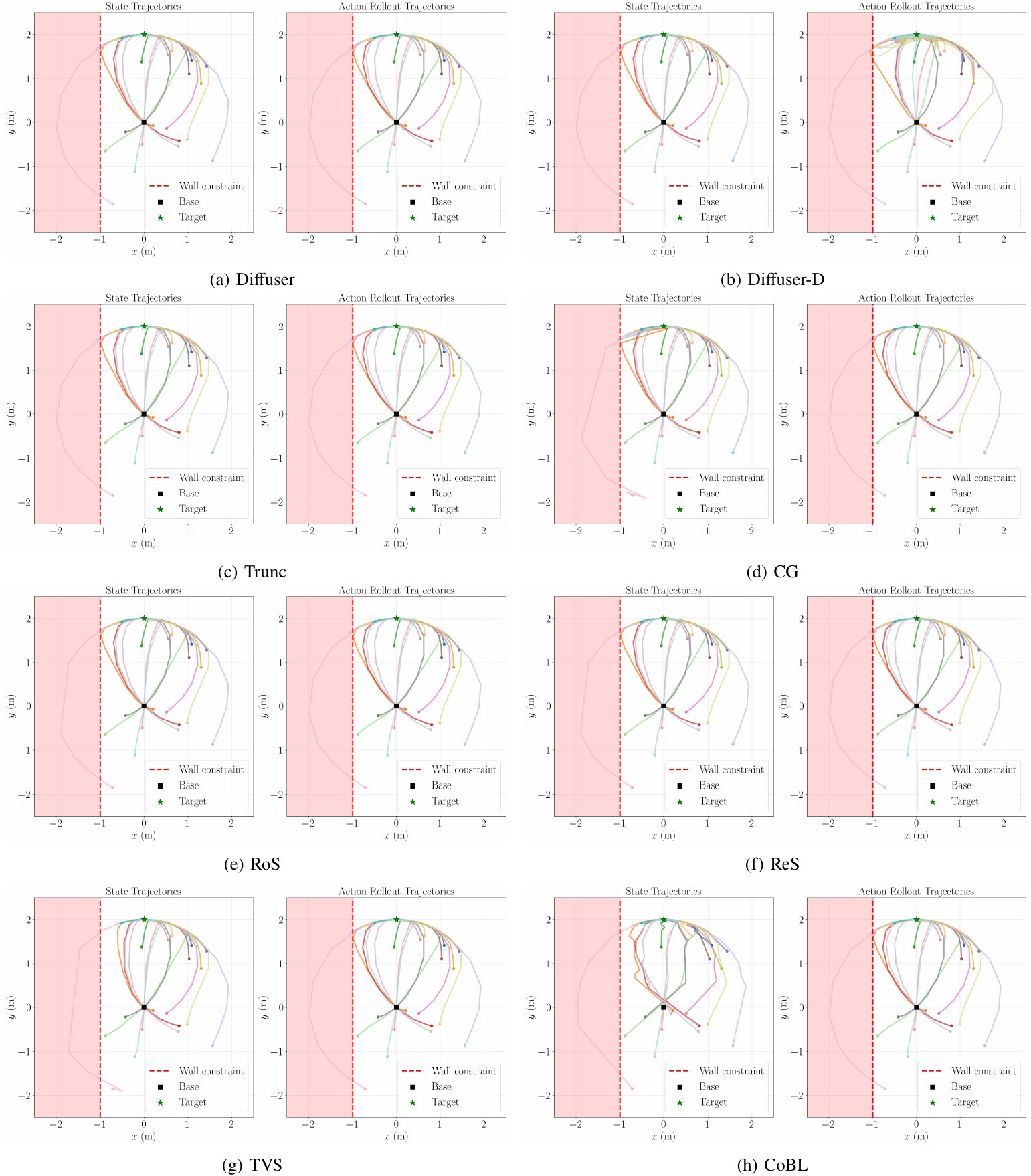}
	\end{figure*}
	\begin{figure*}[t]
		\centering
		\includegraphics[width=1\linewidth]{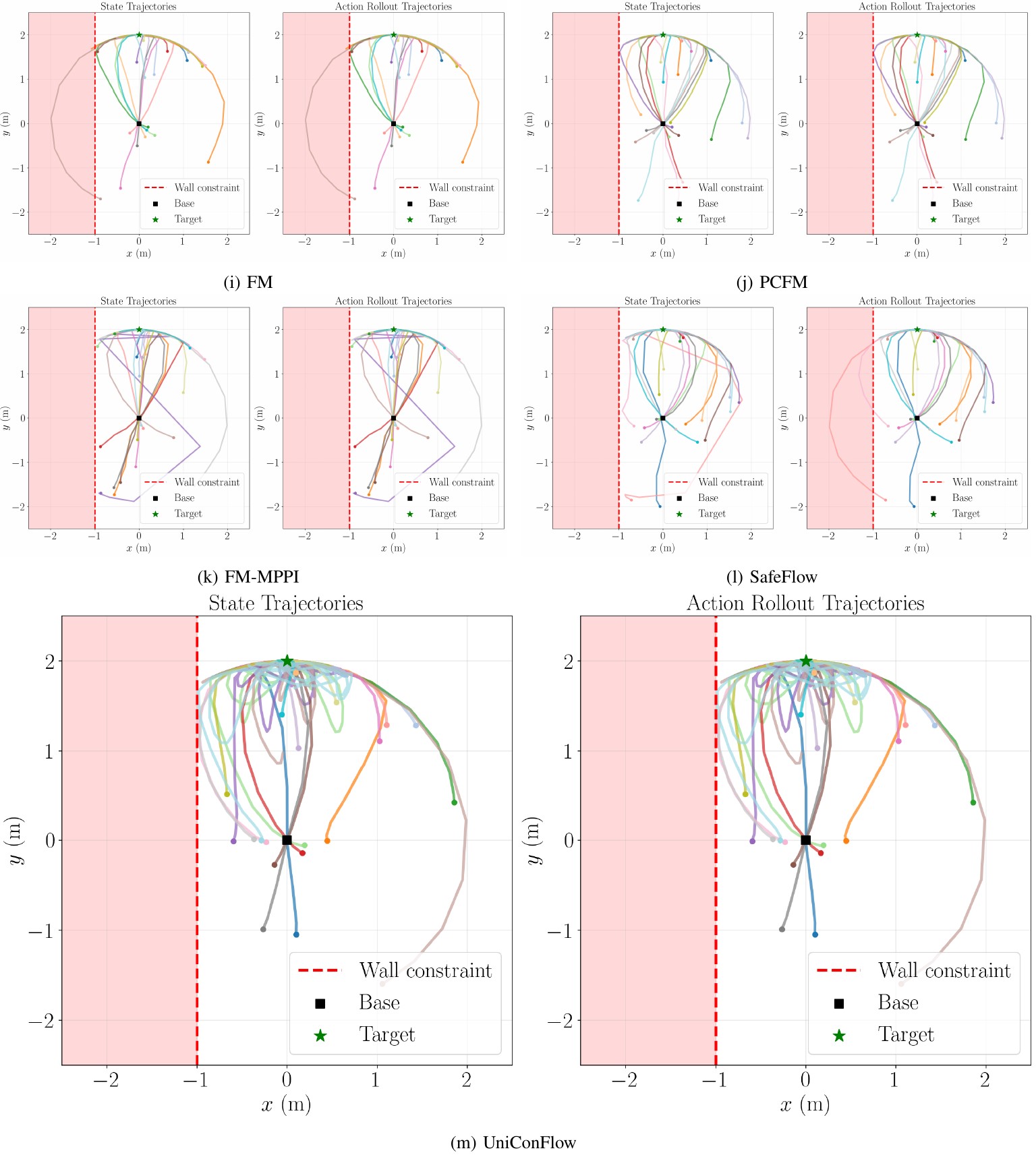}
		\caption{Visualization of the motion of the tip of the inverted pendulum over time.}
		\label{fig_pen_add}
	\end{figure*}

	The average values of all metrics over 1000 trials, along with the standard deviation, are provided in \cref{tab_pen}.
	\begin{table*}[t]
		\centering
		\caption{Results comparison on different metrics against baselines in the double inverted pendulum scenario.}
		\label{tab_pen}
		\begin{tabular}{lcccccccc}
			\toprule
			\textbf{Method} &
			{\textbf{SR-S}(\%)  ($\uparrow$)} &
			{\textbf{SR-A}(\%)  ($\uparrow$)} &
			{\textbf{AR}(\%)  ($\uparrow$)} &
			{\textbf{TSR}(\%)  ($\uparrow$)} &
			{\textbf{KC-F} ($\downarrow$)} &
			{\textbf{KC-I} ($\downarrow$)} &
			{\textbf{Cost}  ($\downarrow$)} &
			{\textbf{Time}(ms)  ($\downarrow$)} \\
			\midrule
			Diffuser     & 86.10 & 86.40 & 95.53 & 83.66 & 0.007$\pm$0.000 & 1.18$\pm$0.50 & 535.30 & \textbf{1.9}$\pm$0.001 \\
			Diffuser-D   & 86.10 & 74.60 & 76.27 & 69.06 & \textbf{0.000}$\pm$\textbf{0.000} & \textbf{0.0000}$\pm$\textbf{0.0000} & 601.52$\pm$202.28 & 2.9$\pm$0.001 \\
			Trunc        & 86.10 & 86.10 & 95.92 & 83.66 & 0.007$\pm$0.000 & 1.18$\pm$0.51 & 535.29 & 2.2$\pm$0.0001\\
			CG           & 87.35 & 86.70 & 91.85 & 83.92 & 0.087$\pm$0.004 & 1.48$\pm$0.69 & 536.40 & 133.3$\pm$0.001 \\
			RoS          & 87.62 & 86.30 & 94.28 & 84.19 & 0.033$\pm$0.008 & 1.68$\pm$0.51 & 533.67 & 139.9$\pm$0.001\\
			ReS          & 87.62 & 85.80 & 98.49 & 85.19 & 0.033$\pm$0.008 & 1.20$\pm$0.54 & 535.68 & 137.2$\pm$0.001\\
			TVS          & 97.49 & 86.10 & 98.57 & 93.05 & 0.067$\pm$0.002 & 1.45$\pm$0.58 & 532.35 & 140.7$\pm$0.001\\
			CoBL           & 88.66 & 85.29 & 86.29 & 85.51 & 0.093$\pm$0.002    & 1.68$\pm$0.57 & 547.82 & 2.5$\pm$0.001\\
			FM           & 86.10 & 88.94 & 99.17 & 85.12 & 0.004$\pm$0.000    & 1.18$\pm$0.52 & 531.76 & 1.7$\pm$0.001\\
			PCFM         & 94.12 & 86.30 & 92.61 & 92.04 & 0.003$\pm$0.000   & 1.14$\pm$0.34 & 522.46 & 8.4$\pm$0.001\\
			FM-MPPI      & \textbf{100} & \textbf{100} & 99.97 & 99.97 & \textbf{0.000$\pm$0.000} & \textbf{0.00$\pm$0.00} & 530.79 & 12.9$\pm$0.001\\
			SafeFlow & \textbf{100} & 86.9 & 96.07 & 82.79 & 0.007$\pm$0.003 & 1.13$\pm$0.69 & 535.18$\pm$131.15& 2.2$\pm$0.001\\
			\textbf{UniConFlow} & \textbf{100} & \textbf{100} & \textbf{100} & \textbf{100} & \textbf{0.000$\pm$ 0.000} & \textbf{0.00$\pm$0.00} & \textbf{507.81} & 2.6$\pm$0.001\\
			\bottomrule
		\end{tabular}
	\end{table*}

	\subsection{Car Racing Experiment}
	\subsubsection{Dataset Creation}
	\label{subsubsec_Dataset_Creation_for_Car_Racing_Experiment}
	To train and evaluate UniConFlow on realistic driving behaviors, we construct an expert trajectory dataset on the Nürburgring Nordschleife racetrack introduced above. 
	All trajectories are generated by solving optimal control problems (OCPs) for the model \eqref{eq_dynamics_car}. 
	
	We start from randomly sampled initial states and control sequences in both forward and backward driving directions along the track. 
	For each scenario, we formulate a finite–horizon optimal control problem over the state sequence \(\{\boldsymbol{s}^k\}_{k=0}^{H}\) and control sequence \(\{\boldsymbol{a}^k\}_{k=0}^{H-1}\), governed by the vehicle dynamics \eqref{eq_dynamics_car}. 
	Let $\boldsymbol{\xi}_{\mathrm{ref}}^{k} = [x_{\mathrm{ref}}^{k}, y_{\mathrm{ref}}^{k}]^\top$ denote the raceline reference position at step $k$.
	
	The objective function balances raceline tracking, speed regulation, control smoothness, and terminal accuracy, and is designed as follows
	\begin{subequations}\label{eq_J_car}
		\begin{align}
			&J_{\text{car}}
			= \sum_{k=0}^{H-1} \ell_k
			+ \ell_H(\boldsymbol{s}^H,\boldsymbol{s}^{\mathrm{goal}}), \\
			&\ell_k
			= w_{\mathrm{trk}}
			\big\|\boldsymbol{\xi}^k - \boldsymbol{\xi}_{\mathrm{ref}}^k\big\|_2^2
			+ w_{\mathrm{vel}}
			\big(v^k - v_{\mathrm{ref}}^k\big)^2
			+ w_{\mathrm{ctr}}
			\big\|\boldsymbol{a}^k\big\|_2^2 \nonumber \\
			&\qquad + w_{\mathrm{sm}}
			\big\|\boldsymbol{a}^k - \boldsymbol{a}^{k-1}\big\|_2^2, \\
			&\ell_H(\boldsymbol{s}^H,\boldsymbol{s}^{\mathrm{goal}})
			= w_H \big\|\boldsymbol{s}^H-\boldsymbol{s}^{\mathrm{goal}}\big\|_2^2,
		\end{align}
	\end{subequations}
	where $\ell_k$ denotes the stage cost at time step $k$, comprising position tracking, speed tracking, control effort, and control smoothness penalties, while $\ell_H$ is the terminal cost enforcing convergence to the desired goal state. $ w_{\mathrm{trk}}, w_{\mathrm{vel}}, w_{\mathrm{ctr}}, w_{\mathrm{sm}} w_{\mathrm{vel}}, w_H \in \mathbb{R}_{>0}$ are positive weighting coefficients of the stage and terminal cost terms.
	Therefore, the OCP of car racing is thus
	\begin{subequations}
		\begin{align}
			\min \quad
			& J_{\text{car}}~ \text{in}~ \eqref{eq_J_car}, \\
			\text{s.t.} \quad
			& \boldsymbol{s}^{k+1}
			= f(\boldsymbol{s}^k,\boldsymbol{a}^k),
			\qquad k = 0,\dots,T-1, \\
			& \boldsymbol{a}_{\min}
			\le \boldsymbol{a}^k
			\le \boldsymbol{a}_{\max},
			\qquad k = 0,\dots,T-1, \\
			& 0 \le v^k \le v_{\max},
			\qquad k = 0,\dots,T, \label{eq_vmax_car} \\
			& \theta_{\min} \le \theta^k \le \theta_{\max},
			\qquad k = 0,\dots,T, \label{eq_thetamax_car} \\
			& d_{\mathrm{trk}}^{L}(\boldsymbol{\xi}^k) >0 ,
			\qquad k = 0,\dots,T, \label{eq_track_safety_L} \\
			& d_{\mathrm{trk}}^{R}(\boldsymbol{\xi}^k) >0,
			\qquad k = 0,\dots,T. \label{eq_track_safety_R}
		\end{align}
	\end{subequations}
	where the bounds on velocity \eqref{eq_vmax_car}, steering angle
	\eqref{eq_thetamax_car}, and the distance constraints
	\eqref{eq_track_safety_L}-\eqref{eq_track_safety_R} enforce collision-free driving along the track.
	We solve this OCP for randomly sampled initial conditions in both the
	forward and reverse driving directions along the track.
	In total, we generate $5000$ expert trajectories for forward driving and
	another $5000$ trajectories for reverse driving.
	Each solved OCP yields a sequence of $100$ control inputs and
	$101$ discrete states.

	\subsubsection{Implementation Details}
	
	\label{subsubsec_CEM_cost_function_car}
	
	We now detail the CEM subproblems used on the three types of windows (frozen, violation, and recovery) introduced in \cref{subsec:unicon_pipeline}.  
	For any window $\mathcal{W} \subseteq \{0,\dots,H\}$ with entry state $\boldsymbol{s}^{k^-}$ (the first index in $\mathcal{W}$), a control sequence on this window is denoted by
	\[
	\bm{U}_{\mathcal{W}}
	:= \{\boldsymbol{a}^k\}_{k \in \mathcal{W}}
	\in \mathbb{C}_a^{|\mathcal{W}|},
	\]
	where $\mathbb{C}_a$ is the admissible set of controls determined by the input bounds and the action constraints in~\eqref{eq_inequality_car}.  
	Starting from the fixed entry state $\boldsymbol{s}^{k^-}$, the corresponding state sequence
	$\{\boldsymbol{s}^k(\bm{U}_{\mathcal{W}})\}_{k \in \mathcal{W}}$ is obtained by rolling out the nonholonomic car dynamics~\eqref{eq_dynamics_car}.
	
	Each CEM problem solves a finite-horizon optimal control problem of the form
	\begin{align}
		\bm{U}_{\mathcal{W}}^{\star}
		=
		\arg\min_{\bm{U}_{\mathcal{W}} \in \mathbb{C}_a^{|\mathcal{W}|}} 
		J(\bm{U}_{\mathcal{W}}),
		\label{eq:general_CEM_obj}
	\end{align}
	which CEM approximates via iterative sampling and refitting of a Gaussian distribution over $\bm{U}_{\mathcal{W}}$.

	Frozen-window refinement:
	For each frozen window $\mathcal{W}_{\mathrm{frz}}^{(i)}$, we denote by
	$\{\hat{\boldsymbol{s}}^k\}_{k \in \mathcal{W}_{\mathrm{frz}}^{(i)}}$
	the PTZF-guided states obtained after guided sampling.  
	The CEM objective aligns the dynamics rollout with these reference states while encouraging smooth controls:
	\begin{align}
		J_{\mathrm{pref}}^{(i)}(\bm{U}_{\mathrm{frz}}^{(i)})
		&=
		\sum_{k \in \mathcal{W}_{\mathrm{frz}}^{(i)}}
		\left\| 
		\boldsymbol{s}^k(\bm{U}_{\mathrm{frz}}^{(i)}) - \hat{\boldsymbol{s}}^k 
		\right\|_2^2
		\\
		&\quad
		+ \lambda_{\mathrm{sm}}
		\sum_{k \in \mathcal{W}_{\mathrm{frz}}^{(i)} \setminus \{\max \mathcal{W}_{\mathrm{frz}}^{(i)}\}}
		\left\| 
		\boldsymbol{a}^{k+1} - \boldsymbol{a}^{k} 
		\right\|_2^2,\nonumber
		\label{eq:J_pref}
	\end{align}
	where $\bm{U}_{\mathrm{frz}}^{(i)} := \{\boldsymbol{a}^k\}_{k \in \mathcal{W}_{\mathrm{frz}}^{(i)}}$ and $\lambda_{\mathrm{sm}} \in \mathbb{R}_{>0}$ is a smoothness weight.  
	The last index of the window is excluded from the smoothness sum so that the term $\boldsymbol{a}^{k+1}$ is well-defined for all summation indices.
	
	Safety-aware CEM on violation windows:
	Recall from \cref{subsec:unicon_pipeline} that the violation windows 
	$\{\mathcal{W}_{\mathrm{vio}}^{(i)}\}_{i=1}^{m}$ are obtained by dilating the contiguous sets of indices where the state constraint $h_{\mathrm{car}}(\boldsymbol{s}^k) \le 0$ is violated.  
	For window $i$, let $\mathcal{W}_{\mathrm{vio}}^{(i)} = \{k_i^{-},\dots,k_i^{+}\}$ and let $\boldsymbol{s}^{k_i^{-}}$ be the frozen entry state at the beginning of this window.
	
	We denote by $d_{\mathrm{obs}}(\boldsymbol{\xi}^k)$ the signed distance to the union of all obstacles and by
	$d_{\mathrm{trk}}^{L}(\boldsymbol{\xi}^k)$, $d_{\mathrm{trk}}^{R}(\boldsymbol{\xi}^k)$
	the signed distances to the left and right track boundaries, respectively, as in the construction of $h_{\mathrm{car}}(\boldsymbol{s}^k)$ in~\eqref{eq:h_car_state}.  
	To obtain smooth penalty functions compatible with the rasterized track map, we introduce
	\begin{align}
		\phi_{\mathrm{obs}}(\boldsymbol{s}^k) &:= d_{\mathrm{obs}}(\boldsymbol{\xi}^k), \\
		\phi_{\mathrm{trk}}(\boldsymbol{s}^k) &:= 
		-\min\big\{ d_{\mathrm{trk}}^{L}(\boldsymbol{\xi}^k),\, d_{\mathrm{trk}}^{R}(\boldsymbol{\xi}^k) \big\},
	\end{align}
	and define the associated soft penalties
	\begin{align}
		\psi_{\mathrm{obs}}(\boldsymbol{s}) &= \max\big(0,\; - \phi_{\mathrm{obs}}(\boldsymbol{s})\big)^2, \\
		\psi_{\mathrm{trk}}(\boldsymbol{s}) &= \max\big(0,\; \phi_{\mathrm{trk}}(\boldsymbol{s})\big)^2,
	\end{align}
	which penalize obstacle penetration and track-limit violations, respectively.  
	Note that these penalties are consistent with the state constraint $h_{\mathrm{car}}(\boldsymbol{s}^k) \le 0$ in~\eqref{eq:h_car_state}.
	
	For each violation window, the safety-aware CEM problem minimizes
	\begin{align}
		J_{\mathrm{vio}}^{(i)}(\bm{U}_{\mathrm{vio}}^{(i)})
		&= \lambda_{\mathrm{obs}}
		\sum_{k \in \mathcal{W}_{\mathrm{vio}}^{(i)}}
		\psi_{\mathrm{obs}}\big(\boldsymbol{s}^k(\bm{U}_{\mathrm{vio}}^{(i)})\big)
		\nonumber\\
		&\quad
		+ \lambda_{\mathrm{trk}}
		\sum_{k \in \mathcal{W}_{\mathrm{vio}}^{(i)}}
		\psi_{\mathrm{trk}}\big(\boldsymbol{s}^k(\bm{U}_{\mathrm{vio}}^{(i)})\big)
		\nonumber\\
		&\quad
		+ \lambda_{\mathrm{rmse}}
		\sum_{k \in \mathcal{W}_{\mathrm{vio}}^{(i)}}
		\left\|
		\boldsymbol{s}^k(\bm{U}_{\mathrm{vio}}^{(i)}) - \hat{\boldsymbol{s}}^k
		\right\|_2^2
		\nonumber\\
		&\quad
		+ \lambda_{\mathrm{term}}
		\left\|
		\boldsymbol{s}^{k_i^{+}}(\bm{U}_{\mathrm{vio}}^{(i)}) - \hat{\boldsymbol{s}}^{k_i^{+}}
		\right\|_2^2
		\nonumber\\
		&\quad
		+ \lambda_{\mathrm{sm}}
		\sum_{k \in \mathcal{W}_{\mathrm{vio}}^{(i)} \setminus \{k_i^{+}\}}
		\left\| \boldsymbol{a}^{k+1} - \boldsymbol{a}^{k} \right\|_2^2,
		\label{eq:J_vio}
	\end{align}
	where $\bm{U}_{\mathrm{vio}}^{(i)} := \{\boldsymbol{a}^k\}_{k \in \mathcal{W}_{\mathrm{vio}}^{(i)}}$, and
	$\lambda_{\mathrm{obs}}, \lambda_{\mathrm{trk}}, \lambda_{\mathrm{rmse}}, \lambda_{\mathrm{term}}, \lambda_{\mathrm{sm}} \in \mathbb{R}_{>0}$ are weighting coefficients.  
	All candidate control sequences sampled by CEM are rolled out through the car dynamics~\eqref{eq_dynamics_car}, so every candidate $\bm{U}_{\mathrm{vio}}^{(i)}$ is dynamically feasible by construction.
	
	Recovery windows:
	Between each violation window $\mathcal{W}_{\mathrm{vio}}^{(i)}$ and the subsequent frozen segment, we introduce a recovery window 
	$\mathcal{W}_{\mathrm{rec}}^{(i)} = \{\kappa_i^{-},\dots,\kappa_i^{+}\}$ that connects the last state of the optimized violation segment to the first state of the next frozen segment.  
	Let $\boldsymbol{s}^{\mathrm{vio},k_i^{+}}$ denote the terminal state of the optimized violation window and 
	$\boldsymbol{s}^{\mathrm{frz},\kappa_i^{+}}$ the fixed target state at the beginning of the next frozen segment.  
	Since the recovery windows are chosen away from obstacles, we omit obstacle penalties and focus on alignment and smoothness:
	\begin{align}
		J_{\mathrm{rec}}^{(i)}(\bm{U}_{\mathrm{rec}}^{(i)})
		&= \lambda_{\mathrm{start}}
		\left\|
		\boldsymbol{s}^{\kappa_i^{-}}(\bm{U}_{\mathrm{rec}}^{(i)}) - \boldsymbol{s}^{\mathrm{vio},k_i^{+}}
		\right\|_2^2
		\nonumber\\
		&\quad
		+ \lambda_{\mathrm{end}}
		\left\|
		\boldsymbol{s}^{\kappa_i^{+}}(\bm{U}_{\mathrm{rec}}^{(i)}) - \boldsymbol{s}^{\mathrm{frz},\kappa_i^{+}}
		\right\|_2^2
		\nonumber\\
		&\quad
		+ \lambda_{\mathrm{sm}}
		\sum_{k \in \mathcal{W}_{\mathrm{rec}}^{(i)} \setminus \{\kappa_i^{+}\}}
		\left\| \boldsymbol{a}^{k+1} - \boldsymbol{a}^{k} \right\|_2^2,
		\label{eq:J_tr}
	\end{align}
	where $\bm{U}_{\mathrm{rec}}^{(i)} := \{\boldsymbol{a}^k\}_{k \in \mathcal{W}_{\mathrm{rec}}^{(i)}}$ and
	$\lambda_{\mathrm{start}}, \lambda_{\mathrm{end}} \in \mathbb{R}_{>0}$ weight the entry and exit alignment terms.  
	After each recovery-window optimization, we roll out the dynamics on $\mathcal{W}_{\mathrm{rec}}^{(i)}$ and keep all previously optimized windows fixed.

	\paragraph{Obstacle Modeling.}
	All obstacles in the car-racing scenarios are modeled as 2D ellipses on
	the rasterized track map.
	Each ellipse is specified by a center
	$\boldsymbol{\xi}^{(o)}_{\mathrm{c}} = [x^{(o)}_{\mathrm{c}},
	y^{(o)}_{\mathrm{c}}]^\top$ in pixel coordinates, axis lengths
	$(a^{(o)},b^{(o)})$ (in pixels) along the major and minor axes, and an
	orientation angle $\theta^{(o)}$ (in degrees).
	The signed distance $d_{\mathrm{obs}}^{(o)}(\boldsymbol{\xi})$ used in
	\eqref{eq:h_car_state} is computed from this ellipse model, and the
	aggregated obstacle distance $d_{\mathrm{obs}}(\boldsymbol{\xi})$ is
	defined as the minimum over all components.
	
	In the forward-driving scenario, three static ellipses are placed along
	the track:
	\begin{itemize}
		\item Obstacle~1:
		$(x_{\mathrm{c}},y_{\mathrm{c}}) = (348, 290)$,
		$(a,b) = (40,10)$, $\theta = 30^\circ$;
		\item Obstacle~2:
		$(x_{\mathrm{c}},y_{\mathrm{c}}) = (331, 307)$,
		$(a,b) = (40,10)$, $\theta = 30^\circ$;
		\item Obstacle~3:
		$(x_{\mathrm{c}},y_{\mathrm{c}}) = (340, 298)$,
		$(a,b) = (30,10)$, $\theta = -45^\circ$.
	\end{itemize}
	
	In the reverse-driving scenario, a single bone-shaped obstacle is
	implemented as the union of three overlapping ellipses:
	\begin{itemize}
		\item Component~1:
		$(x_{\mathrm{c}},y_{\mathrm{c}}) = (46, 28)$,
		$(a,b) = (55,22)$, $\theta = 120^\circ$;
		\item Component~2:
		$(x_{\mathrm{c}},y_{\mathrm{c}}) = (257, 238)$,
		$(a,b) = (40,20)$, $\theta = 60^\circ$;
		\item Component~3:
		$(x_{\mathrm{c}},y_{\mathrm{c}}) = (504, 403)$,
		$(a,b) = (45,105)$, $\theta = -15^\circ$.
	\end{itemize}
	These ellipses define the $d_{\mathrm{obs}}^{(o)}(\cdot)$ terms in
	\eqref{eq:h_car_state} for both forward and reverse driving.
	
	\paragraph{Cost.}
	For the raceline experiment, we evaluate the generated control sequence
	$\boldsymbol{a}^k = [\delta^k,\tau^k]^\top$ along the horizon
	$k = 0,\dots,H$ using a lightweight quadratic cost on control magnitude
	and smoothness,
	\begin{equation*}
		J_{\mathrm{car}}
		= \sum_{k=0}^{H}
		\bigl(
		w_{\mathrm{ctrl}} J_{\mathrm{ctrl}}^k
		+ w_{\mathrm{smooth}} J_{\mathrm{smooth}}^k
		\bigr),
	\end{equation*}
	where
	\begin{subequations}
		\begin{align*}
			J_{\mathrm{ctrl}}^k
			&= (\delta^k)^2 + (\tau^k)^2, \\
			J_{\mathrm{smooth}}^k
			&= \begin{cases}
				0, & k = 0,\\[0.5ex]
				(\delta^k - \delta^{k-1})^2
				+ (\tau^k - \tau^{k-1})^2, & k \ge 1.
			\end{cases}
		\end{align*}
	\end{subequations}
	The weights are $ w_{\mathrm{ctrl}} = 10^{-3},~
	w_{\mathrm{smooth}} = 5 \times 10^{-2}$.

	\subsubsection{Additional Results}
	\label{subsubsection_arcar}
	To visualize the evolution of the car's state over time, we plot the state sequence and action rollout sequence for all approaches (except the results in \cref{fig_cartrajectories}). 
	\begin{figure*}[t]
		\centering
		\includegraphics[width=1\linewidth]{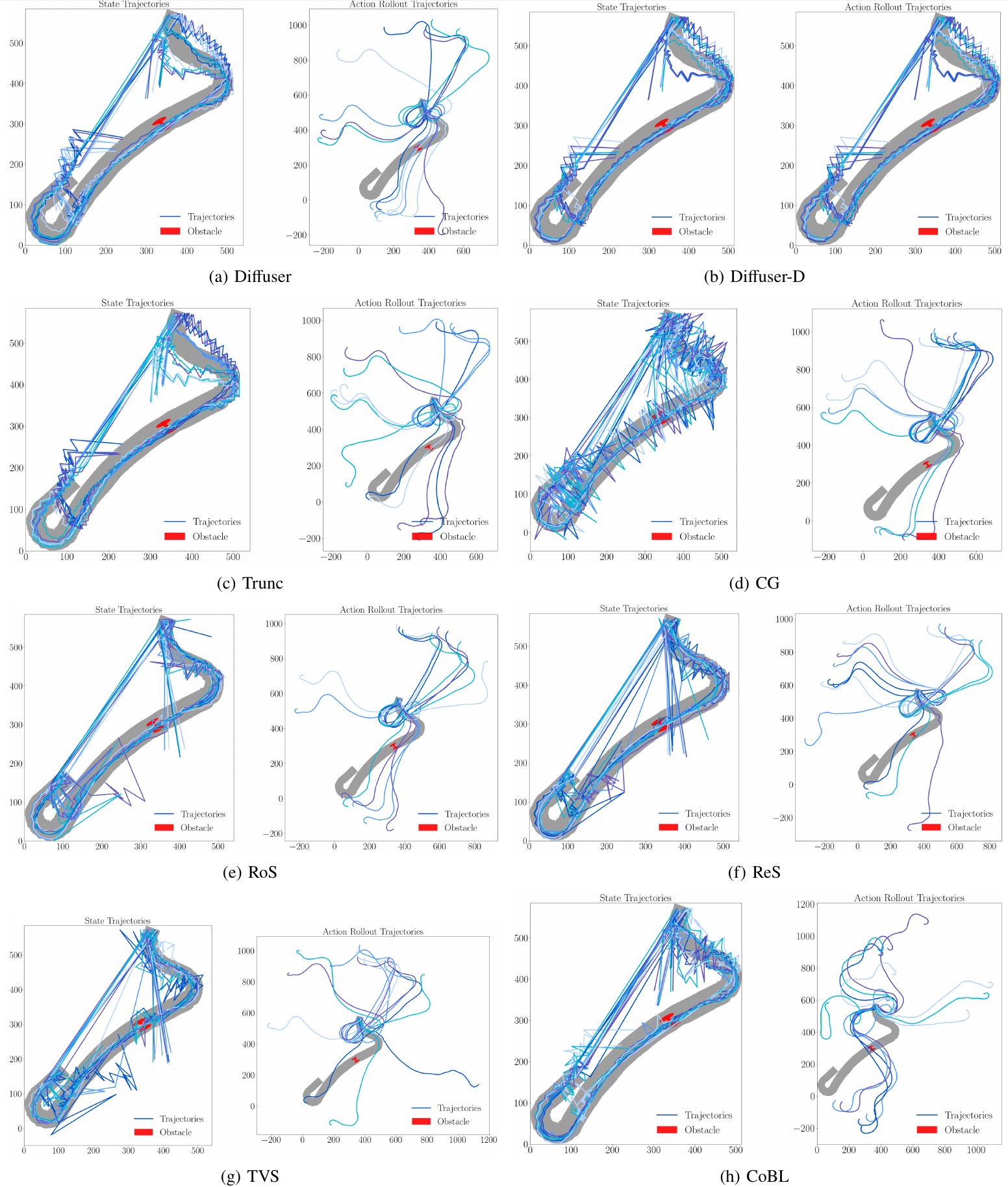}
	\end{figure*}
	\begin{figure*}[t]
		\centering
		\includegraphics[width=1\linewidth]{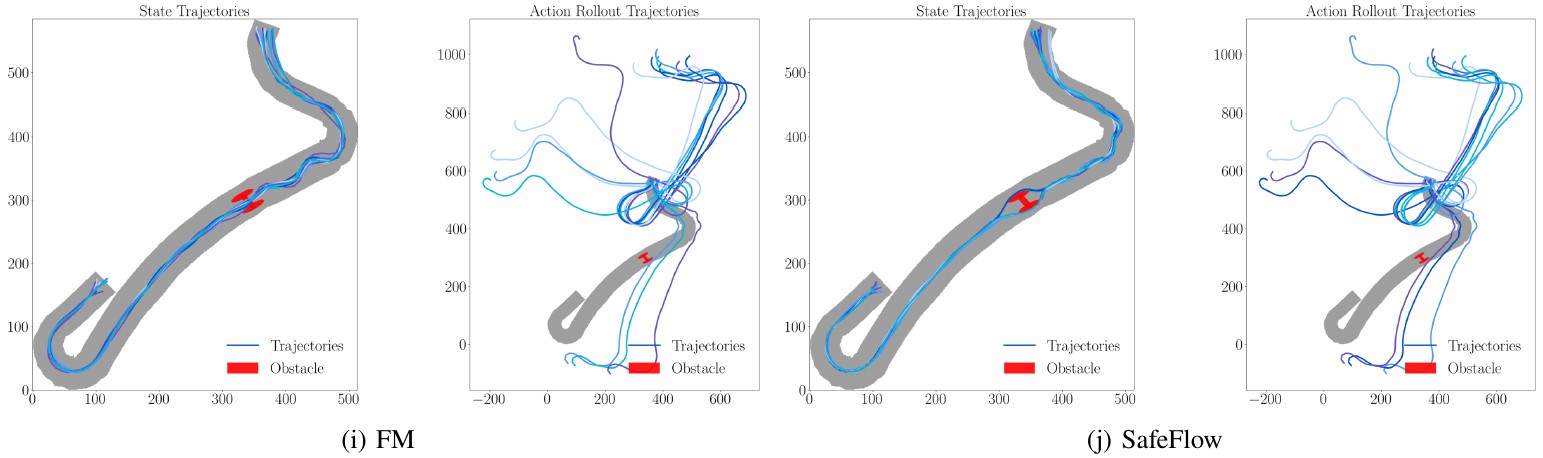}
		\caption{Visualization of the motion of the car's states over time in the forward driving scenario.}
		\label{fig_car_add_forward}
	\end{figure*}
	
	\begin{figure*}[t]
		\centering
		\includegraphics[width=1\linewidth]{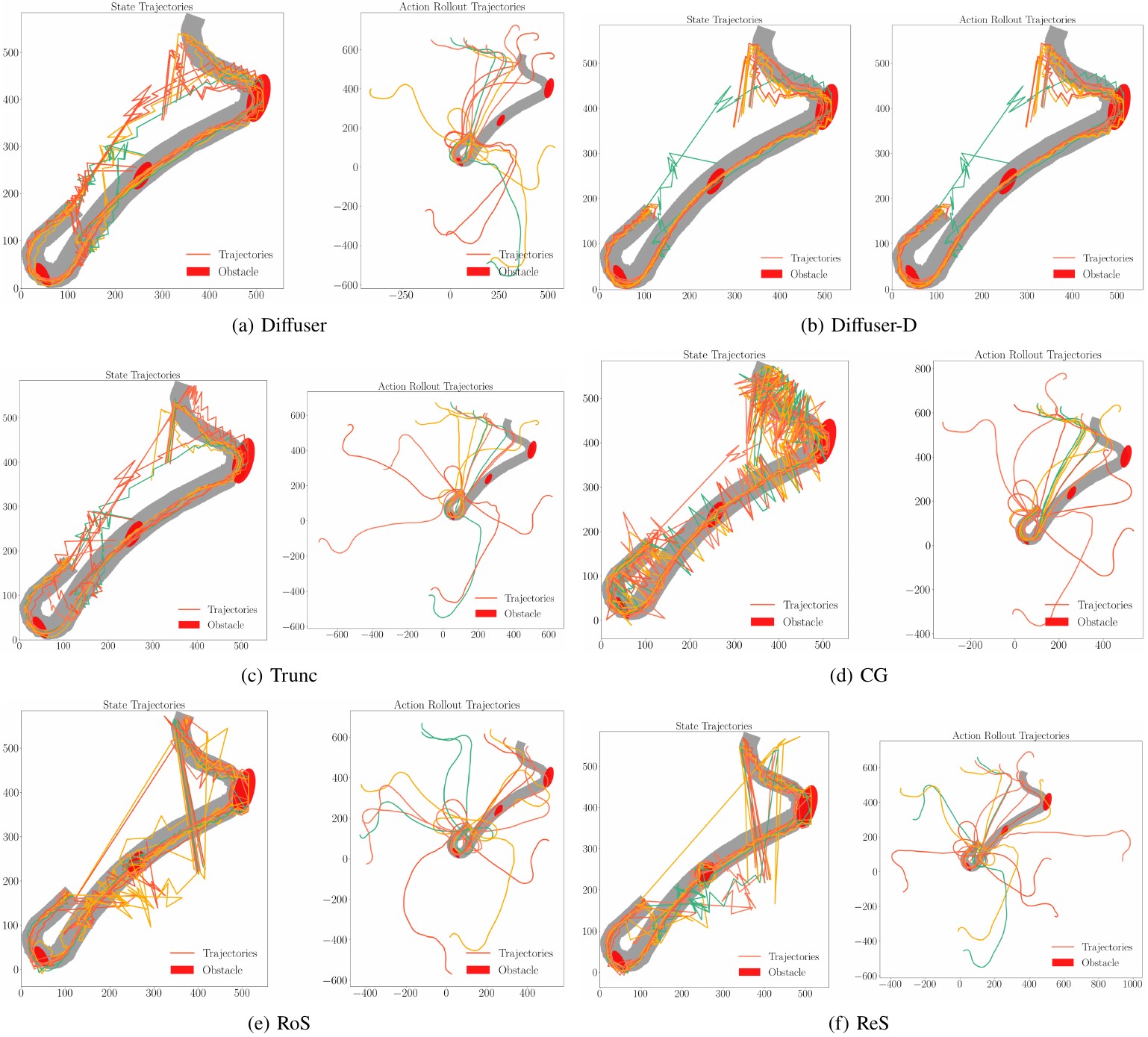}
	\end{figure*}
	\begin{figure*}[t]
		\centering
		\includegraphics[width=1\linewidth]{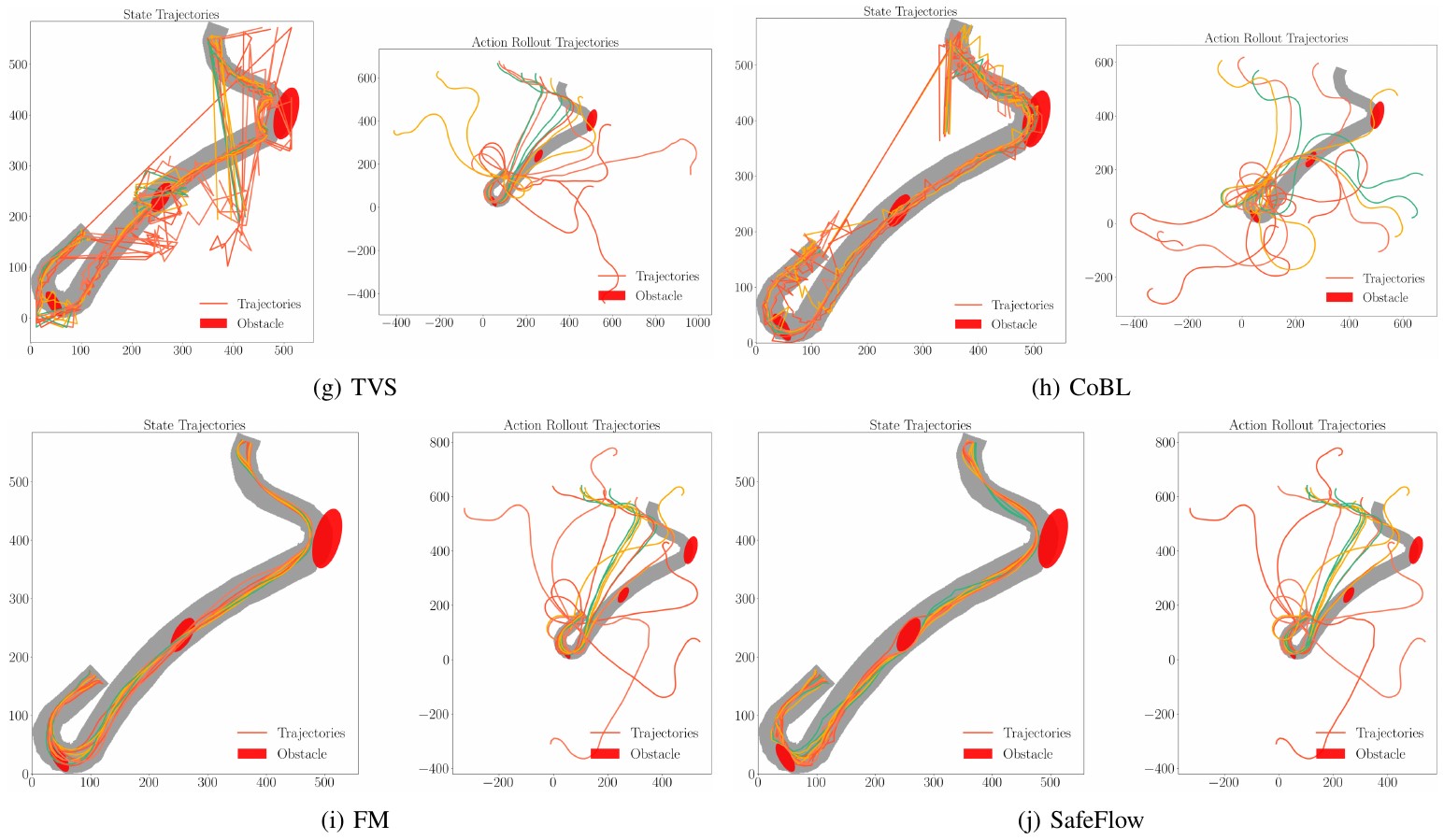}
		\caption{Visualization of the motion of the car's states over time in the reverse driving scenario.}
		\label{fig_car_add_forward}
	\end{figure*}
	
	In order to show the quantitative values for comparison, we provide the average values of all metrics over 1000 trials, along with the standard deviation for the forward driving and reverse driving cases in \cref{tab_forward_driving,tab_reverse_driving}, respectively.
	\begin{table*}[t]
		\centering
		\caption{Results comparison on different metrics against baselines for the car racing scenario (forward driving).}
		\label{tab_forward_driving}
		\begin{tabular}{lccccccccc}
			\toprule
			\textbf{Method} & \textbf{SR-S} (\%) & \textbf{SR-A} (\%) $\uparrow$ & \textbf{AR} (\%) $\uparrow$ & \textbf{TSR} (\%) $\uparrow$ & \textbf{KC-F} $\downarrow$ &\textbf{KC-I} $\downarrow$ & \textbf{Cost} $\downarrow$ & \textbf{Time} (ms) $\downarrow$ \\
			\midrule
			Diffuser     & 0.00 & 0.00 & \textbf{100.00} & 0.00 & 40.48$\pm$ 16.23 & 5.30 $\pm$ 1.09 &  99.62 $\pm$ 30.52&  2.10  $\pm$ 0.0001 \\
			Diffuser-D   & 0.00 & 0.00 &  93.72  & 0.00 & \textbf{0.00} $\pm$ 0.00& \textbf{0.00} $\pm$ 0.00 & 423.86 $\pm$ 51.21 &  2.30  $\pm$ 0.0001\\
			Trunc        & 0.00 & 0.00 & \textbf{100.00} & 0.00 & 41.16 $\pm$ 17.01 & 4.96 $\pm$ 2.79 & 103.27 $\pm$ 28.08&  2.10  $\pm$ 0.0001 \\
			CG           & 0.00 & 0.00 & \textbf{100.00} & 0.00 & 53.90 $\pm$ 9.14 & 4.45 $\pm$ 1.57& 123.67 $\pm$ 34.57& 286.40  $\pm$ 0.0001\\
			RoS          & 7.27 & 0.00 & \textbf{100.00} & 0.00 & 54.70 $\pm$ 14.22& 4.27 $\pm$ 1.59 & 130.71 $\pm$ 41.50 & 312.10  $\pm$ 0.0001\\
			ReS          & 14.55 & 0.00 & \textbf{100.00} & 0.00 & 52.63 $\pm$ 14.25 & 4.73 $\pm$ 1.74 & 113.59 $\pm$ 42.13&  23.90  $\pm$ 0.0001\\
			TVS          & 34.55 & 0.00 & \textbf{100.00} & 0.00 & 60.19 $\pm$ 12.60 & 4.17 $\pm$ 1.51 & 138.10 $\pm$ 43.59 & 314.10  $\pm$ 0.0001\\
			CoBL         & 1.29 & 0.00 & \textbf{100.00} & 0.00 & 43.44 $\pm$ 13.52 & 4.71 $\pm$ 1.61& 113.26 $\pm$ 30.85& 446.70  $\pm$ 0.0001\\
			FM           & 0.00 & 0.00 & \textbf{100.00} & 0.00 & 45.64 $\pm$ 13.03 & 5.72 $\pm$ 1.03 & 97.34 $\pm$ 22.18& \textbf{2.08}  $\pm$ 0.0001\\
			PCFM         & 0.00 & 0.02 &  0.00  & 0.00 & 37.17 $\pm$ 11.32  & 5.17 $\pm$ 1.51  & 108.81$\pm$ 25.47 & 180000.00 \\
			FM-MPPI      & 0.00 & 0.00 &  0.00  & 0.00 & 22.06 $\pm$ 9.90  & 3.19 $\pm$ 0.64  & 66731.02 $\pm$ 346.72 & 180000.00 \\
			SafeFlow     & \textbf{100.00} & 0.00 & \textbf{100.00} & 0.00 &  54.83 $\pm$ 16.47 & 7.29 $\pm$ 0.90 & 98.84 $\pm$ 19.87  & 2.60  $\pm$ 0.0001   \\
			UniConFlow   & \textbf{100.00} & \textbf{100.00} & \textbf{100.00} & \textbf{100.00} & \textbf{0.00$\pm$ 0.00} & \textbf{0.00$\pm$ 0.00} & \textbf{63.79$\pm$ 15.27 } & {2.70}  $\pm$ 0.0001 \\
			\bottomrule
		\end{tabular}
	\end{table*}

	\begin{table*}[t]
		\centering
		\caption{Results comparison on different metrics against baselines for the car racing scenario (reverse driving).}
		\label{tab_reverse_driving}
		\begin{tabular}{lccccccccc}
			\toprule
			\textbf{Method} & \textbf{SR-S}(\%) & \textbf{SR-A}(\%) $\uparrow$ & \textbf{AR}(\%) $\uparrow$ & \textbf{TSR}(\%) $\uparrow$ & \textbf{KC-F} $\downarrow$ &\textbf{KC-I} $\downarrow$ & \textbf{Cost} $\downarrow$ & \textbf{Time} (ms) $\downarrow$ \\
			\midrule
			Diffuser     & 0.00  & 0.00 & \textbf{100.00} & 0.00            & 26.29$\pm$ 12.323           & 4.53 $\pm$ 2.84       & 126.59 $\pm$ 27.34  & 2.90  $\pm$ 0.00\\
			Diffuser-D   & 0.00  &  0.00 & 91.22   & 0.00                   & \textbf{0.00}$\pm$ 0.00   & \textbf{0.00} $\pm$ 0.00 & 348.22  $\pm$ 79.85 & 3.00 $\pm$ 0.000\\
			Trunc        & 0.91  &  0.00 & \textbf{100.00} &  0.00          & 24.74$\pm$17.17          & 4.55 $\pm$ 2.76         & 129.77 $\pm$ 33.11  & 2.90 $\pm$ 0.00\\
			CG           & 0.00  &  0.00 & \textbf{100.00} & 0.00           & 23.95$\pm$6.29           & 4.05  $\pm$ 1.57      & 132.74 $\pm$ 28.57  & 178.40$\pm$ 0.00 \\
			RoS          & 0.00  &  0.00 & \textbf{100.00} & 0.00           & 24.50$\pm$17.73          & 4.03 $\pm$ 1.59         & 137.03 $\pm$ 27.88  & 177.30 $\pm$ 0.00\\
			ReS          & 0.91  &  0.00 & \textbf{100.00} & 0.00           & 26.58$\pm$11.34           & 3.97 $\pm$ 1.74         & 133.89 $\pm$ 28.39  & 18.90 $\pm$ 0.00\\
			TVS          & 66.36 &  0.00& \textbf{100.00} &  0.00           & 27.95$\pm$15.97           & 3.70 $\pm$ 1.51         & 138.27 $\pm$ 40.58  & 178.20 $\pm$ 0.00\\
			CoBL         & 0.00 & 0.00 & \textbf{100.00}  & 0.00            & 26.23$\pm$19.90           & 4.04 $\pm$ 1.61         &149.15  $\pm$ 27.56 &  449.10 $\pm$ 0.00\\
			FM           & 0.00  &  0.00 & \textbf{100.00} & 0.00           & 42.81$\pm$11.52           & 4.59 $\pm$ 2.01         & 121.18 $\pm$ 23.84  & 2.80 $\pm$ 0.00\\
			PCFM      & 0.00  &  0.05 & 0.00   & 0.00                       & 37.17$\pm$8.18           & 4.32 $\pm$ 1.00         & 109.11 $\pm$ 21.02  & 180000.00 $\pm$ 0.00\\
			FM-MPPI      & 0.00  &  0.00 & 0.00  & 0.00                     & 20.59$\pm$5.21         & 36.92 $\pm$ 15.90        &63242.24 $\pm$ 981.17 & 180000.00 $\pm$ 0.00\\
			SafeFlow   & \textbf{100.00} & 0.00 & \textbf{100.00} & 0.00    & 32.71$\pm$10.06       & 5.42 $\pm$ 1.98 & 107.73 $\pm$ 35.81         & 3.30   $\pm$ 0.00 \\
			UniConFlow   & \textbf{100.00} & \textbf{100.00} & \textbf{100.00} & \textbf{100.00} & \textbf{0.00$\pm$ 0.00} & \textbf{0.00 $\pm$ 0.00} & \textbf{64.21$\pm$ 10.73} & \textbf{2.50$\pm$ 0.00} \\
			\bottomrule
		\end{tabular}
	\end{table*}

	\subsection{Manipulator Experiment}
	
	\subsubsection{Dataset Creation}
	To obtain expert demonstrations on the manipulator tasks, we employ a
	receding-horizon MPC scheme on the full joint-space dynamics
	\eqref{eq_arm_dynamics}.
	At each control step with sampling time $\Delta t = 1/500\,\mathrm{s}$,
	MPC optimizes a finite-horizon sequence of states
	$\{\boldsymbol{s}^k\}_{k=0}^{H}$ and controls
	$\{\boldsymbol{a}^k\}_{k=0}^{H-1}$, where
	$\boldsymbol{s}^k = [\boldsymbol{q}^{k\top},\dot{\boldsymbol{q}}^{k\top}]^\top
	\in \mathbb{R}^{14}$ and
	$\boldsymbol{a}^k = \boldsymbol{\tau}^k \in \mathbb{R}^7$.
	The optimization problem has the generic form
	\begin{subequations}
		\label{eq:ocp_arm}
		\begin{align}
			\min_{\{\boldsymbol{s}^k,\boldsymbol{a}^k\}}
			\quad & J_{\mathrm{arm}} \\
			\text{s.t.}\quad
			& \boldsymbol{s}^0 = \boldsymbol{s}_{\mathrm{cur}}, \\
			& \boldsymbol{s}^{k+1}
			= \boldsymbol{f}(\boldsymbol{s}^k,\boldsymbol{a}^k),
			\quad k = 0,\dots,H-1, \\
			& h_{\mathrm{arm}}(\boldsymbol{s}^k) \le 0,
			\quad k = 0,\dots,H,
		\end{align}
	\end{subequations}
	where $J_{\mathrm{arm}}$ is the MPPI-style cost introduced in the
	Implementation Details, combining end-effector tracking, obstacle
	penalties, control effort, smoothness, and terminal accuracy, and
	$h_{\mathrm{arm}}(\cdot)$ is the aggregated obstacle-avoidance
	constraint from~\eqref{eq_h_arm_state} and all physical constraints on Franka arm.
	At each step, only the first control input of the optimal sequence is applied to the system, the horizon is shifted forward, and the OCP is resolved, yielding a standard receding-horizon MPC loop.
	
	For the pseudo-2D figure-eight scenario, the reference trajectory
	$\{\boldsymbol{p}_{\mathrm{ref}}^k\}$ is defined in the end-effector
	$yz$-plane, while the MPC runs on the full 7-DoF dynamics and enforces
	all joint and torque bounds.
	After a short transient, the closed-loop system converges to a stable
	limit cycle in joint space that realizes a smooth figure-eight motion of
	the stick tip.
	We record approximately five cycles of this motion, resulting in a
	single long rollout of about $2.6\times 10^4$ time steps.
	From this rollout we construct a dataset of $5000$ expert trajectories
	by random sliding-window sampling.
	Each trajectory has a horizon of roughly $8800$ time steps and covers
	one complete figure-eight cycle.

	For the 3D circular-motion scenario, we use the same MPC formulation
	\eqref{eq:ocp_arm} but track a three-dimensional reference
	$\{\boldsymbol{p}_{\mathrm{ref}}^k\} \subset \mathbb{R}^3$.
	The stick tip starts from a high point on the $z$-axis, gradually enters
	a circular trajectory in free space, traces one full revolution with
	simultaneous variation in all $(x,y,z)$ coordinates, and finally returns
	to the initial high point.
	Running MPC on this reference again produces a stable limit cycle in
	joint space.
	We record this closed-loop behavior and build another dataset of $5000$
	expert trajectories using the same sliding-window scheme, with each
	trajectory spanning about $8800$ time steps and covering one complete
	3D circle.

	\subsubsection{Implementation Details}
	\paragraph{Obstacle Modeling.}
	In both the pseudo-2D and 3D manipulator experiments, physical
	obstacles are modeled as 3D ellipsoids in the robot base frame.
	An ellipsoid is described by its center
	$\boldsymbol{c}^{(i)} \in \mathbb{R}^3$, semi-axes
	$\boldsymbol{r}^{(i)} = [r_x^{(i)},r_y^{(i)},r_z^{(i)}]^\top$, and an
	(optional) rotation around the $x$-axis.
	For an end-effector position $\boldsymbol{p}_{\mathrm{ee}}$, the CBF
	used in the cost $J_{\mathrm{arm}}$ and in
	\eqref{eq_h_arm_state} has the generic quadratic form
	\[
	h_i(\boldsymbol{p}_{\mathrm{ee}})
	= (\boldsymbol{p}_{\mathrm{ee}} - \boldsymbol{c}^{(i)})^\top
	Q^{(i)} (\boldsymbol{p}_{\mathrm{ee}} - \boldsymbol{c}^{(i)}) - 1,
	\]
	with a positive-definite shape matrix $Q^{(i)}$ determined by
	$\boldsymbol{r}^{(i)}$ and the chosen rotation.
	In the pseudo-2D case, only the $yz$-projection
	$\boldsymbol{p}_{yz}^k(\boldsymbol{q})$ of the end-effector and of the
	ellipsoids is used when computing the distances
	$d_{\mathrm{obs}}^i(\boldsymbol{p}_{yz}^k(\boldsymbol{q}))$.
	
	For the 3D circular-motion experiment, a single ellipsoidal obstacle is
	placed close to the center of the circle, with $ \boldsymbol{c}_{\mathrm{circ}}
	= [0.55,\; 0.05,\; 0.605]^\top~\mathrm{m},\qquad
	\boldsymbol{r}_{\mathrm{circ}}
	= [0.02,\; 0.015,\; 0.015]^\top~\mathrm{m}$,
	and its principal axes aligned with the world axes.
	
	For the pseudo-2D figure-eight experiment, three ellipsoids are used to
	form a composite obstacle region:
	\begin{itemize}
		\item \emph{Side obstacle}:
		$\boldsymbol{c}_{\mathrm{side}}
		= [0.68,\; 0.26,\; 0.70]^\top~\mathrm{m}$,
		$\boldsymbol{r}_{\mathrm{side}}
		= [0.03,\; 0.15,\; 0.01]^\top~\mathrm{m}$,
		rotated by $0.2$\,rad about the $x$-axis;
		\item \emph{Central obstacle}:
		$\boldsymbol{c}_{\mathrm{cent}}
		= [0.68,\; -0.12,\; 0.65]^\top~\mathrm{m}$,
		$\boldsymbol{r}_{\mathrm{cent}}
		= [0.03,\; 0.08,\; 0.10]^\top~\mathrm{m}$,
		axis-aligned;
		\item \emph{Carved hole}:
		$\boldsymbol{c}_{\mathrm{hole}}
		= [0.68,\; -0.06,\; 0.64]^\top~\mathrm{m}$,
		$\boldsymbol{r}_{\mathrm{hole}}
		= [0.03,\; 0.05,\; 0.05]^\top~\mathrm{m}$,
		axis-aligned.
	\end{itemize}
	The “hole’’ ellipsoid defines a carved opening inside the central
	obstacle: trajectories are required to remain outside the solid
	ellipsoids but are allowed to pass through the interior of the hole,
	which realizes a narrow safe corridor around the figure-eight reference.
	These ellipsoidal shapes are projected to the $yz$-plane and used to
	compute the distance terms $d_{\mathrm{obs}}^i(\cdot)$ in
	\eqref{eq_h_arm_state}.
	
	\paragraph{Cost.}
	For the manipulator benchmark, we use a cost over the
	discrete horizon $k = 0,\dots,H$.
	Let $\boldsymbol{s}^k = [\boldsymbol{q}^{k\top},\dot{\boldsymbol{q}}^{k\top}]^\top
	\in \mathbb{R}^{14}$ be the state and
	$\boldsymbol{a}^k = \boldsymbol{\tau}^k \in \mathbb{R}^7$ the control input.
	Denote the end–effector position and its reference by
	$\boldsymbol{p}_{\mathrm{ee}}^k$ and $\boldsymbol{p}_{\mathrm{ref}}^k$,
	and recall the aggregated obstacle constraint $h_{\mathrm{arm}}(\boldsymbol{s}^k)$
	in~\eqref{eq_h_arm_state}.
	The total cost is
	\begin{align}
		J_{\mathrm{arm}}
		&= w_{\mathrm{obs}} J_{\mathrm{obs}}
		+ w_{\mathrm{track}} J_{\mathrm{track}}
		+ w_{\mathrm{ctrl}} J_{\mathrm{ctrl}} \\
		&\quad+ w_{\mathrm{smooth}} J_{\mathrm{smooth}}
		+ w_{\mathrm{vel}} J_{\mathrm{vel}}
		+ w_{\mathrm{term}} J_{\mathrm{term}},
	\end{align}
	with
	\begin{subequations}
		\begin{align}
			J_{\mathrm{obs}}
			&= \sum_{k=0}^{H}
			\bigl[\max\{0,\,h_{\mathrm{arm}}(\boldsymbol{s}^k)\}\bigr]^2, \\
			J_{\mathrm{track}}
			&= \sum_{k=0}^{H}
			\bigl\|\boldsymbol{p}_{\mathrm{ee}}^k - \boldsymbol{p}_{\mathrm{ref}}^k\bigr\|_2^2, \\
			J_{\mathrm{ctrl}}
			&= \sum_{k=0}^{H}
			\bigl\|\boldsymbol{\tau}^k\bigr\|_2^2, \\
			J_{\mathrm{smooth}}
			&= \sum_{k=1}^{H}
			\bigl\|\boldsymbol{\tau}^k - \boldsymbol{\tau}^{k-1}\bigr\|_2^2, \\
			J_{\mathrm{vel}}
			&= \sum_{k=0}^{H}
			\bigl\|\dot{\boldsymbol{q}}^{k} - \dot{\boldsymbol{q}}_{\mathrm{ref}}^{k}\bigr\|_2^2, \\
			J_{\mathrm{term}}
			&= \bigl\|\boldsymbol{p}_{\mathrm{ee}}^{H} - \boldsymbol{p}_{\mathrm{ref}}^{H}\bigr\|_2^2.
		\end{align}
	\end{subequations}
	The weights are $w_{\mathrm{obs}} = 30.0,~
	w_{\mathrm{track}} = 3.0,~
	w_{\mathrm{ctrl}} = 0.8,~
	w_{\mathrm{smooth}} = 6.0,~
	w_{\mathrm{vel}} = 1.0,~
	w_{\mathrm{term}} = 10.0$.

	\subsubsection{Additional Results}
	\label{subsub_armain}
	To visualize the motion of the manipulator's end-effector state over time, we plot the state sequence and action rollout sequence for all approaches. 
	
	\begin{figure*}[t]
		\centering
		\includegraphics[width=1\linewidth]{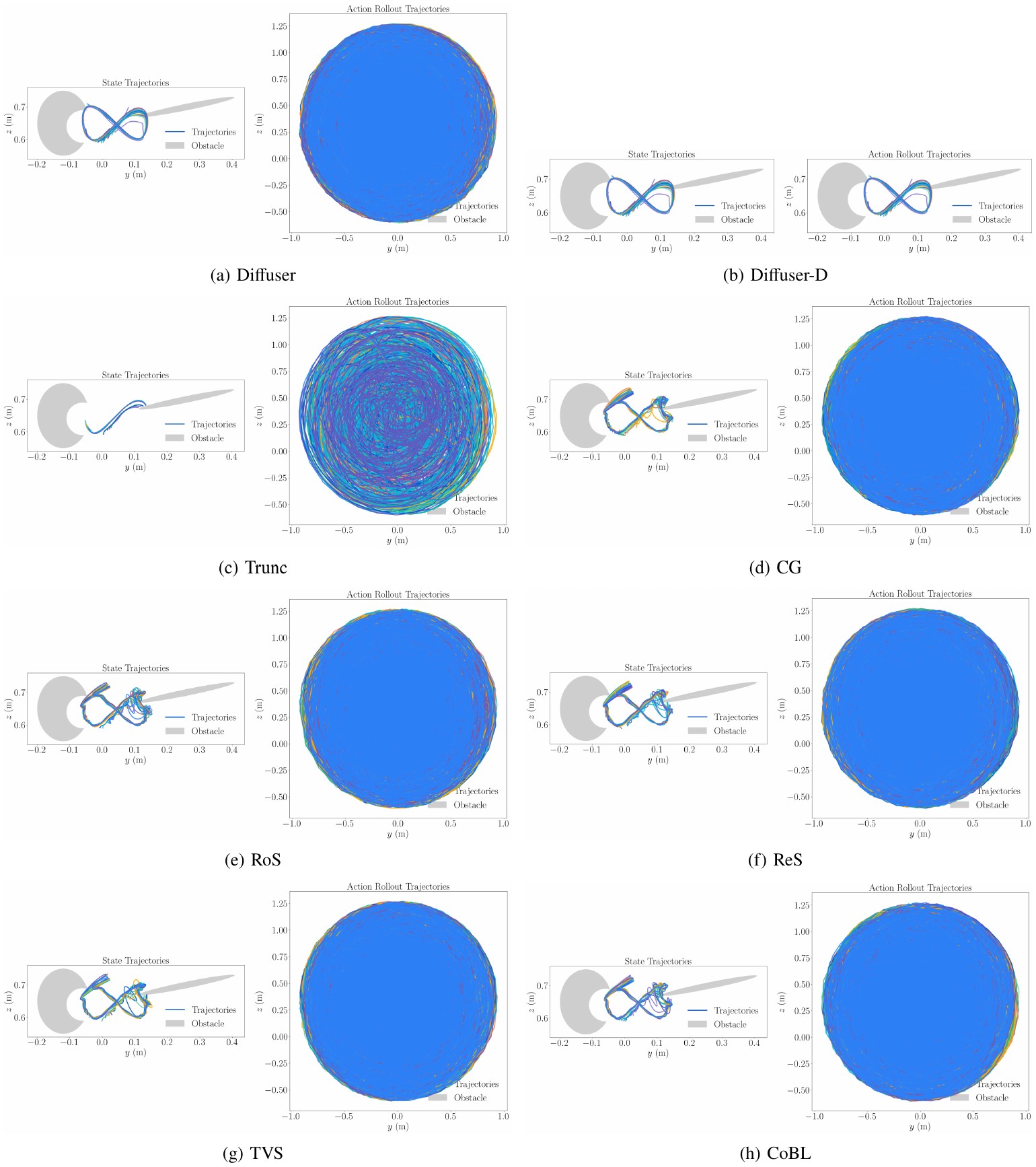}
	\end{figure*}
	\begin{figure*}[t]
		\centering
		\includegraphics[width=1\linewidth]{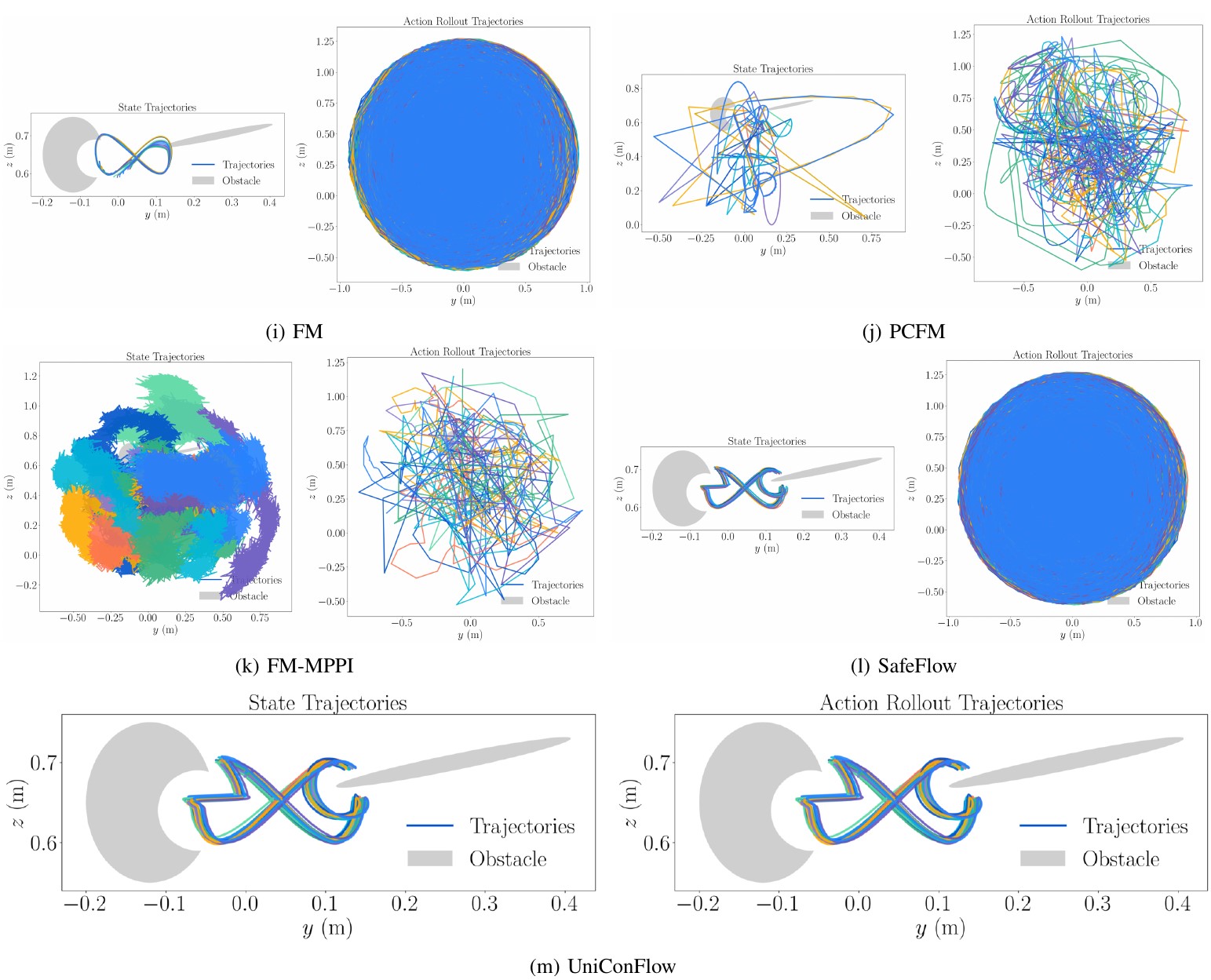}
		\caption{Visualization of the spatial motion of the manipulator's end-effector states over time in the pseudo-2D scenario.}
		\label{fig_car_add_forward}
	\end{figure*}
	
	\begin{figure*}[t]
		\centering
		\includegraphics[width=1\linewidth]{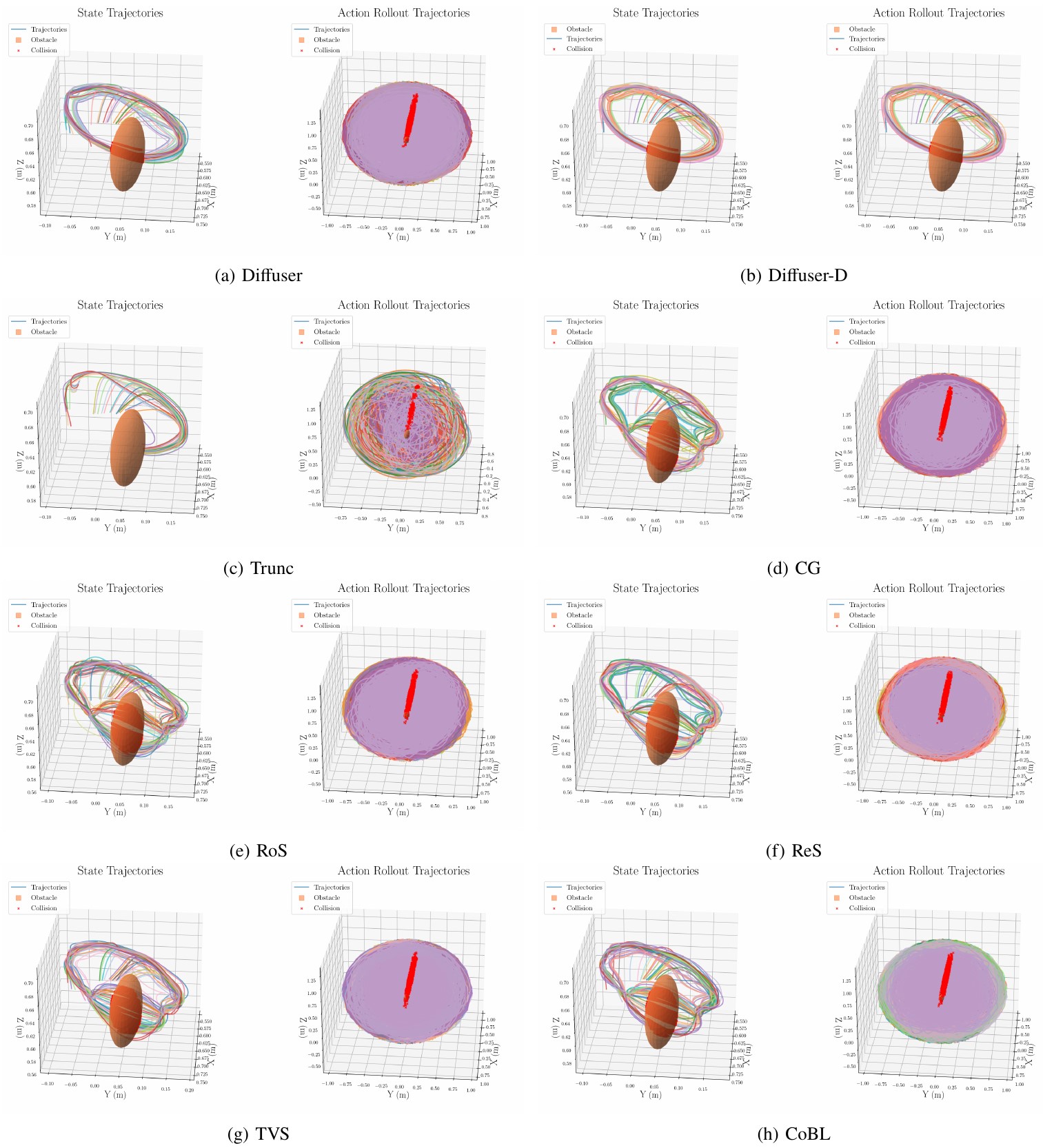}
	\end{figure*}
	\begin{figure*}[t]
		\centering
		\includegraphics[width=1\linewidth]{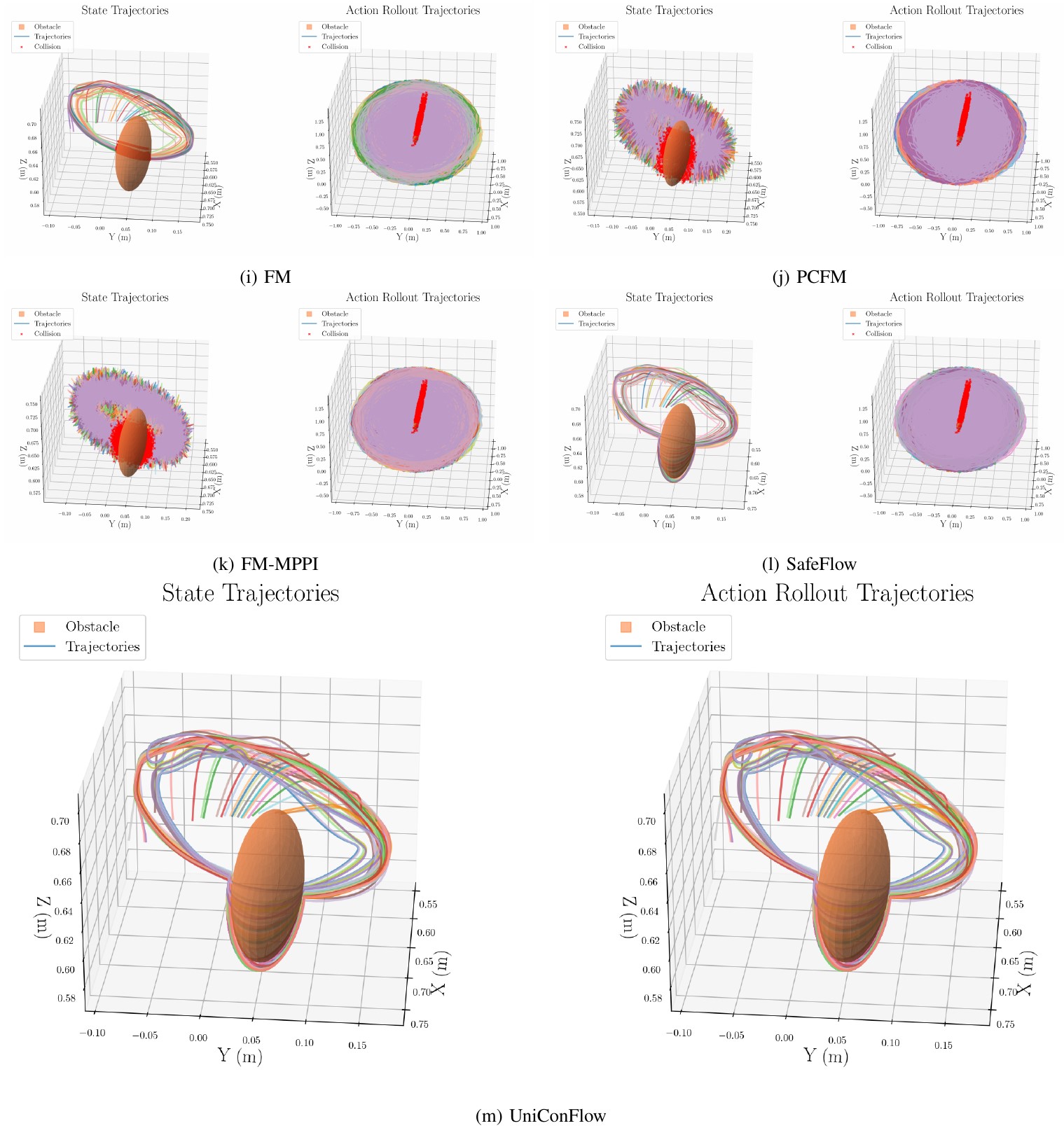}
		\caption{Visualization of the spatial motion of the manipulator's end-effector states over time in the 3D scenario.}
		\label{fig_car_add_forward}
	\end{figure*}
	
	To enable a quantitative comparison, we present the mean and standard deviation of all evaluation metrics over 1000 trials for the forward- and reverse-driving scenarios in \cref{tab_arm_add_2D,tab_arm_add_3D}, respectively.
	\begin{table*}[t]
		\centering
		\caption{Results comparison on different metrics against baselines for the manipulator pseudo-2D case.}
		\label{tab_arm_add_2D}
		\begin{tabular}{lccccccccc}
			\toprule
			\textbf{Method} & \textbf{SR-S} (\%) & \textbf{SR-A} (\%) $\uparrow$ & \textbf{AR} (\%) $\uparrow$ & \textbf{TSR} (\%) $\uparrow$ & \textbf{KC-F} $\downarrow$ &\textbf{KC-I} $\downarrow$ & \textbf{Cost} $\downarrow$ & \textbf{Time} (ms) $\downarrow$ \\
			\midrule
			Diffuser     & 86.29 & 0.00 & 100.00 & 0.00 & 2057.94 $\pm$ 59.74 & 37.09 $\pm$ 2.91  & $\approx 1.66 \times 10^{11}$ & 27.00 $\pm$ 0.02\\
			Diffuser-D   & 86.36 & 0.00 & 100.00 & 0.00 & 2057.89 $\pm$ 51.97 & 37.08 $\pm$ 7.24 & $\approx 1.66 \times 10^{11}$ & 28.00 $\pm$ 0.08\\
			Trunc        & 100.00 & 0.00 & 100.00 & 0.00 & 1228.73 $\pm$ 36.21 & 35.36 $\pm$ 8.17 & $\approx 1.25 \times 10^{10}$ & 59.91 $\pm$ 1.76\\
			CG           & 95.84 & 0.00 & 100.00 & 0.00 & 2059.33 $\pm$ 32.49 & 37.08 $\pm$ 26.42 & $\approx 3.62 \times 10^{10}$ & 4439.57 $\pm$ 0.02\\
			RoS          & 94.51 & 0.00 & 100.00 & 0.00 & 2059.94 $\pm$ 36.84& 37.08 $\pm$ 2.55 & $\approx 3.62 \times 10^{10}$ & 4435.32 $\pm$ 0.04\\
			ReS          & 97.78 & 0.00 & 100.00 & 0.00 & 2059.05 $\pm$ 29.66 & 37.08 $\pm$ 4.75 & $\approx 3.62 \times 10^{10}$ & 4462.81 $\pm$ 0.03\\
			TVS          & 93.39 & 0.00 & 100.00 & 0.00 & 2064.39 $\pm$ 32.14  & 37.07 $\pm$ 5.17 & $\approx 3.62 \times 10^{10}$ & 6491.22 $\pm$ 0.01\\
			CoBL         & 92.79 & 0.00 & 100.00 & 0.00 & 2059.60 $\pm$ 30.76 & 37.08 $\pm$ 6.12 & $\approx 3.66 \times 10^{10}$ & 586.35 $\pm$ 0.01\\
			FM           & 86.07  & 0.00 & 100.00 & 0.00  & 2056.43 $\pm$ 38.751 & 37.21 $\pm$ 2.88 & $\approx 1.66 \times 10^{11}$ & 26.00 $\pm$ 0.02\\
			PCFM       & 91.02  & 0.02 & 0.00  & 0.00  & 37.17 $\pm$ 32.79 & 5.17 $\pm$ 1.82 & $\approx 5.65 \times 10^{9}$    & $>$ 180000.00 \\
			FM-MPPI      & 82.17  & 0.00 & 0.00  & 0.00  & \textbf{0.00} $\pm$ 0.00  & \textbf{0.00} $\pm$ 0.00 & $\approx3.72 \times 10^{9}$ & $>$ 180000.00 \\
			SafeFlow & \textbf{100.00} & {0.00} & \textbf{100.00}& {0.00} & 2069.25 $\pm$ 21.85 & 33.50$\pm$ 1.08  &$\approx 1.62 \times 10^{10}$ & 35.00 $\pm$0.02\\
			UniConFlow   & \textbf{100.00} & \textbf{100.00} & \textbf{100.00} & \textbf{100.00} & \textbf{0.0000} $\pm$ 0.00 & \textbf{0.0000}$\pm$ 0.00 & \textbf{64.79 $\pm$15.72} & \textbf{32.00} $\pm$ 2.16\\
			\bottomrule
		\end{tabular}
	\end{table*}

	\begin{table*}[t]
		\centering
		\caption{Results comparison on different metrics against baselines for the manipulator 3D case.}
		\label{tab_arm_add_3D}
		\begin{tabular}{lccccccccc}
			\toprule
			\textbf{Method} & \textbf{SR-S} (\%) & \textbf{SR-A} (\%) $\uparrow$ & \textbf{AR} (\%) $\uparrow$ & \textbf{TSR} (\%) $\uparrow$ & \textbf{KC-F} $\downarrow$ &\textbf{KC-I} $\downarrow$ & \textbf{Cost} $\downarrow$ & \textbf{Time} (ms) $\downarrow$ \\
			\midrule
			Diffuser     & 0.00  & 0.00  & 100.00 & 0.00  & 692.44$\pm$2.43 & 0.06$\pm$0.05 &  $\!\approx\! 1.17 \!\times\! 10^{7}\pm$4.40 $\!\times\! 10^{5}$  &103.3$\pm$0.7  \\
			Diffuser-D   & 0.00  &  0.00 & 100.00 & 0.00  & 0.00$\pm$0.00     & 0.00$\pm$0.00 &  $\!\approx\! 1.18 \!\times\! 10^{6}\pm$4.67 $\!\times\! 10^{5}$    &103.4$\pm$0.9  \\
			Trunc        & 100.0 &  7.02 & 100.00 & 7.02  & 55.93$\pm$28.76 & 0.08$\pm$0.10 &  $\!\approx\! 2.37 \!\times\! 10^{7}\pm$7.17 $\!\times\! 10^{5}$  & 24.5$\pm$7.4  \\
			CG           & 12.30  & 0.00 & 100.00 & 0.00  & 692.37$\pm$2.64  & 0.06$\pm$0.05 &  $\!\approx\! 1.19 \!\times\! 10^{7}\pm$5.12 $\!\times\! 10^{5}$   &102.9$\pm$0.5\\
			RoS          & 7.02  &  0.00 & 100.00 & 0.00  & 677.34$\pm$2.43  & 0.06$\pm$0.03 &  $\!\approx\! 1.17 \!\times\! 10^{7}\pm$4.15 $\!\times\! 10^{5}$   &107.1$\pm$0.2\\
			ReS          & 3.95  &  0.00 & 100.00 & 0.00  & 689.28$\pm$2.54 & 0.05$\pm$0.04  &  $\!\approx\! 1.16 \!\times\! 10^{7}\pm$5.03 $\!\times\! 10^{5}$    &107.6$\pm$0.1 \\
			TVS          & 8.81  &  0.00 & 100.00 & 0.00  & 693.66$\pm$2.35  & 0.05$\pm$0.04 & $\!\approx\! 1.18 \!\times\! 10^{7}\pm$4.76 $\!\times\! 10^{5}$     &106.9$\pm$0.1 \\
			CoBL         & 6.92  &  0.00 & 100.00 & 0.00  & 648.72$\pm$5.72  & 0.04$\pm$0.02 & $\!\approx\! 1.19 \!\times\! 10^{7}\pm$4.75 $\!\times\! 10^{5}$     &109.8$\pm$0.1 \\
			FM           & 0.00  &  0.00 & 100.00 & 0.00  & 692.81$\pm$2.318  & 0.05$\pm$0.01 &  $\!\approx\! 1.16 \!\times\! 10^{7}\pm$4.24 $\!\times\! 10^{5}$   & 102.7$\pm$0.7 \\
			PCFM       & 0.00  &  0.05 & 0.00   & 0.00  &  463.13$\pm$1.225   & 0.039$\pm$0.02 &  $\!\approx\! 7.12 \!\times\! 10^{5}\pm$1.22 $\!\times\! 10^{3}$    & $>$180000.00       \\
			FM-MPPI      & 0.00  &  0.00 & 0.00  & 0.00  & 513.97$\pm$2.014  & 0.021$\pm$0.00    &$\!\approx\! 3.51 \!\times\! 10^{7}\pm$2.28 $\!\times\! 10^{5}$   & $>$180000.00       \\
			SafeFlow     & \textbf{100.00} & 0.00 & \textbf{100.00}  & 0.00 & 704.90$\pm$1.18  & 0.06$\pm$0.02  &  $\!\approx\! 1.59 \!\times\! 10^{7}\pm$2.12 $\!\times\! 10^{5}$ &108.1$\pm$0.3   \\
			UniConFlow   & \textbf{100.00} & \textbf{100.00} & \textbf{100.00} & \textbf{100.00} & \textbf{0.00}& \textbf{0.00}  &  $ 783.95 \pm$ 12.98  & \textbf{105.8}$\pm$0.2 \\
			\bottomrule
		\end{tabular}
	\end{table*}
	
	\subsection{Additional Ablation Study}
	\label{subsec_aas}
\begin{figure*}[t]
		\centering
		\includegraphics[width=1\linewidth]{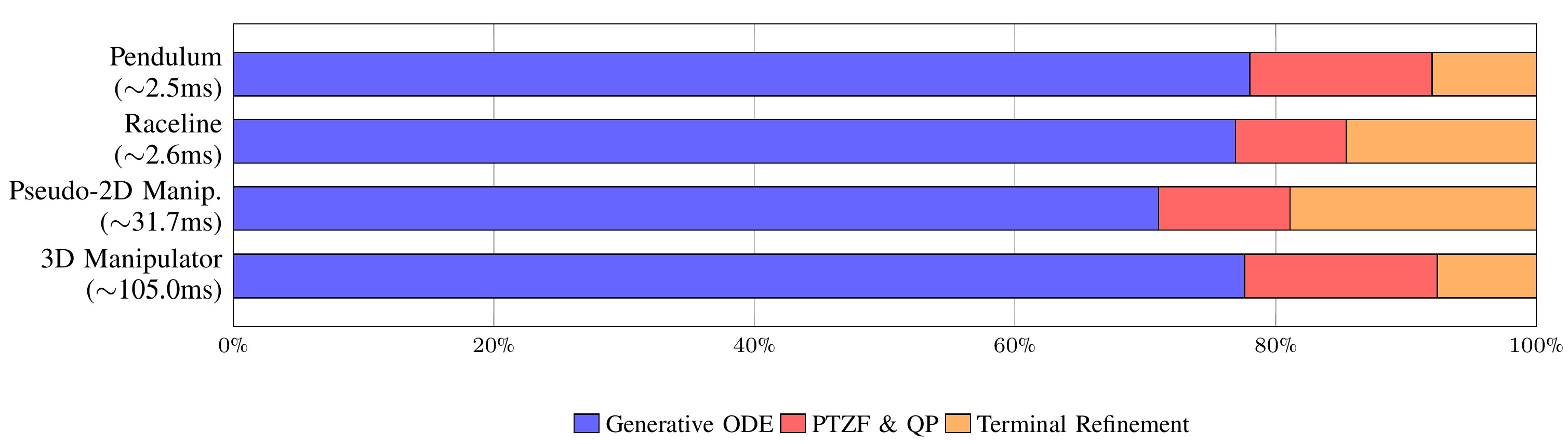}
		\caption{Percentage breakdown of the computational cost for the UniConFlow inference pipeline.}
		\label{fig:computation_bar_chart}
	\end{figure*}

	The percentage breakdown of inference cost across tasks is visualised in \cref{fig:computation_bar_chart}. 
	The generative model ODE integration consistently dominates the total budget across all dimensionalities, accounting for $78\%$, $76.9\%$, $71.0\%$, and $77.6\%$ of total inference time for the pendulum, raceline, pseudo-2D manipulator, and 3D manipulator, respectively. 
	The PTZF constraint evaluation and QP solving remain a modest fraction in low- and medium-dimensional cases ($8.5\%$ and $14.0\%$), growing to $14.8\%$ in the high-dimensional 3D manipulator. 
	Terminal refinement contributes $8.0\%$ to $14.6\%$ in the low- and medium-dimensional cases but decreases to $7.6\%$ for the 3D manipulator shown in \cref{tab:ablation_terminal}. 
	
	At each generative time step $t_\ell$, we measure the positive slack magnitude:
	\begin{equation}
		D(t_\ell):=\|\delta_{t_\ell}^{+}\|_1 = \sum_i \max(\delta_i(t_\ell),0),
		\label{eq:raw_slack_def}
	\end{equation}
	which directly measures how much positive slack is being used at that moment.
	Moreover, we report three statistics:
	\begin{align}
		\bar D_{\mathrm{raw}}
		&:=
		\frac{\sum_{\ell=1}^{L} D(t_\ell)\Delta t_\ell}
		{\sum_{\ell=1}^{L}\Delta t_\ell},
		\label{eq:raw_slack_mean}
		\\
		D_{\mathrm{raw}}^{\max}
		&:=
		\max_{1\le \ell\le L} D(t_\ell),
		\label{eq:raw_slack_peak}
		\\
		t_{\max}
		&:=
		\arg\max_{1\le \ell\le L} D(t_\ell),
		\label{eq:raw_slack_peak_time}
	\end{align}
	Here, $\bar D_{\mathrm{raw}}$ is the time-averaged raw slack magnitude, 
	$D_{\mathrm{raw}}^{\max}$ is the peak raw slack magnitude, 
	$t_{\max}$ is the generative time at which that peak occurs.
	
	\begin{table}[t]
		\centering
		\caption{
			Slack magnitude analysis across tasks.
		}
		\label{tab:direct_raw_slack}
		\footnotesize
		\setlength{\tabcolsep}{6pt}
		\renewcommand{\arraystretch}{1.12}
		\begin{tabular}{l l c c  c }
			\toprule
			\textbf{Experiment}  & $\bar D_{\mathrm{raw}}$ & $D_{\mathrm{raw}}^{\max}$ & $t_{\max}$ \\
			\midrule
			Raceline
			& $1.0049$
			& $1.7129$ 
			& $0.9411 $ 
			\\
			
			Pendulum
			& $1.7209\times 10^{-2}$
			& $3.7317 \times 10^{-2}$ 
			& $0.8813$ 
			\\
			
			2D manipulator
			& $1.1258\times 10^{-1}$
			& $5.7431\times 10^{-1}$ 
			& $0.7913$ 
			\\
			
			3D manipulator
			& $1.6400\times 10^{-1}$
			& $2.7364\times 10^{-1}$ 
			& $0.5597$ 
			\\
			\bottomrule
		\end{tabular}
	\end{table}
	As shown in Table~\ref{tab:direct_raw_slack}, the slack magnitudes remain small and 
	controlled across all tasks. For the Pendulum task, both $\bar{D}_{\mathrm{raw}}$ 
	and $D_{\mathrm{raw}}^{\max}$ are on the order of $10^{-2}$, which is virtually 
	negligible. For the more complex 2D and 3D Manipulator tasks, the time-averaged 
	slack stays well bounded ($0.11$-$0.16$), and notably the 3D task exhibits a lower 
	peak than the 2D case, demonstrating that slack usage does not 
	grow with problem dimensionality. Furthermore, the peak slack consistently occurs 
	late in the ODE integration ($t_{\max} \in [0.56, 0.94]$ across all tasks), 
	consistent with our phased activation design, confirming that slack is never used to 
	arbitrarily warp the trajectory in the early generative phase.
	Taken together, these results do not support the interpretation that UniConFlow degenerates into a globally soft-penalty planner. 
	A more accurate interpretation is that slack acts only as a bounded auxiliary buffer whose relative contribution remains small across tasks, while the main correction is still carried by the PTZF-guided dynamics and the final enforcement stage.
	
	\subsection{Training Details}
	The training hyperparameters for the double inverted pendulum in
	\cref{subsec_double_pendulum}, the long-horizon car racing experiment
	(\cref{subsec_car_racing}), and the robotic manipulation task
	(\cref{subsec_manipulator}) are summarized as follows.
	For all tasks, we split the dataset into $90\%$ training and $10\%$
	validation trajectories.
	Training is performed on a workstation running Ubuntu~22.04 LTS with an
	AMD Ryzen 9 7950X CPU, 64\,GB RAM, and an NVIDIA GeForce RTX~3090~Ti
	GPU (24\,GB GDDR6X).
	
	All diffusion-based and flow-matching models are trained with AdamW and a constant learning rate of $2\times 10^{-4}$ without any learning rate schedule.
	We use mini-batches of size $64$ for all experiments.
	The total numbers of optimization steps are
	$5\times 10^{4}$ for the pendulum task,
	$1\times 10^{5}$ for the raceline task, and
	$2\times 10^{5}$ for the manipulator task, which corresponds to roughly $500$ epochs for the diffusion models and about $1200$ epochs for the FM models on each dataset.
	
	
	\section{Discussion, Limitation, and Future Work}
	
	In this paper, the training dataset is established by solving OCPs or MPC to track the desired trajectories or positions. Although diffusion models are able to deliver near-globally optimal outcomes in far less time than conventional methods, there is no guarantee that Generative models can generate truly optimal solutions for OCPs or MPC. Notably, post-processing approaches that combine with traditional solvers struggle with high-dimensional, long-horizon problems with constraints and often fail to complete within practical time limits—particularly in dynamic systems and robotics, where real-time constraints make them unusable. UniConFlow is designed as a model-agnostic safety layer that operates directly on the sampling ODE/SDE during inference. Because the QP-based guidance mechanism acts on the continuous-time denoising trajectory rather than on the network architecture itself, it is compatible with any diffusion- or FM-based planner or policy, such as Diffusion Policy~\cite{chi2023diffusionpolicy} and DiT-based Vision-Language-Action models~\cite{yu2026safenightvlaseeingunseen}. This plug-and-play characteristic means that safety and kinodynamic consistency can be retrofitted onto pretrained generative policies without any model retraining or architectural modification.
	
	\todo{
	The certification results established for the PTZF-guided flow are continuous-time results under the assumptions of the theoretical analysis. In implementation, the guided ODE is evaluated with a finite-step numerical solver, and small residual violations may remain because of time discretization, finite-precision arithmetic, and numerical gradient evaluation. Therefore, the Stage-1 output should be interpreted as a constraint-aware generative trajectory and an effective warm start for final certification. The terminal refinement stage closes this numerical gap and returns the fully certified trajectory of the complete UniConFlow pipeline. Table III shows that Stage 1 already achieves high certified rates before refinement, demonstrating that PTZF guidance performs the primary constraint-shaping role, while terminal refinement provides strict numerical enforcement. A promising direction for future work is to develop discretization-aware certificates that explicitly account for numerical integration errors.
	}
	
	\todo{
	A further implementation-level limitation arises in singular or near-singular configurations, where the constraint Jacobians used by the QP guidance may become rank deficient or numerically ill-conditioned. In the current implementation, parallel resampling is used to generate multiple candidate trajectories and retain a feasible candidate for which the guidance computation remains numerically well-conditioned. Although this strategy improves practical robustness, it should be regarded as an engineering safeguard rather than a fundamental mathematical resolution of singularity. Future work will investigate rank-aware regularization, damped pseudoinverse constructions, and manifold-based guidance formulations that explicitly account for singular constraint geometry.
	}
	
	In addition to these numerical considerations, our real-world deployment on the Franka Emika arm reveals a discrepancy between the generated state sequences and the actual robot execution. This discrepancy primarily arises from the sim-to-real dynamics gap, as unmodeled effects, such as joint friction and actuator dynamics, cause the physical system to deviate from the nominal model used during trajectory generation. Although the current framework ensures theoretical consistency with respect to the nominal dynamics under the stated assumptions, it does not explicitly account for unstructured physical uncertainties during inference. Future research could extend UniConFlow with robust trajectory generation capabilities by incorporating dynamics randomization, bounded model uncertainty, or uncertainty-aware guidance into the flow-matching vector field.
	
	\todo{
	Finally, the current implementation requires explicit and differentiable constraint gradients. In the real-world experiments, irregular physical obstacles are therefore represented by conservative analytical approximations, such as ellipsoids or unions of ellipsoids. These representations provide smooth collision gradients and mitigate the non-differentiability and measurement noise associated with raw, unstructured perception data. However, they may introduce conservative safety margins, obscure fine geometric details, and reduce the usable workspace. Accordingly, the current safety certificate is established with respect to the adopted analytical obstacle representation and its prescribed safety margin, rather than directly with respect to arbitrary raw sensor observations. Future work will investigate continuous differentiable scene representations and uncertainty-aware neural distance fields that can provide smooth collision gradients directly from sensor observations while reducing geometric conservatism.
	}
	
\end{document}